\documentclass{article} % For LaTeX2e
\usepackage{iclr2027_conference,times}
\usepackage{hyperref}
\usepackage{url}

\usepackage[utf8]{inputenc} % allow utf-8 input
\usepackage[T1]{fontenc}    % use 8-bit T1 fonts
\usepackage{graphicx}
\usepackage{subcaption}
\usepackage{booktabs} % for professional tables
\usepackage{enumitem}
\usepackage{amsfonts}       % blackboard math symbols
\usepackage{nicefrac}       % compact symbols for 1/2, etc.
\usepackage{microtype}      % microtypography
\usepackage{xcolor}         % colors
\usepackage{empheq}
\usepackage{standalone}
\usepackage{multirow}
\usepackage{siunitx}
\usepackage{wrapfig}
\usepackage{xspace}

\usepackage{algorithm}
\usepackage{algorithmic}

\usepackage{amsmath}
\usepackage{amssymb}
\usepackage{mathtools}
\usepackage{amsthm}

\theoremstyle{plain}
\newtheorem{theorem}{Theorem}[section]
\newtheorem{proposition}[theorem]{Proposition}

\theoremstyle{definition}

\theoremstyle{remark}
\newtheorem{remark}[theorem]{Remark}

\usepackage[capitalize,noabbrev]{cleveref}

\crefname{property}{Property}{Properties}

\newcommand{\GaussianBlobs}{\texttt{GaussianBlobs}\xspace}
\newcommand{\Pinball}{\texttt{Pinball}\xspace}
\newcommand{\Cylinder}{\texttt{CylinderFlow}\xspace}
\newcommand{\DeepJEB}{\texttt{DeepJEB}\xspace}

\title{GRIFDIR: Graph Resolution-Invariant Diffusion Models over Irregular Domains}
\author{%
James Rowbottom \\
Department of Applied Mathematics and Theoretical Physics\\
University of Cambridge\\
UK \\
\AND
Elizabeth L.~Baker\\
Department of Applied Mathematics and Computer Science\\
Technical University of Denmark\\
Denmark \\
\And
Nick Huang \\
Department of Mathematics\\
Simon Fraser University\\
Canada \\
\And
Ben Adcock \\
Department of Mathematics\\
Simon Fraser University\\
Canada \\
\And
Carola-Bibiane Schönlieb \\
Department of Applied Mathematics and Theoretical Physics\\
University of Cambridge\\
UK \\
\And
Alexander Denker \\
Helmholtz Imaging \\
Deutsches Elektronen-Synchrotron DESY, Germany \\
}

\newcommand{\data}{\text{data}}
\newcommand{\rbb}{\mathbb{R}}
\newcommand{\drm}{\mathrm{d}}
\newcommand{\pbb}{\mathbb{P}}
\newcommand{\ebb}{\mathbb{E}}

\newcommand{\imagespace}{\vspace{-0.4cm}}

\newcommand{\R}{\mathbf{R}}

\iclrfinalcopy % Uncomment for camera-ready version, but NOT for submission.
\begin{document}

\maketitle

\begin{abstract}
Score-based diffusion models in infinite-dimensional function spaces provide a mathematically principled framework for modelling function-valued data, offering key advantages such as resolution invariance and the ability to handle irregular discretisations. 
However, practical implementations have struggled to fully realise these benefits. 
Existing backbones like Fourier neural operators are often biased towards regular grids and fail to generalise to complex domain topologies. 
We introduce an architecture, {GRIFDIR}, for function-space diffusion models that represents generalised graph convolutional kernels as finite element functions, 
allowing the score network to operate directly on unstructured meshes over domains of arbitrary shape.
We demonstrate the efficacy of our network architecture through a series of unconditional and conditional sampling experiments across diverse geometries, including non-convex and multiply-connected domains.
Our results show that the proposed method maintains resolution invariance and achieves high fidelity in capturing functional distributions on non-trivial geometries.
% {\color{red} add \textbf{GRIFDIR}}
\end{abstract}

\section{Introduction}
Score-based diffusion models have recently been extended to infinite-dimensional Hilbert spaces, providing a principled framework for modelling function-valued data \citep{lim2025score,pidstrigach2024infinite,hagemann2025multilevel}. 
This is particularly relevant for scientific machine learning applications, where data arises naturally as solutions to partial differential equations (PDEs), spatial fields, or discretisations of continuous operators \citep{yao_guided_2025,baker2026supervised,baldassari2023conditional}.
Instead of modelling finite-dimensional vectors, infinite-dimensional diffusion models learn distributions directly over functions.

There are two main approaches for applying diffusion models to function-valued data. 
The first is the \emph{discretise-first} approach, where functions are discretised onto a fixed grid and standard finite-dimensional diffusion models are applied. 
The second is the \emph{discretise-last} approach, where diffusion models are formulated directly in function space and discretisation is introduced only at training or sampling time.
This provides several desirable properties:
\begin{enumerate}[nosep,leftmargin=*]
    \item \textbf{Resolution invariance}: Models trained at one resolution can generalise to different resolutions. 
    \item \textbf{Handling irregular data}: 
    % Function-space 
    Models naturally support unstructured meshes and complex geometries.
    \item \textbf{Theoretical guarantees:} Convergence results hold independently of the discretisation resolution.
\end{enumerate}
Furthermore, modelling directly in function spaces allows the incorporation of prior knowledge about the structure of the data, such as smoothness or Hilbert space constraints, which can improve both modelling flexibility and theoretical convergence guarantees \citep{pidstrigach2024infinite}.
Despite these advantages, existing numerical implementations do not fully realise these theoretical benefits. 
Convolutional U-Net architectures perform well on fixed, regular grids but lack resolution invariance and struggle with irregular geometries \citep{ronneberger2015u,hagemann2025multilevel}. 
Fourier Neural Operators (FNOs) \citep{li2020fourier}, on the other hand, provide resolution invariance but do not naturally handle irregular geometries or even non-regular grids. 
There is still a gap between the theoretical promises of infinite-dimensional diffusion models and practical architectures that fully exploit these promises.
\vspace{-0.1cm}

\begin{figure}[t]
    \centering
    \includegraphics[width=1.0\linewidth]{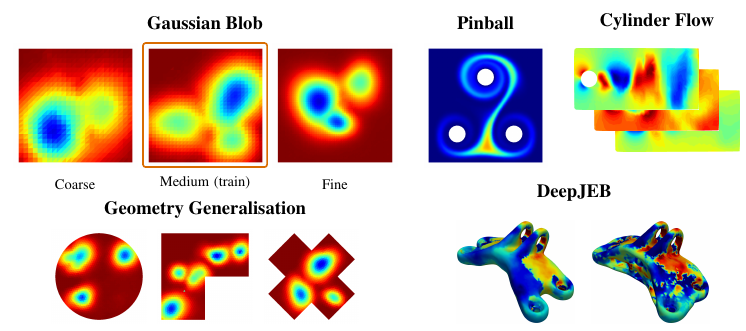}
    % \vspace{-1cm}
    \caption{GRIFDIR allows sampling from different resolutions (see \GaussianBlobs) without retraining, as well as works on non standard domains, e.g., including holes and non-convex geometries. Further, GRIFDIR scales to 3D generation. Details for these experiments in Section~\ref{sec:results}.}
    \label{fig:overview}
    \imagespace
\end{figure}

\paragraph{Contribution} 
We identify key properties that a neural network architecture should satisfy in order to fully leverage the advantages of the infinite-dimensional diffusion model paradigm over the finite-dimension paradigm, including resolution invariance and support for irregular domain geometries. We explore the abilities of existing models to fulfil these key properties.
We further propose \textbf{GRIFDIR} (\underline{G}raph \underline{R}esolution-\underline{I}nvariant \underline{F}EM \underline{D}iffusion over \underline{Ir}regular domains), a graph neural operator for infinite-dimensional diffusion models. 
Our approach is designed to satisfy the identified desirable properties, while remaining computationally efficient and easy to train. 
We demonstrate that the proposed architecture enables practical function-space diffusion modelling across varying domains and discretisations (see \cref{fig:overview}). 
We further present conditional sampling results. 

% We propose \textit{GRIFDIR}, a graph neural operator for infinite-dimensional diffusion models. 
% Our approach is designed to satisfy key desirable properties, including resolution invariance and support for irregular domain geometries, while remaining computationally efficient and easy to train. 
% We demonstrate that the proposed architecture enables practical function-space diffusion modelling across varying domains and discretisations. 
% We further present conditional sampling results with this new model. 

The paper is structured as follows: 
In \cref{sec:inf_dim_diffusion} we provide background on infinite-dimensional diffusion models and recent conditional sampling techniques. 
In \cref{sec:architecture} we provide desiderata on suitable network architectures and explain the GNN-based network architecture. In \cref{sec:experiment_setup} we explain the experiment setups and in \cref{sec:invariance,sec:geometry,sec:realistic}, we show results on varying domains and datasets. 
% We discuss additional related work in Appendix~\ref{app:realted_work}.

\section{Infinite-dimensional Diffusion Models}
\label{sec:inf_dim_diffusion}
Similarly to finite-dimensions, we define score-based diffusion models in infinite-dimensions with a forward and reverse stochastic differential equation (SDE). 
However, the SDEs are defined directly in a Hilbert space $\mathcal{H}$ and some care has to be taken in defining the time-reversal since Lebesgue measures (and therefore the standard densities) do not exist in Hilbert space. 
We will refer to elements in the Hilbert space by $a \in \mathcal{H}$ and with $A$ (and $A_t$) to random variables with realisations in $\mathcal{H}$.

Following \citet{pidstrigach2024infinite} we consider the forward SDE starting from the prior data distribution $\pi$, with support in $\mathcal{H}$, with $C$ a trace-class covariance operator and $W^\mathcal{H}$ the cylindrical Wiener process associated to $\mathcal{H}$
\begin{align}
    \label{eq:inf_dim_forward}
    d A_t = - \frac{1}{2} A_t  dt + \sqrt{C}dW^\mathcal{H}_t, \quad A_0 \sim \pi.
\end{align}
The time-reversal $Z_t \coloneqq A_{T-t}$ is given by
\begin{align}\label{eq:backwardSDE}
    \drm Z_t &= \left[\frac{1}{2}Z_t + s(T-t, Z_t)\right] \drm t +\sqrt{C}\drm W^\mathcal{H}_t,\\
    s(t, a) &\coloneqq -{(1-e^{-t})}^{-1}\left(a - e^{-\frac{t}{2}}\mathbb{E}[A_0 | A_t = a]\right)\label{eq:score},
\end{align}
where $Z_0 \sim \text{Law}(A_T)\approx \mathcal{N}(0,C)$ and $Z_T \sim \pi$.
Alternatively, we can formulate the Variance Exploding (VE) SDE \citep{song2021scorebased} in this infinite-dimensional setting. 
Let $\sigma_t > 0$ be a monotonically increasing noise schedule. 
Then the forward VE-SDE is given by
\begin{align}
    \label{eq:inf_dim_forward_ve}
    \drm A_t = g_t \sqrt{C} \drm W^\mathcal{H}_t, \quad A_0 \sim \pi, \text{ where } g_t \coloneqq \sqrt{\frac{\drm}{\drm t}\sigma_t^2.}
\end{align}
The transition measure is given by $\text{Law}(A_t \mid A_0) = \mathcal{N}(A_0, \sigma_t^2 C)$. 
Similarly to the previous case, we define a score operator $s_{\text{VE}}(t, a)$ which absorbs $C$:
\begin{align}
s_{\text{VE}}(t, a) &\coloneqq -\frac{1}{\sigma_t^2}\left(a - \mathbb{E}[A_0 \mid A_t = a]\right)\label{eq:score_ve},\\
\drm Z_t &= g_{T-t}^2 s_{\text{VE}}(T\!-\!t, Z_t) \drm t + g_{T-t} \sqrt{C} \drm W^\mathcal{H}_t,\label{eq:backwardSDE_ve}
\end{align}
where $Z_0 \sim \text{Law}(A_T) \approx \mathcal{N}(0, \sigma_T^2 C)$ and $Z_T \sim \pi$. 
One can then use the infinite-dimensional version of score matching to learn $s(t,a)$ or $s_\text{VE}(t,a)$ from data.

\subsection{Conditional Sampling}
\label{sec:conditional_sampling}
One class of function-valued sampling problems arises from (Bayesian) inverse problems. Let $L:\mathcal{H} \to \mathcal{Y}$ be a forward operator between Hilbert spaces, and the $Y = \mathcal{L}(a_0) + \eta$, with $\eta \in \mathcal{Y}$ being observation noise and $a_0 \sim \pi$. 
The inverse problem is to infer $a_0$ given some observation $Y=y$, which we model as sampling from the posterior distribution $\pi^y \coloneqq \pi_{| Y=y}$. 
Following \citet{Dashti2017}, we assume that $\pi_y \ll \pi$ for $y$ $\pi$-almost surely. 
Then, for some potential $\Phi: \mathcal{H} \times \mathcal{Y} \to \rbb$, we write the likelihood as 
\begin{align}\label{eq:likelihood}
\frac{\drm \pi^y}{\drm \pi}(a_0) = \frac{1}{\xi}\exp(-\Phi(a_0, y)), \text{ where } \xi = \ebb_{a_0\sim \pi}[\exp(-\Phi(a_0, y))].
\end{align}
To sample from the conditional distribution $\pi^y$ using diffusion models, one replaces the unconditional score in the time-reversed SDEs with a conditional score $s^y(t, a)$. 
While one can directly train a conditional model if paired data is available \citep{baldassari2023conditional}, a significant advantage of diffusion models is the ability to leverage a pre-trained, unconditional prior to solve various inverse problems without complete retraining. 
 
In finite-dimensional settings, conditional sampling fundamentally relies on Bayes' theorem (see Appendix~\ref{app:finite_dim_story}). In function spaces, this relationship can be formalised via the $h$-transform \citep{rogers2000diffusions}.
The conditional score splits into the unconditional score and a guidance term:
\begin{align}
    s^y(t, a) &= s(t,a) + C \nabla \log h^y(t, a), \\
    h^y(t, a) &\coloneqq \xi^{-1}\ebb_{\pbb}\!\left[\exp(-\Phi(A_0, y))\mid A_t = a\right],\label{eq:guidance_inf}
\end{align}
where $C$ denotes the covariance operator, $\mathbb{P}$ is the path measure of the forward SDE, and $\Phi$ is the potential defined in \eqref{eq:likelihood}; see Theorem 3.2 in \citet{baker2026supervised} for a proof. 
Substituting this conditional score into the time-reversed SDEs yields samples from the posterior $\pi^y$. 

Because the expectation in the guidance term \eqref{eq:guidance_inf} is generally intractable, practical implementations require approximations. Broadly, methods either circumvent the guidance term by learning a joint embedding \citep{yao_guided_2025}, utilise training-free approximations like the Tweedie estimate \citep{lin2026decoupled}, or perform supervised fine-tuning to explicitly learn the guidance \citep{baker2026supervised}. We defer a detailed discussion of these conditional sampling paradigms to Appendix~\ref{app:bg_guidance}.

\section{Architecture for Infinite-Dimensional Diffusion Models}
\label{sec:architecture}

\subsection{Desirable Properties for the Diffusion Network}
\label{sec:desirable_properties}
A central challenge in the infinite-dimensional framework is to parametrise the score function as a \textit{neural operator} $s_\theta: [0,T] \times \mathcal{H} \to \mathcal{H}$ that maps between function spaces \citep{kovachki2023neural}. 
While neural operators are most prominently used to learn solution operators of PDEs \citep{azizzadenesheli2024neural}, they provide a natural backbone for parametrising the infinite-dimensional score. 
To fully leverage the infinite-dimensional framework on irregular data, the score network should exhibit four key properties:
\begin{enumerate}[nosep,leftmargin=*]
    \item \label[property]{property:multiscale}\textbf{Multiscale processing:} Local fine-scale features and global structures must be captured, analogously to U-Nets.
    \item \label[property]{property:resolution}\textbf{Resolution and discretisation invariance:} The operator must define a mapping between function spaces. A model trained at resolution $N$ must generalise to $N' \neq N$ without retraining. 
    \item\label[property]{property:weird_domains} \textbf{Domain flexibility:} The architecture must be able to encode and decode functions on arbitrary and complex domain geometries without structural modification. 
    \item \label[property]{property:global} \textbf{Global structure:} 
    The model should capture and maintain global information about the target distribution when evaluating the score.
    %The model should maintain a compressed global representation of the target distribution. %We argument the bipartite nature of message passing GNNs does not possess this property in isolation.
    % The model should construct a compressed representation %at the coarsest scale to store and exchange global information about the learned distribution.
\end{enumerate}
The last point is essential, since the score function depends on the full input configuration, not only on local features. 
Architectures based solely on message passing propagate information through local interactions and therefore, lack an explicit global state (unless augmented with pooling or hierarchical mechanisms). 
As a result, they may struggle to capture long-range dependencies and global structure. 

\subsection{Existing Neural Operator Architectures}
\label{sec:related_operators}
\paragraph{Grid-based neural operators}
Standard U-Nets satisfy \cref{property:multiscale} but fail \cref{property:resolution} and \cref{property:weird_domains}, as convolutions are restricted to fixed regular grids. 
A widespread class of neural operators similarly relies on regular grids, including CNO~\citep{raonic2023convolutional} and FNO~\citep{li2020fourier}, which satisfy \cref{property:multiscale} and \cref{property:resolution} up to spectral truncation but fail \cref{property:weird_domains} on irregular geometries \citep{bartolucci2023reno}. 
While adaptations exist, e.g., domain masking, non-uniform Fourier transforms \citep{lingsch2023beyond}, or spatial deformation maps \citep{li2023fourier}, they introduce computational overhead and additional training complexity.

\paragraph{Message-Passing GNN}
Message-Passing (MP) GNNs offer a natural alternative to handle complex geometries. 
Methods like MeshgraphNets \citep{pfaff2020learning} and MP-PDEs \citep{brandstetter2022message} operate on unstructured meshes and thus satisfy domain flexibility. 
Given node features and physical coordinates, standard message passing iteratively updates each node by aggregating permutation-invariant messages from its neighbouring nodes and applying an update function.
% Given node features $a_i^{(k)}$ of node $i$ with physical coordinates $x_i$ at layer $k$, standard message passing updates the features via
% \begin{equation*}
%     a_i^{(k+1)} = f_\text{upd}^{(k)}\left(a_i^{(k)}, \bigoplus_{j \in \mathcal{N}(i)} f_\text{msg}^{(k)}\left(a_i^{(k)}, a_j^{(k)}, x_i, x_j\right)\right),
% \end{equation*}
% where $\mathcal{N}(i)$ denotes the immediate graph neighbours of node $i$, $f_\text{msg}^{(k)}$ is a message function, \smash{$\bigoplus$} is a permutation-invariant aggregation operator (such as the sum or mean), and $f_\text{upd}^{(k)}$ is an update function. 
Standard MP-GNNs fail to achieve resolution invariance (\cref{property:resolution}). 
A $K$-layer MP-GNN aggregates information over a physical radius proportional to $K \bar{d}$, where $\bar{d}$ is the average edge length of the mesh. 
As the mesh is refined ($\bar{d} \to 0$), the physical receptive field shrinks.

\paragraph{Graph Neural Operators}
Graph Neural Operators (GNOs) \citep{li2020neural,wen_geometry_2025} attempt to resolve the shrinking receptive field of standard MP-GNNs by decoupling the aggregation neighbourhood from the topological mesh. 
Instead of aggregating over immediate graph edges, GNOs aggregate over a fixed physical radius $r$. 
For a signal $a: \Omega \to \mathbb{R}^c$, the operator $\mathcal{K}$ is defined as
\begin{equation}
    \label{eq:gno_discrete}
    (\mathcal{K}a)(x_i) = \int_{B_r(x_i)} \kappa_\theta(x_i, y)\, a(y)\, \mathrm{d}y \approx \frac{1}{|N_r(i)|} \sum_{j \in N_r(i)} \kappa_\theta(x_i, x_j)\, a(x_j),
\end{equation}
where $\kappa_\theta$ is a learnable continuous kernel parametrised by a neural network, $B_r(x_i)$ is a ball of radius $r$ centred at~$x_i$, and $N_r(i)$ contains all nodes within $B_r(x_i)$. This formulation maintains a constant physical receptive field.

% For a signal $a: \Omega \to \mathbb{R}^c$, the operator $\mathcal{K}$ is defined as
% \begin{equation}
%     (\mathcal{K}a)(x_i) = \int_{B_r(x_i)} \kappa_\theta(x_i, y) a(y) \, \mathrm{d}y,
% \end{equation}
% where $\kappa_\theta$ is a learnable continuous kernel parametrised by a neural network, and $B_r(x_i)$ is a ball of radius $r$ centred at~$x_i$. 
% In practice, the integral is approximated by an empirical mean over nodes $N_r(i)$ within $B_r(x_i)$
% \begin{equation} \label{eq:gno_discrete}
%     (\mathcal{K}a)(x_i) \approx \frac{1}{|N_r(i)|} \sum_{j \in N_r(i)} \kappa_\theta(x_i, x_j) a(x_j),
% \end{equation}
% where $N_r(i)$ contains all nodes within the physical neighbourhood $B_r(x_i)$. 
% This formulation maintains a constant physical receptive field.

\paragraph{Continuous Kernels}
The choice of how to parametrise the kernel $\kappa_\theta$ is central to these operators. 
GNOs parametrise $\kappa_\theta$ as a continuous function of relative coordinates $x_j - x_i$. 
Common approaches include using B-spline bases as in SplineCNN \citep{fey2018splinecnn}, Gaussian mixture models in MoNet \citep{monti2017geometric}, or MLPs as seen in QuadConv \citep{doherty2023quadconv}, NNConv \citep{simonovsky2017dynamic}, PointNet++ \citep{qi2017pointnetpp}, or PointTransformer \citep{zhao2021point}.

% Finite Element Networks \cite{lienen2022learning} take a more principled approach,
% expanding $u(x) = \sum_j c_j\,\varphi^{(j)}(x)$ in a piecewise linear (P1) basis and recovering the Galerkin weak form $\mathbf{A}\,\partial_t \mathbf{c} = \mathbf{m}$ with mass matrix $\mathbf{A}_{ij} = \langle \varphi^{(i)}, \varphi^{(j)} \rangle_\Omega$. 
% However, computing $m_i = \langle F, \varphi^{(i)} \rangle_\Omega$ requires integrating a learned nonlinearity $F$ against each basis function, which is prohibitive; the authors therefore approximate $F$ as piecewise constant per mesh cell. 
% The Galerkin integral of a neural network against the basis is consequently avoided entirely.

\subsection{GRIFDIR: A Multiscale Diffusion GNN}
\label{sec:fem_conv}

\begin{figure}[t]
    \centering

    \begin{minipage}[t]{0.48\linewidth}
        \centering
        \includegraphics[width=\linewidth]{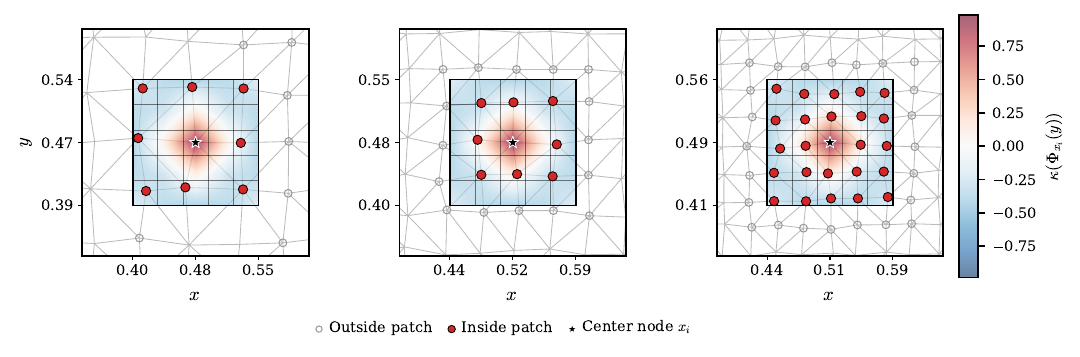}
        \caption{FEM convolution across increasing mesh resolutions. The reference patch %(black square, $w{=}0.15$) 
        and the learned filter (parameterised on a $5{\times}5$ grid) remain fixed in physical space, ensuring resolution-invariant aggregation over nodes (red dots).}
        \label{fig:fem_convolution}
    \end{minipage}
    \hfill
    \begin{minipage}[t]{0.48\linewidth}
        \centering
        \includegraphics[width=0.9\linewidth]{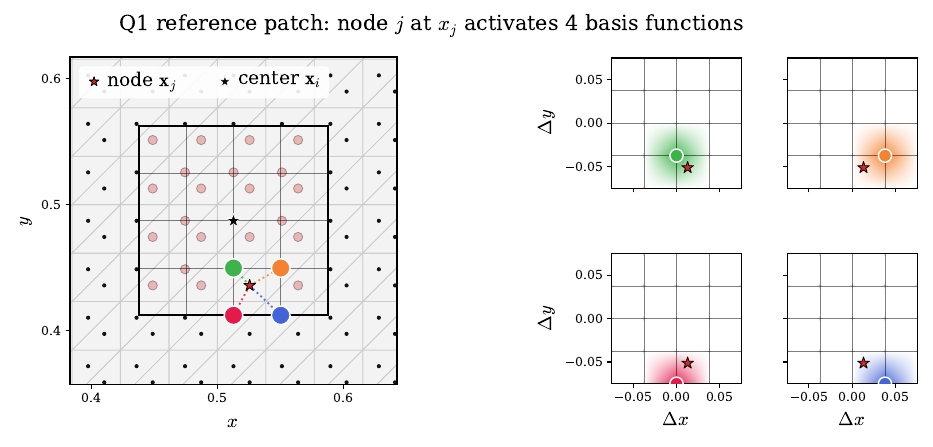}
        \caption{Q1 basis function for the FEM convolution for a reference patch. %of size $w=0.15$ (black square).
        Every node only activates four neighbouring Q1 basis functions (red, blue, orange, green).}
        \label{fig:fem_convolution_part2}
    \end{minipage}
    \imagespace
\end{figure}

We propose \textbf{GRIFDIR}: a multiscale graph neural operator defined over a hierarchy of meshes, combined with a finite-element-based convolution operator and a latent transformer. 
The architecture operates on functions defined over general meshes, achieving resolution-invariance (\cref{property:resolution}), support for complex geometries (\cref{property:weird_domains}), hierarchical multi-scale processing (\cref{property:multiscale}), and a global compressed latent representation via the transformer (\cref{property:global}).

We first fix the notation of the mesh. We will define the architecture first for domains $\Omega \subset \mathbb{R}^2$ and later generalise this to higher dimensions. 
% {\color{red} We can also directly write everything with $\Omega \subset \mathbb{R}^d$.} 
A discretisation of $\Omega$ is a finite set of nodes $\mathcal{T}=\{x_1, \dots, x_N\} \subset \Omega$ together with a finite element basis $\{ \psi_j\}_{j=1}^N$. The basis functions $\psi_j$ satisfy $\psi_j(x_k) = \delta_{jk}$ and are local, i.e., $\text{supp}(\psi_j)$ has a diameter $h_j$ around $x_j$ (for piecewise linear elements $h_j$ is the size of the largest triangle incident to $x_j$). We further assume that the basis functions are normalised, i.e., $\sum_{j=1}^N \psi_j(x) = 1$ for all $x \in \Omega$.

% A feature field $a : \Omega \to \mathbb{R}$ is represented on the mesh by its nodal values $\{a(x_j)\}_{j=1}^N$. 

\textbf{Resolution-invariant FEM convolution.}
We generalise CNN convolutions to unstructured meshes by parametrising the convolutional filter as a finite-element function on a fixed \emph{reference patch} in physical space, and projecting neighbouring nodes into the filter's coordinate frame, rather than evaluating a learned kernel in the coordinate frame of the domain. In particular, we define the \emph{reference patch} $\hat{\Omega} = [-1,1]^2 \subset \mathbb{R}^2$ with a regular $P \times P$ Cartesian grid $\{\xi_p\}_{p=1}^{P^2} \subset \hat\Omega$ and a bilinear (Q1) basis $\{\phi_p\}_{p=1}^{P^2}$ with $\phi_p(\xi_q) = \delta_{pq}$. The learnable convolutional kernel is then defined as 
\begin{align*}
    \kappa: \hat{\Omega} \to \mathbb{R}, \quad \kappa(\xi) = \sum_{p=1}^{P^2} w_p \phi_p(\xi)
\end{align*}
with $w=(w_1, \dots, w_{P^2}) \in \mathbb{R}^{P^2}$ as trainable parameters. For every node $x_i$, let $B_r(x_i) := \{\, y \in \Omega : \|y - x_i\|_\infty \le r \,\}$ be the physical patch around $x_i$ (a square of side $2r$, centred at $x_i$). We now use $\kappa$ to define our FEM convolution. We project the nodes to the references patch via a rescaling map 
\begin{align*}
    \Phi_{x_i} : B_r(x_i) \to \hat\Omega, \qquad \Phi_{x_i}(y) = \frac{y - x_i}{r}.
\end{align*}
We have $\Phi_{x_i}(x_i) = 0$. Composing the kernel with $\Phi_{x_i}$, i.e.\ $\kappa(\Phi_{x_i}(\cdot))$, evaluates the fixed reference kernel in physical coordinates centred and scaled at $x_i$. For a feature field $a : \Omega \to \mathbb{R}$, the FEM convolution of $a$ with $\kappa$ at node $x_i$ is
\begin{align}
    (\mathcal{K}a)(x_i) = \int_{B_r(x_i)} \kappa\bigl(\Phi_{x_i}(y)\bigr)\, a(y)\, \mathrm{d}y.
    \label{eq:fem_conv_integral}
\end{align}
We evaluate \eqref{eq:fem_conv_integral} with a nodal (mass-lumped) quadrature rule on the mesh, replacing $a$ by its interpolant $a_h = \sum_j a(x_j)\psi_j$ and using  $\int_{B_r(x_i)} \psi_j(y)\,f(y)\,\mathrm{d}y \approx V_{i,j}\,f(x_j)$, where $V_{i,j} := \int_{B_r(x_i)} \psi_j(y)\,\mathrm{d}y$ is the \emph{quadrature weight} of neighbour $j$ at node $i$:
\begin{align}
    (\mathcal{K}^{\text{FEM-Conv}}a)(x_i)\!:=\!\sum_{j \in N(i)} V_{i,j}\,\kappa\bigl(\Phi_{x_i}(x_j)\bigr)\,a(x_j)
\!=\!\sum_{p=1}^{P^2} w_p \sum_{j \in N(i)} V_{i,j}\,\phi_p\bigl(\Phi_{x_i}(x_j)\bigr)\,a(x_j),
    \label{eq:fem_message}
\end{align}
 where $N(i) = \{j: x_j \in B_r(x_i) \}$ are the neighbours of node $i$:
% The reference patch $\hat{\Omega} \subset \mathbb{R}^2$ has physical width $w = 2r$. 
% For each node $i$ at position $x_i$ we define the rescaling map
% \begin{align*}
%     \Phi_{x_i}: B_r(x_i) \to \hat{\Omega}, \quad \Phi_{x_i}(x) = \frac{x - x_i}{r}.
% \end{align*}
% Within $\hat{\Omega}$ we place a $P \times P$ regular grid and associate with each grid point $p$ a Q1 bilinear basis function $\phi_p: \hat{\Omega} \to \mathbb{R}$. 
% Writing the filter as $\kappa(\xi) = \sum_p w_p\,\phi_p(\xi)$, the convolution integral becomes
% \begin{align}
% \label{eq:fem_message}
% (\mathcal{K}a)(x_i)\!= \int_{B_r(x_i)} \kappa\bigl(\Phi_{x_i}(y)\bigr)\,a(y)\,\mathrm{d}y
%    &= \sum_{p=1}^{P^2} w_p \! \int_{B_r(x_i)}\!\phi_p\!\bigl(\Phi_{x_i}(y)\bigr)\,a(y)\,\mathrm{d}y
%    \\ &\approx \sum_{p=1}^{P^2} w_p \cdot \sum_{j \in N(i)} V_{i,j}
%         \phi_p\bigl(\Phi_{x_i}(x_j)\bigr)\,a(x_j),\nonumber
% \end{align}
% where $N(i) = \{j : x_j \in B_r(x_i)\}$ are all the nodes inside $B_r(x_i)$, $V_{i,j} = \int_{B_r(x_i)} \phi_j(x)dx$  
% and $w_p$ are the trainable weights of the FEM convolution. {\color{red} For $h \to 0 $ the FEM convolution converges to the true convolution, see Proposition \ref{prop:kernel_consistency}.}
% Note, the $V_{i,j}$ takes the density of nodes into account, and we refer to using this as \textit{lumped aggregation}. However, for uniform meshes, we also consider the more simple $V_{i, j} = \frac{1}{N(i)}$ which we call \textit{mean aggregation}.
A multi-channel extension mirrors a standard CNN and is described in Appendix~\ref{app:fem_conv_mult_channels}. We refer to the weighted aggregation using the quadrature weights $V_{i,j}$ as \textit{lumped aggregation}. The lumped aggregation is of particular importance for non-uniform node densities and ensures that the sum in \eqref{eq:fem_message} converges to the true integral in \eqref{eq:fem_conv_integral}.

\begin{remark}
The construction extends to $d$-dimensional domains by replacing the Q1 basis with a $d$-fold multilinear basis on a $P^{\times d}$ reference grid, giving $P^d$ filter weights and at most $2^d$ nonzero basis overlaps per neighbour. 
While this is tractable for the $2$ or $3$-dimensional domains, which are considered in recent applications \citep{yao_guided_2025,huang2024diffusionpde,ju2026function}, for $d \gg 3$ alternative methods have to be considered, e.g., based on sparse-grids \citep{bungartz2004sparse}. 
\end{remark}

    % \begin{wrapfigure}{r}{0.49\linewidth}
    % \centering
    % \includegraphics[width=1.\linewidth]{figures/q1_basis_activation.pdf}
    % \caption{Q1 basis function for the FEM convolution for a reference patch of size $w=0.15$ (black square). Every node $i$ only activates four neighbouring Q1 basis functions (red, blue, orange, green).}
    % \label{fig:fem_convolution_part2}
    % \end{wrapfigure}
    
% %OLD version
This parametrisation offers three advantages over MLP or spline kernels. First, it is a strict generalisation of standard convolutions, see Proposition \ref{prop:gen_cnns} for a formal statement.
%On a uniform Cartesian grid (with $r$ equal to half the grid spacing), each projected neighbour maps exactly to a grid point of the reference patch, and the FEM convolution reduces to a standard $P{\times}P$ discrete convolution. 
The architecture therefore inherits the inductive biases of CNNs that have proven effective in image-based diffusion models. %, and extends them to arbitrary unstructured geometries. %this 
Second, since each basis function $\phi_p$ is compactly supported, any projected neighbour $x_j$ has non-zero overlap with at most four basis functions via its bilinear stencil, see Figure~\ref{fig:fem_convolution_part2}. 
Each node in $B_r(x_i)$ thus contributes to exactly four filter weights, giving $\mathcal{O}(4\,|N(i)|)$ operations per node regardless of $P^2$, in contrast to MLP kernels which require a full forward pass per edge. Third, as the mesh is refined, the FEM-Conv \eqref{eq:fem_message} converges to the true integral, see Proposition \ref{prop:mesh_refinement_consistency} for a formal statement with proof.

\textbf{Multiscale mesh hierarchy.}
To enable multiscale processing (\cref{property:multiscale}) while preserving domain flexibility (\cref{property:weird_domains}), we embed the FEM convolution within a hierarchical mesh architecture. 
This multiscale processing has been widely used in GNNs \citep{gao2019graph,li2020multipole,wurth2025diffusion,bouziani_structure-preserving_2024}.
For any domain we construct a hierarchy of $L$ meshes $\{ \mathcal{T}^{(\ell)} \}_{\ell=0}^{L-1}$, where $\mathcal{T}^{(0)}$ is the finest level, yielding progressively smaller node sets $\mathcal{Z}^{(\ell)}$ with $N_0 > \dots > N_{L-1}$. 
Each mesh $\mathcal{T}^{(\ell)}$ induces a graph $\mathcal{G}^{(\ell)} = (\mathcal{Z}^{(\ell)}, \mathcal{E}^{(\ell)})$ with edges connecting adjacent nodes. 

\textbf{Coupling FEM convolutions to the hierarchy.} 
Both the physical radius $r^{(\ell)}$ and the patch width $w^{(\ell)}$ at each level $\ell$ are hyperperameters. We set  $r^{(\ell)} = \mu\,\tilde{d}^{(\ell)}$ and patch width $w^{(\ell)} = 2 r^{(\ell)}$, where $\tilde{d}^{(\ell)}$ is the median edge length (distance between nodes) of $\mathcal{T}^{(\ell)}$ with $\mu>0$. 
The same physical radius and patch width are then used for different meshes. 
The physical extent of the filter therefore grows with the coarseness of the mesh, mirroring the expansion of the receptive field of a U-Net. 
Transfers between mesh resolution (up- and downsampling) are implemented as learned restriction and prolongation maps. 
Noise (i.e., diffusion time) and positional context are reinjected at each level via FiLM modulation and learned per-level projections of the mesh coordinates. 
At the coarsest level, domain-conditioned cross attention bridges the $N_{L-1}$ coarse-mesh tokens to a latent Diffusion Transformer~\citep{peebles2023dit} with AdaLN-Zero time conditioning, which serves as the global latent memory of the model (\cref{property:global}). 
Full architectural details are provided in Appendix~\ref{app:architecture}.

\textbf{Conditioning.} We concatenate both the diffusion time (noise level) $t$, embedded via a Fourier embedding, and an optional condition vector $c \in \mathbb{R}^m$ (such as physical parameters for a family of PDEs) alongside the nodal coordinates $x_i$ to each node.The diffusion time is additionally re-injected at every hierarchy level via FiLM \citep{perez2018film}, and at the coarsest level via AdaLN-Zero \citep{peebles2023dit} in the latent transformer.

\section{Results}\label{sec:results}
We evaluate the different properties: invariance to resolution and mesh discretisation (\cref{sec:invariance}), generalisation across domain geometries (\cref{sec:geometry}), and scaling to realistic and higher-dimensional data (\cref{sec:realistic}). Beyond unconditional sampling, we evaluate conditional sampling for inverse problems. In the Appendix, \Cref{app:ablations} isolates the contribution of each architectural component, whilst \cref{app:conditional} compares approximations of the guidance term via Fun-DPS \citep{yao_guided_2025} and Fun-DAPS \citep{lin2026decoupled} (see \cref{app:bg_guidance} for further details).

\subsection{Experimental Setup}\label{sec:experiment_setup}

\textbf{Datasets} We make use of the following datasets, which are described in more detail in \cref{app:datasets}.
The first is \GaussianBlobs, see \cref{fig:gaussian_blob_gt} for some ground truth samples. 
The second is the \Pinball dataset \citep{tomasetto2025reduced}. 
Samples of the \Pinball dataset are shown in \cref{fig:pinball_dataset}. We also use the \Cylinder dataset \citep{han2022predicting} as used in \cite{shi2025meshinformed}.
Finally, we use \DeepJEB \citep{hong2025deepjeb}, a 3D dataset, to show how our method scales.

\textbf{Baselines} We compare our GRIFDIR with diffusion models based on MP-GNN \citep{brandstetter2022message}, GAOT \citep{wen_geometry_2025}, FNO \citep{li2020fourier},
% and Geo-FNO \citep{li2023fourier},
convolutional neural networks and MINO \citep{shi2025meshinformed}. 
These approaches are described in Appendix~\ref{app:baselines_arch}. 

\textbf{Diffusion model setup}
In our experiments, we use the VE-SDE \eqref{eq:inf_dim_forward_ve} using the specific parametrisation from \citet{karras2022elucidating}. 
We consider a noise schedule with $\sigma(0) = \sigma_\text{min} = 0.001$ and $\sigma_T=\sigma_\text{max} = 40.0$.
To construct the covariance operator $C$, we follow \citet{yao_guided_2025} 
% and use the family $C_\ell$ 
and use radial basis function kernels 
with length-scale $\ell=0.1$
\begin{align}
\label{eq:covariance}
K_\ell(x,y)\!\coloneqq\!\exp\!\left(-\frac{|x-y|^2}{2\ell^2}\right),  \qquad
(C_\ell f)(x) \coloneqq \int_\Omega K_\ell(x,y) f(y) dy.
\end{align}
For unconditional sampling, we use the Heun sampler \citep{karras2022elucidating,yao_guided_2025}.

% \textbf{Conditional Sampling}
% Alongside unconditional sampling, we consider sampling for inverse problems. 
% We consider two linear forward operators: a sparse point-wise sensor evaluation and a PDE-constrained operator governed by the Poisson equation. 
% To obtain the conditional samples, we use the \textit{diffusion prior and guidance} paradigm (see \cref{sec:conditional_sampling}), where we first train our model on the unconditional distribution $\pi$ using our network.
% In \cref{app:conditional}, we compare approximations of the guidance term via the Fun-DPS \citep{yao_guided_2025} and Fun-DAPS \citep{lin2026decoupled} frameworks. 
% For more details on these methods, we refer to \cref{app:bg_guidance}. 

\textbf{Metrics} To evaluate the methods we mainly consider the maximum mean discrepancy (MMD) \citep{gretton2012kernel} and the sliced Wasserstein distance (SWD) \citep{bonneel2015sliced}. We also compute two metrics for posterior sampling: the mean squared error (MSE) and the energy score (ES) between the ground truth and posterior estimates. Here, we consider conditional sampling for two linear inverse problems, a sparse point-wise sensor evaluation and a PDE-constrained operator governed by the Poisson equation, where we approximate the conditional score using Fun-DPS. For the MMD we use a Gaussian kernel with length-scale $\ell$. All metrics are described in detail in \cref{app:evaluation_metrics}.

% \textbf{Metrics} To evaluate the methods we mainly consider the maximum mean discrepancy (MMD) \citep{gretton2012kernel} and the sliced Wasserstein distance (SWD) \citep{bonneel2015sliced}. 
% We also compute two posterior metrics based on conditional sampling for sensors using Fun-DPS, where we then use the mean squared error (MSE) and the energy score (ES) between the ground truth and posterior estimates. For the MMD we use a Gaussian kernel with length-scale $\ell$. All metrics are described in  detail in \cref{app:evaluation_metrics}

\subsection{Resolution and Discretisation Invariance}
\label{sec:invariance}

\Cref{property:resolution} requires a single trained model to remain accurate both at unseen resolutions of a uniform mesh and under local changes in node density. We test these two notions separately.

\begin{figure*}[t]
    \centering
    \textbf{Training resolution (2048 elements)}\\[0.1cm]
    \begin{subfigure}{0.16\linewidth}
        \includegraphics[width=\linewidth]{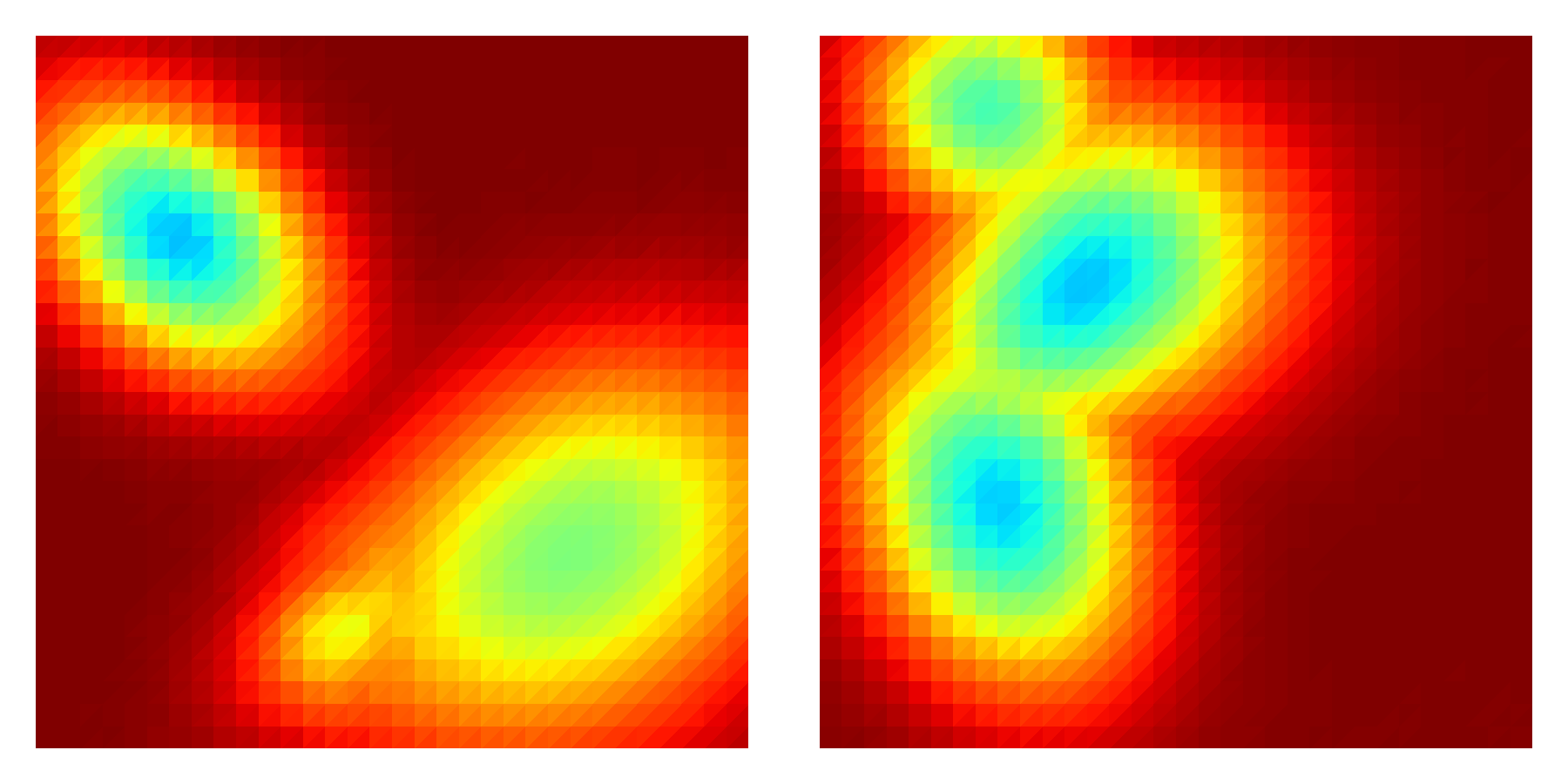}
    \end{subfigure}
    \begin{subfigure}{0.16\linewidth}
        \includegraphics[width=\linewidth]{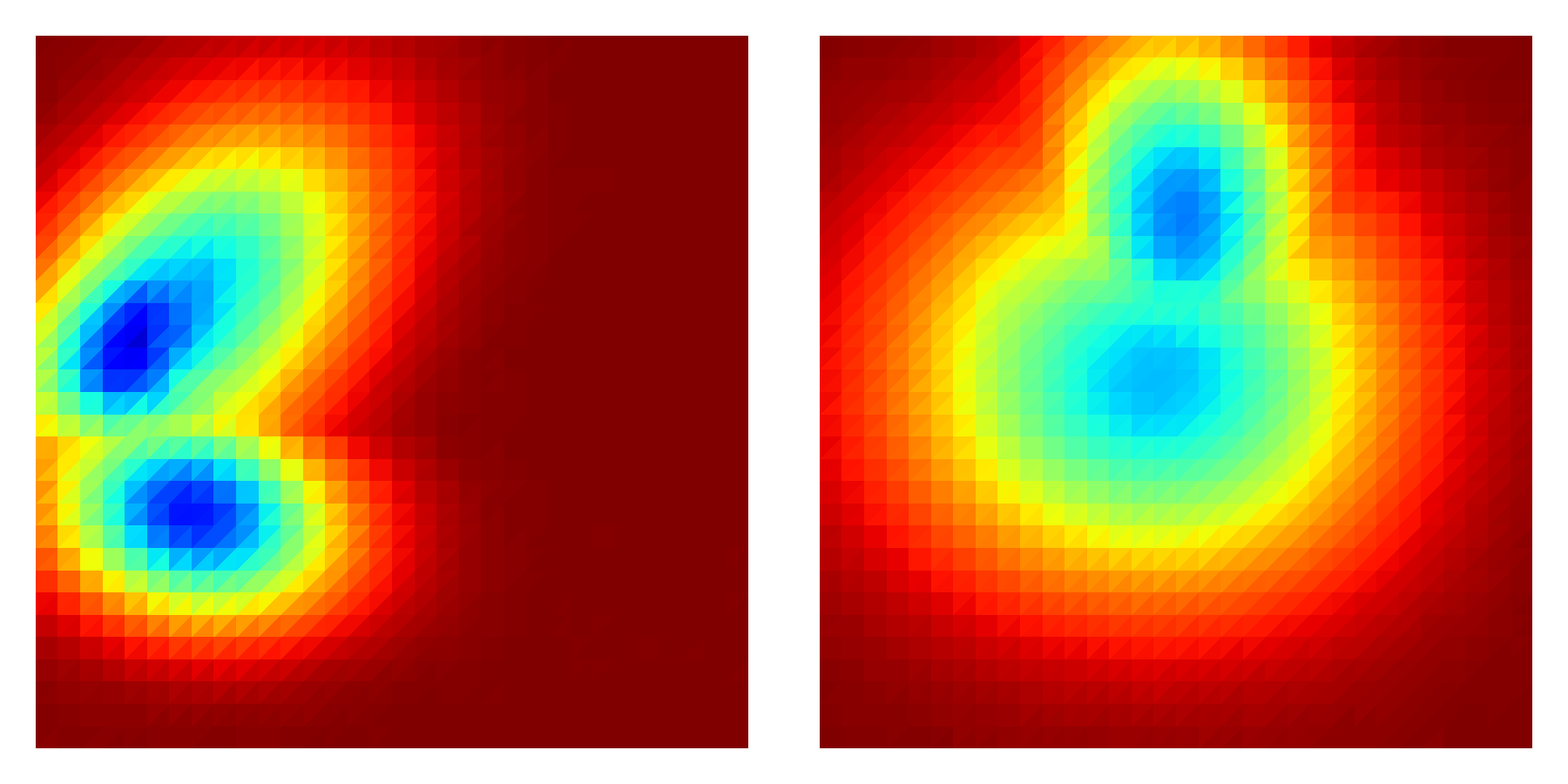}      
    \end{subfigure}
    \begin{subfigure}{0.16\linewidth}
        \includegraphics[width=\linewidth]{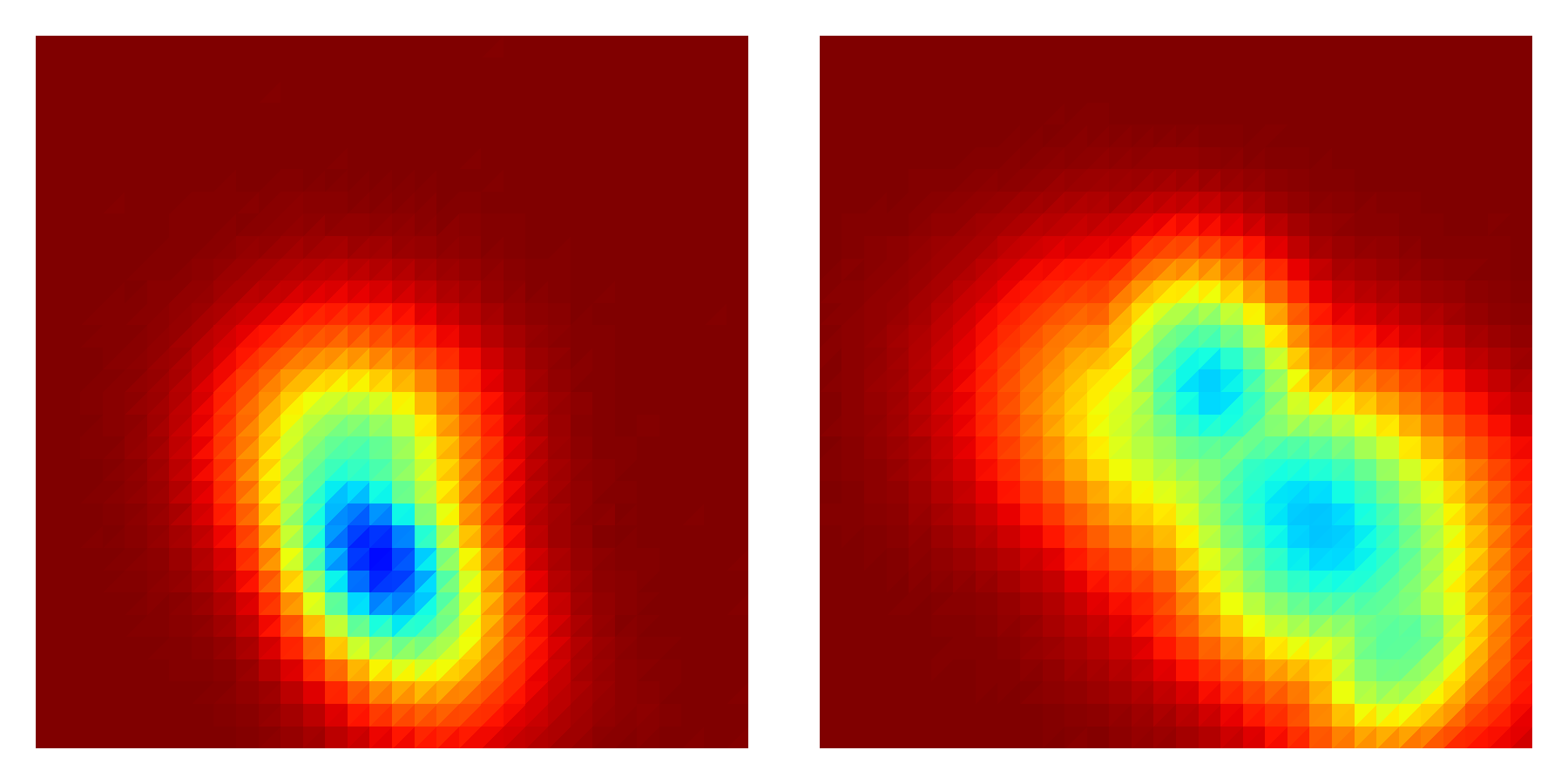}        
    \end{subfigure}
    \begin{subfigure}{0.16\linewidth}
        \includegraphics[width=\linewidth]{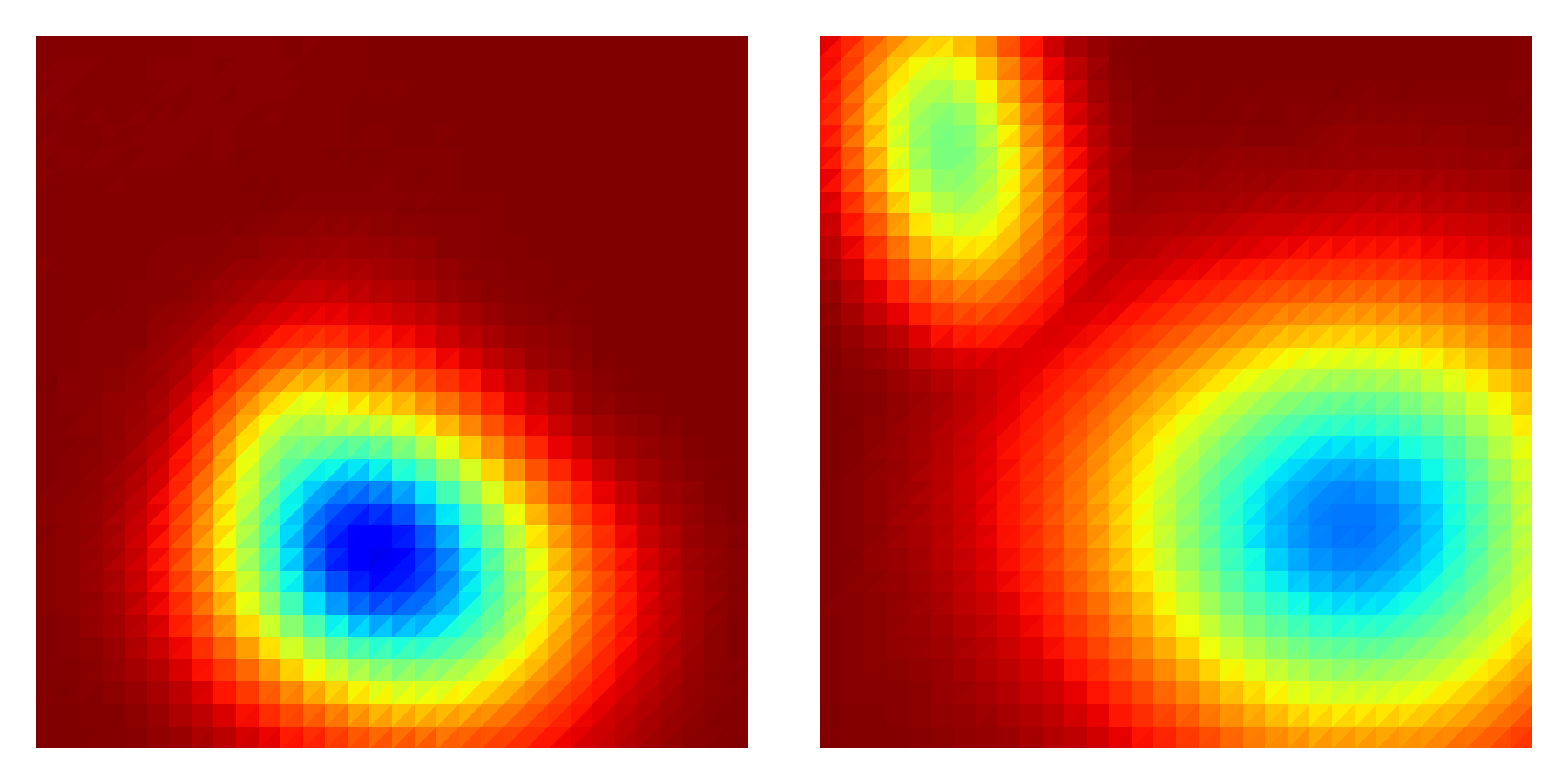}      
    \end{subfigure}
    \begin{subfigure}{0.16\linewidth}
        \includegraphics[width=\linewidth]{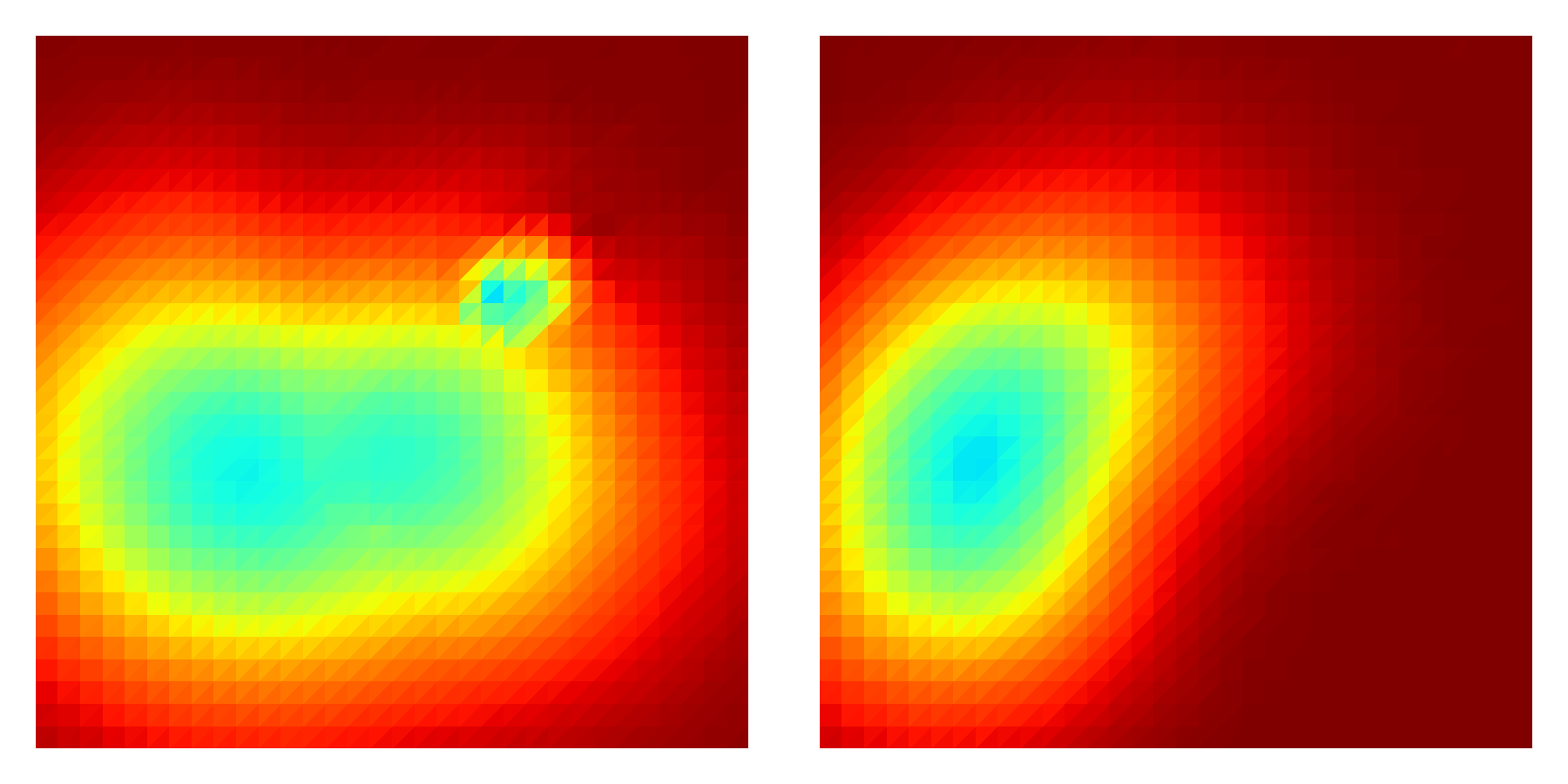}
    \end{subfigure}
    \begin{subfigure}{0.16\linewidth}
        \includegraphics[width=\linewidth]{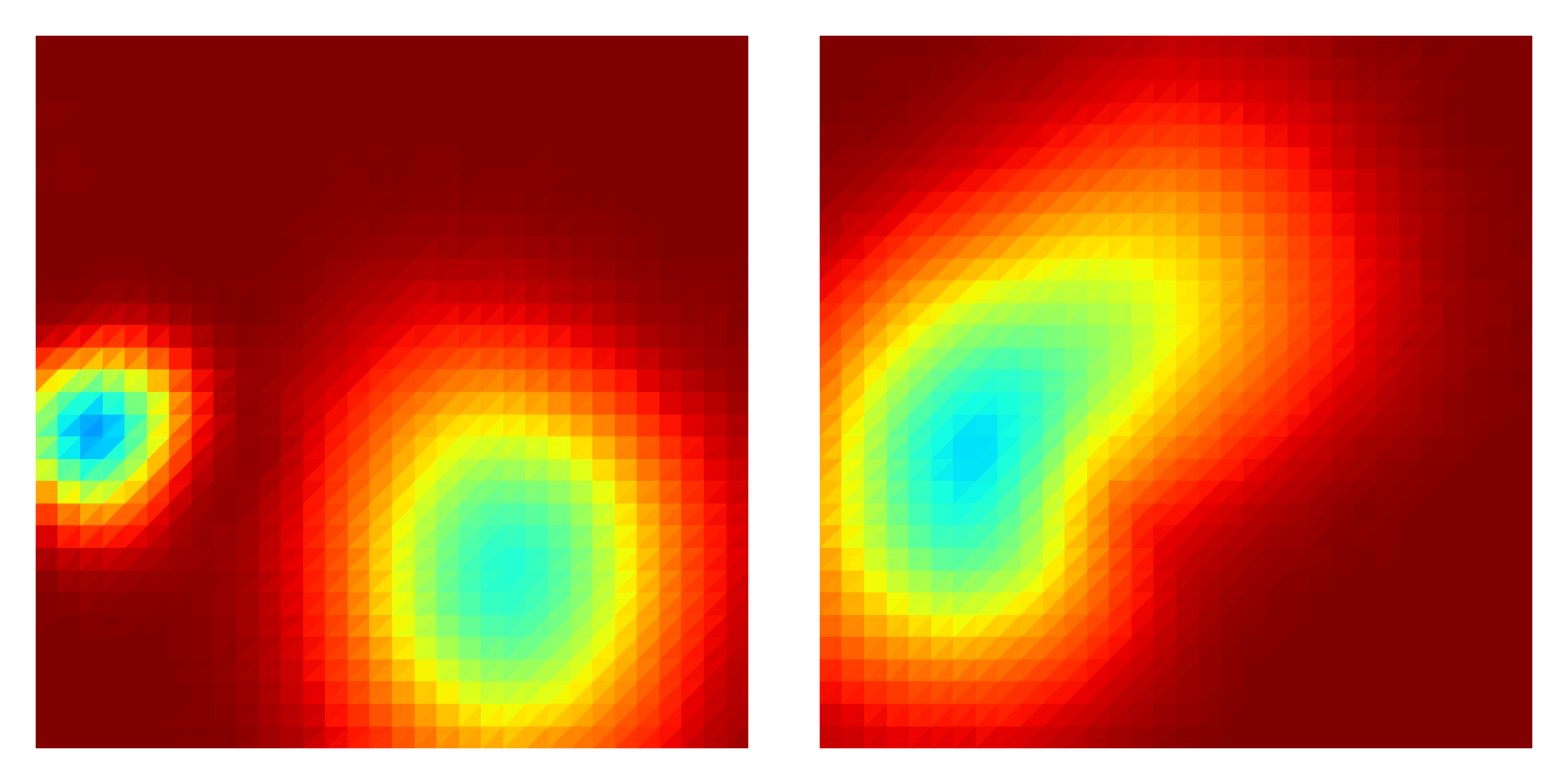}
    \end{subfigure}

    \textbf{Evaluation resolution (8192 elements)}\\[0.1cm]

    % Row 2
    \begin{subfigure}{0.16\linewidth}
        \includegraphics[width=\linewidth]{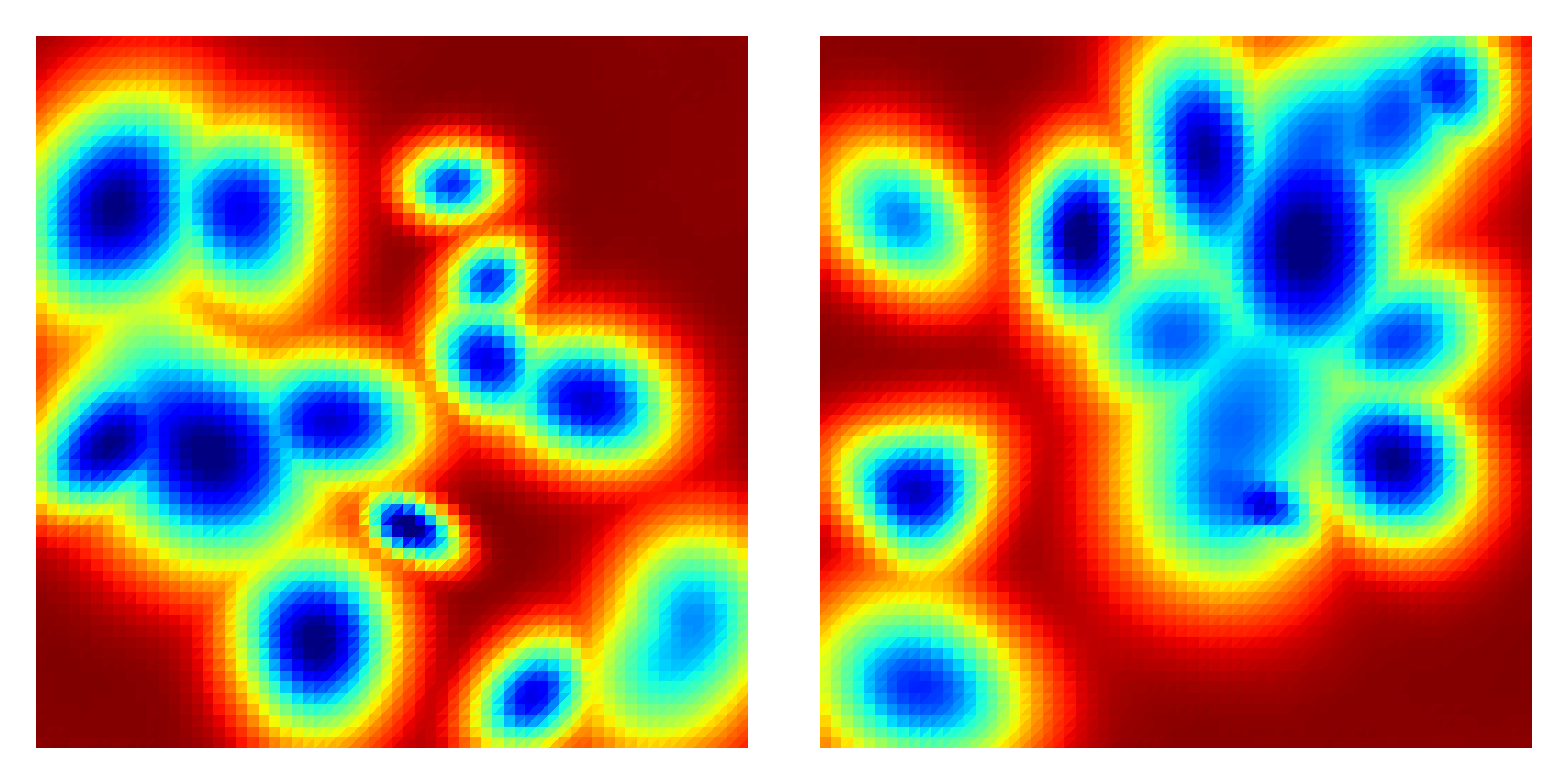}
        \caption*{CNN}
    \end{subfigure}
    \begin{subfigure}{0.16\linewidth}
        \includegraphics[width=\linewidth]{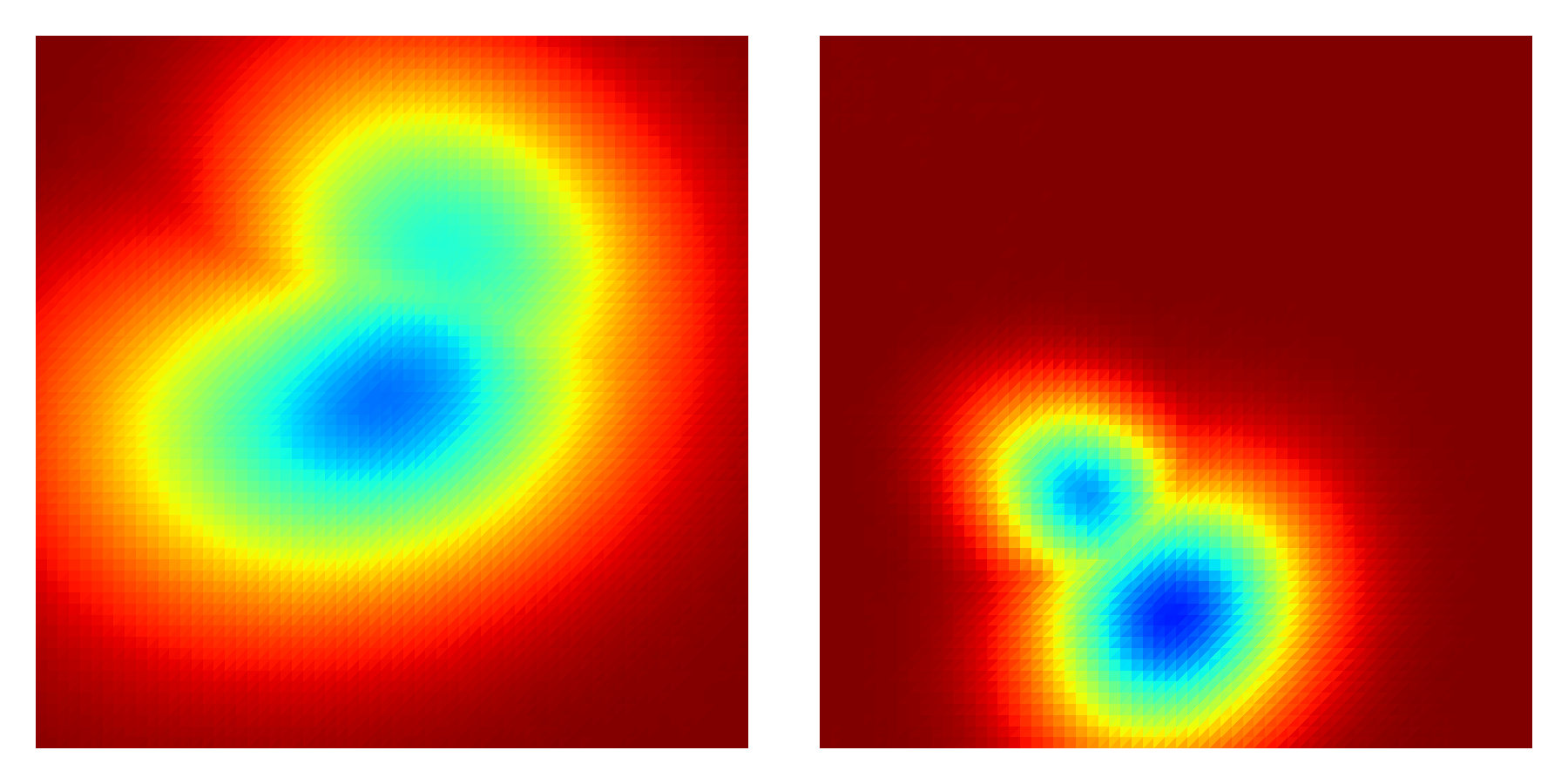}      
        \caption*{FNO}
    \end{subfigure}
    \begin{subfigure}{0.16\linewidth}
        \includegraphics[width=\linewidth]{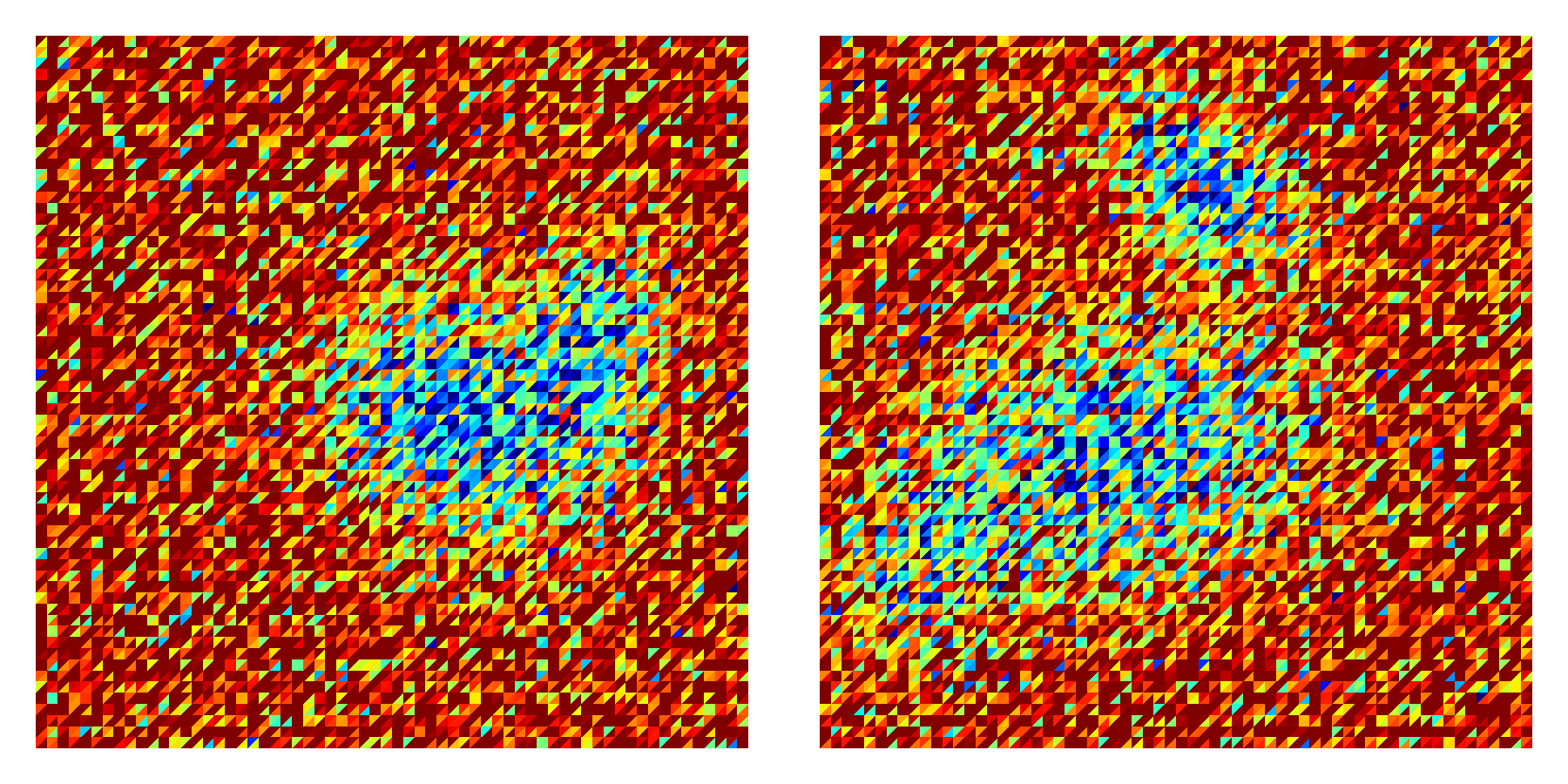}        
        \caption*{GAOT}
    \end{subfigure}
    \begin{subfigure}{0.16\linewidth}
        \includegraphics[width=\linewidth]{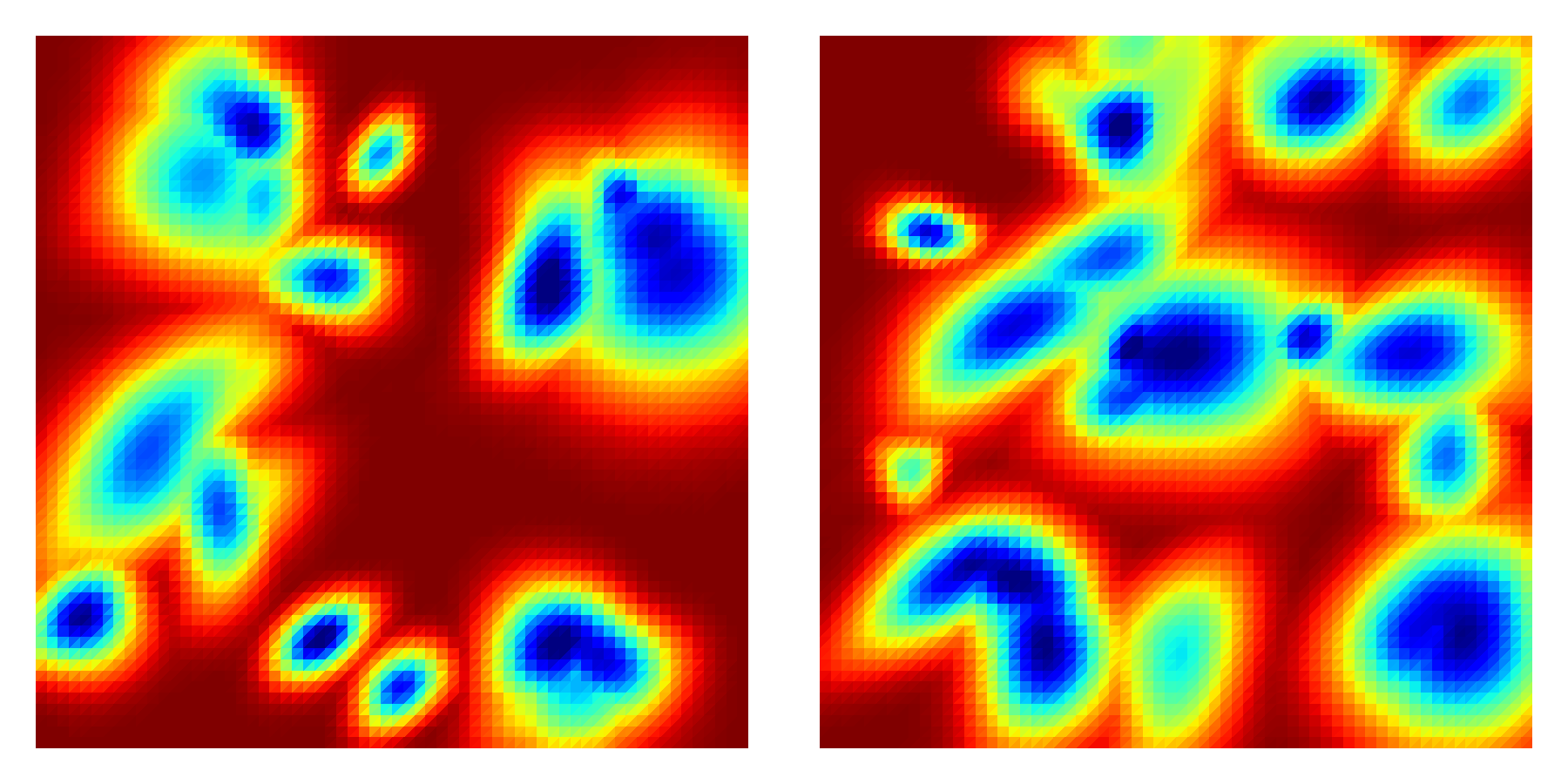}      
        \caption*{MP-GNN}
    \end{subfigure}
    \begin{subfigure}{0.16\linewidth}
        \includegraphics[width=\linewidth]{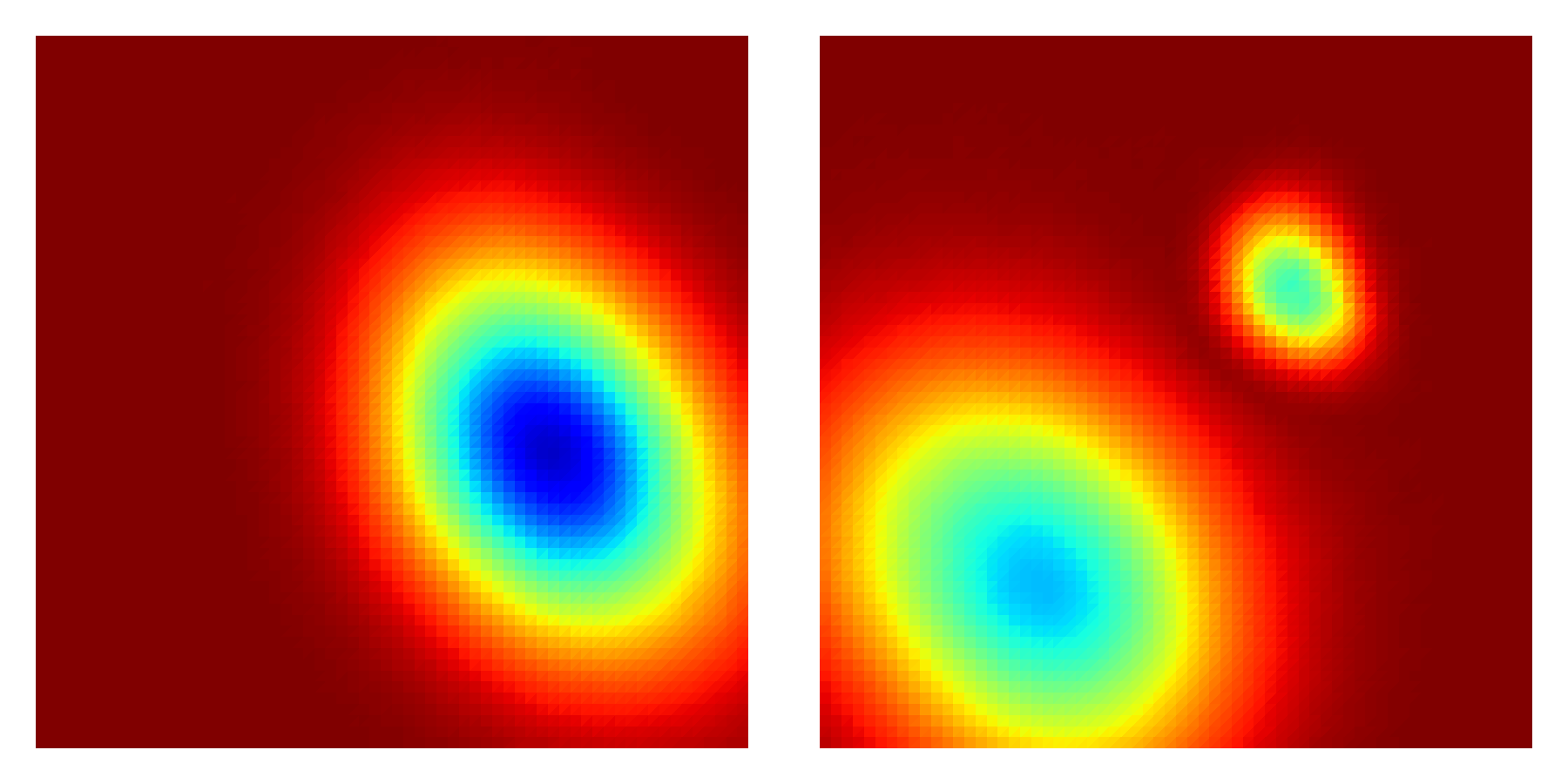}
        \caption*{MINO}
    \end{subfigure}
    \begin{subfigure}{0.16\linewidth}
        \includegraphics[width=\linewidth]{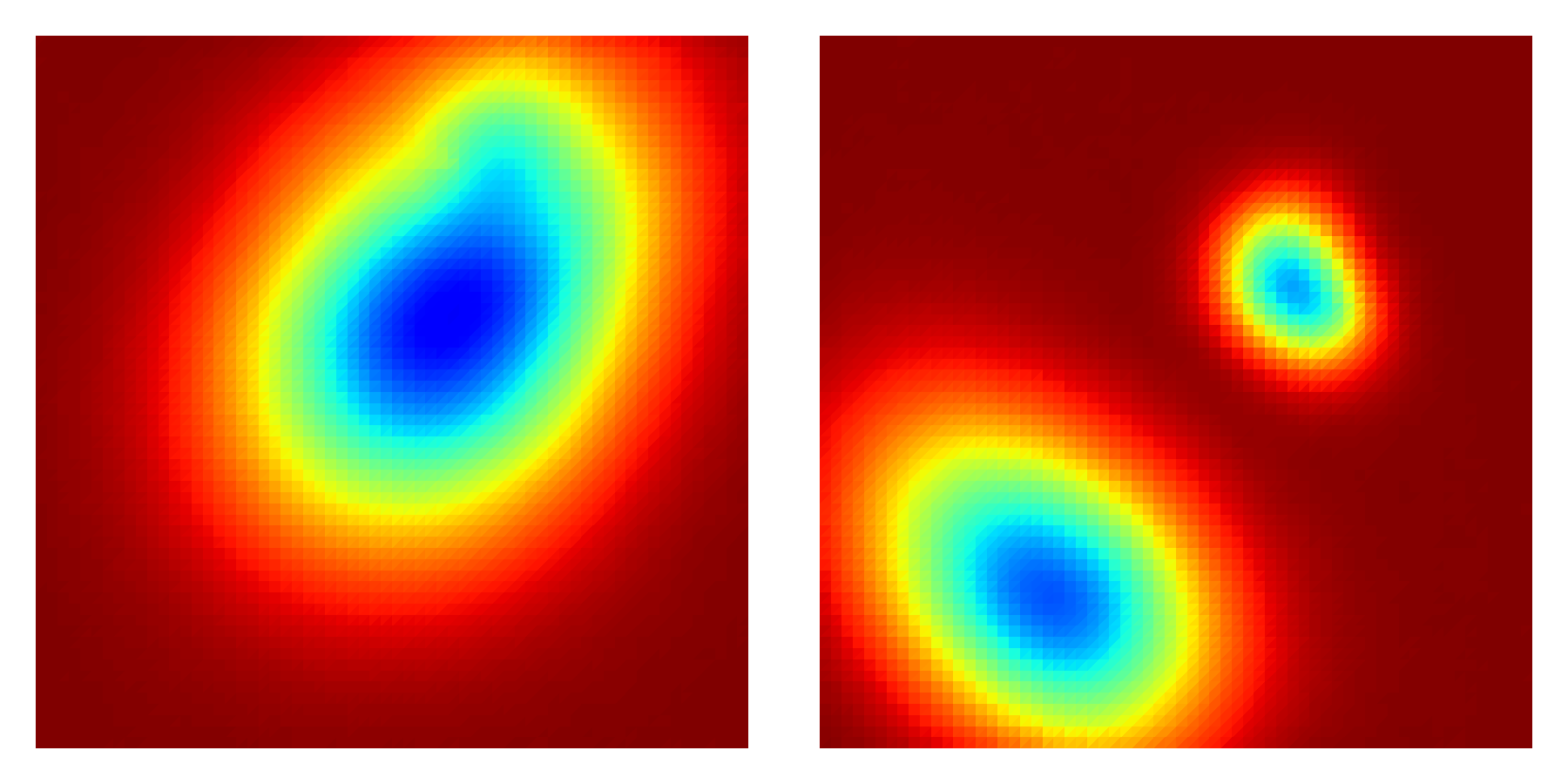}
        \caption*{GRIFDIR (ours)}
    \end{subfigure}
    \caption{
    Resolution generalisation for sampling on \GaussianBlobs sampling.  Models are trained on a coarse triangular mesh with 2048 elements (top) and evaluated on a finer mesh with 8192 elements (bottom).
    Resolution-dependent models degrade or produce artifacts at higher resolution, while resolution-invariant operators maintain spatial structure and sample quality.} %Samples are generated using 200 diffusion steps with a Heun sampler.}
    \label{fig:res_invariant}
    \imagespace
\end{figure*}

\paragraph{Resolution invariance} We train all baselines on a \num{2048}-element mesh and evaluate, without retraining, on a coarser ($1152$-element) and a finer ($8192$-element) mesh of the same square domain for \GaussianBlobs. We use a square domain with a uniform discretisation here specifically so that the CNN and FNO baselines can be included in the comparison. \Cref{tab:res_inv} reports MMD and SWD at all three resolutions, with samples shown in \cref{fig:res_invariant}. GRIFDIR achieves the best MMD and SWD at every resolution tested, including at the training resolution. The CNN and MP-GNN baselines fail in an interesting manner: when the number of mesh elements increases by a factor of four, the generated samples exhibit approximately four times as many blobs. While the performance of the FNO is quite stable across resolution, the baseline FNO can only be applied to rectangular domains with a uniform discretisation.

 %We compare ground truth, CNN, FNO, GAOT, message passing GNN, and the proposed GRIFDIR. 

\begin{table}[!ht]
\centering
\caption{MMD and SWD for \GaussianBlobs at the training resolution (2048 points) and the inference resolution (8192 points or 1152 points) computed with $1024$ samples. We report the mean MMD/SWD over the non-uniform meshes; full results are in Table~\ref{tab:discretisation_invariance}. Best values are \textbf{bold} and second best \underline{underlined}.}
\label{tab:res_inv}
\resizebox{\linewidth}{!}{
\begin{tabular}{lcccccccc}
\toprule
& \multicolumn{2}{c}{{$1152$ pts.}}
& \multicolumn{2}{c}{{$2048$ pts. (train res.)}} 
& \multicolumn{2}{c}{{$8192$ pts.}}
& \multicolumn{2}{c}{{Non-Uniform Mesh}}\\
\cmidrule(lr){2-3} \cmidrule(lr){4-5} \cmidrule(lr){6-7} \cmidrule(lr){8-9}
& MMD $(\downarrow)$ & SWD $(\downarrow)$
& MMD $(\downarrow)$ & SWD $(\downarrow)$
& MMD $(\downarrow)$ & SWD $(\downarrow)$ & MMD $(\downarrow)$ & SWD $(\downarrow)$ \\
\midrule
CNN
& \underline{0.103} & \underline{0.044} & 0.126 & 0.044 & 0.501 & 0.208 & - & - \\
FNO 
& 0.119 & 0.045 & 0.117 & 0.042 & 0.149 & 0.052 & - & - \\
% Geo-FNO
% & & & & & &\\
GAOT
& 0.169 & 0.059 & \underline{0.094} & \underline{0.034} & 0.398 & 0.163 & 0.591 & 3.438 \\
MP-GNN 
& 0.106 & 0.049 & 0.132 & 0.046 & 0.451 & 0.190 & 0.195 & 0.068 \\
MINO
& 0.133 & 0.049 & 0.109 & 0.041 & \underline{0.105} & \underline{0.040} & \textbf{0.128} & \textbf{0.046} \\
GRIFDIR
& \textbf{0.063} & \textbf{0.028} & \textbf{0.089} & \textbf{0.032} & \textbf{0.053} & \textbf{0.025} & \underline{0.160} & \underline{0.057} \\\bottomrule
\end{tabular}}
\end{table}

\paragraph{Discretisation invariance} Resolution invariance alone does not test robustness to \emph{non-uniform} node density. To test density-dependence directly, we use the \GaussianBlobs dataset on a graded 2D square mesh, with element size varying across the domain (fine left, fine centre, fine edge, fine right; see \cref{fig:graded_meshes} for visualisations of the different meshes). We evaluate the same baselines as in the resolution-invariance experiment above directly on these non-uniform meshes, without retraining. \Cref{tab:res_inv} reports the mean MMD and SWD across the four configurations. The individual results per mesh are in \Cref{tab:discretisation_invariance}.

CNN and FNO cannot be evaluated in this setting at all, since both require inputs on a regular grid. Among the remaining baselines, GAOT degrades most sharply. MP-GNN fares moderately better. MINO and GRIFDIR are the only methods that remain broadly stable under non-uniform discretisation, with MINO achieving the best mean scores (0.128, 0.046) and GRIFDIR the second-best on both metrics (0.160, 0.057).

% \begin{table}[!ht]
% \centering
% \caption{We train the models on graded meshes for the \GaussianBlobs dataset and take $1024$ samples for the evaluation. For GRIFDIR we consider three variations, based on using lumping (L) and/or area-weighting (W).}
% \resizebox{\linewidth}{!}{
% \begin{tabular}{lcccccccc}
% \toprule
% & \multicolumn{2}{c}{{Fine left}}
% & \multicolumn{2}{c}{{Fine centre}} 
% & \multicolumn{2}{c}{{Fine edge}} 
% & \multicolumn{2}{c}{{Fine right (train)}} \\
% \cmidrule(lr){2-3} \cmidrule(lr){4-5} \cmidrule(lr){6-7} \cmidrule(lr){8-9}
% & MMD $(\downarrow)$ & SWD $(\downarrow)$
% & MMD $(\downarrow)$ & SWD $(\downarrow)$
% & MMD $(\downarrow)$ & SWD $(\downarrow)$
% & MMD $(\downarrow)$ & SWD $(\downarrow)$ \\
% \midrule
% GAOT 
% & 0.730 & 4.490 & 0.687 & 22.39
% & 0.756 & 20.83 & 0.131 & 0.043\\
% MINO 
% & \textbf{0.104} & \textbf{0.038} & \textbf{0.053} & \textbf{0.026}
% & 0.156 & \underline{0.055} & \textbf{0.075} & \textbf{0.028}\\
% GRIFDIR %{\tiny(no lumped, no area-weighted loss)}
% & \underline{0.239} & \underline{0.103} & 0.251 & 0.105
% & \underline{0.127} & \textbf{0.051} & 0.104 & 0.034\\
% GRIFDIR L%{\tiny(lumped, no area-weighted loss)}
% & \underline{0.239} & 0.106 & 0.251 & 0.106
% & \textbf{0.126} & \textbf{0.051} & 0.104 & 0.034 \\
% GRIFDIR L\&W %{\tiny(lumped, area-weighted loss)}
% & 0.258 & 0.122 & \underline{0.176} & \underline{0.071}
% & 0.179 & 0.066 & \underline{0.088} & \underline{0.029} \\
% \bottomrule
% \end{tabular}}\label{tab:discretisation_invariance}
% \end{table}

% \subsubsection{Geometry generalisation}

\subsection{Domain and Geometry Generalisation}
\label{sec:geometry}
\begin{figure}[t]
    \centering
    \begin{subfigure}[t]{0.16\linewidth}
        \includegraphics[width=\linewidth]{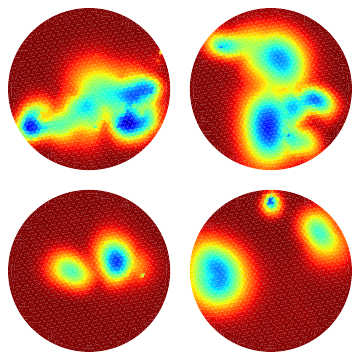}
        \caption*{Circle}
    \end{subfigure}%\hspace{-1pt}
    \begin{subfigure}[t]{0.16\linewidth}
        \includegraphics[width=\linewidth]{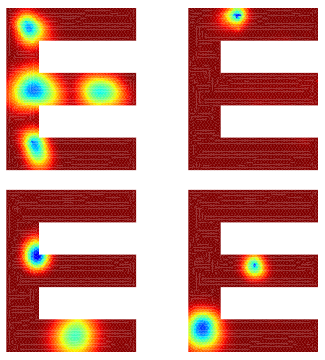}
        \caption*{E-shape}
    \end{subfigure}%\hspace{-1pt}
    \begin{subfigure}[t]{0.16\linewidth}
        \includegraphics[width=\linewidth]{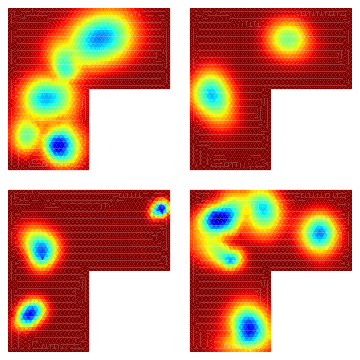}
        \caption*{L-shape}
    \end{subfigure}
    \begin{subfigure}[t]{0.16\linewidth}
        \includegraphics[width=\linewidth]{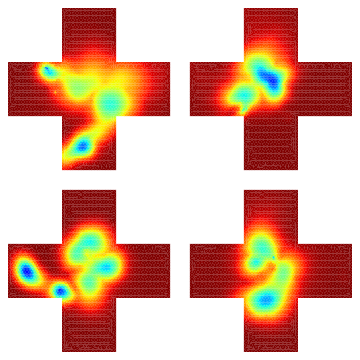}
        \caption*{Plus}
    \end{subfigure}%\hspace{-2pt}
    \begin{subfigure}[t]{0.16\linewidth}
        \includegraphics[width=\linewidth]{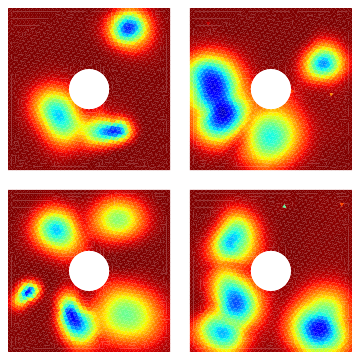}
        \caption*{Square with hole}
    \end{subfigure}%\hspace{-2pt}
    \begin{subfigure}[t]{0.16\linewidth}
        \includegraphics[width=\linewidth]{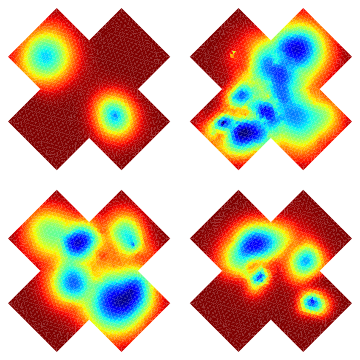}
        \caption*{X-shape}
    \end{subfigure}
    \caption{Samples of the multi domain GRIFDIR model for \GaussianBlobs.} 
    % across $6$ different domains.}
    \label{fig:different_domains}
    \imagespace
\end{figure}

% Next, we test our proposed model on its ability to train on eight different irregularly shaped domains, as described in Table~\ref{tab:gblob_datasets}, including a domain with a hole in the middle in order to verify that the model can cope with discontinuities in the domain. In \cref{fig:different_domains} we plot samples from our GRIFDIR model and see that it generates plausible samples. To embed additional domain context data 1-hot vectors are injected into the cross-attention heads as described in Appendix~\ref{app:architecture}.
Next, we test whether a single GRIFDIR model can be trained to generate samples on eight differently shaped domains (Table~\ref{tab:gblob_datasets}), including non-convex, curved and domains with holes, to verify that the model copes with a range of irregularities in the domain. One set of model weights is trained jointly on a mixed dataset of all eight domains: the shared FEM-convolutional V-cycle acts as a mesh-agnostic encoder and decoder on each domain's own mesh, and at the coarsest level cross-attention heads translate the variable-size coarse mesh into a fixed set of latent tokens shared by every domain, with a one-hot domain embedding added to the attention queries so the same latent space is shared across the geometries (Appendix~\ref{app:architecture}). \cref{fig:different_domains} shows that this single model generates plausible samples on every domain.

\subsection{Realistic and 3D Geometries}
\label{sec:realistic}

\paragraph{Pinball and Cylinder Flow}
Next, we consider 2D datasets with non-standard meshes, namely the \Cylinder and \Pinball dataset, both of which contain holes in the domain. Results for both datasets are in Table~\ref{tab:results_cylinder_pinball}. 
We see across both datasets GRIFDIR has competitive results, with the best or second best results across all metrics. Although MP-GNN does well on \Cylinder, we saw in \cref{fig:res_invariant} it doesn't generalise well to increases in resolution. MINO performs well on \Pinball, but falls behind on \Cylinder. 
We hypothesise that this is to do with the meshes: \Pinball has very uniform meshes (see \cref{fig:pinball_mesh}), whereas \Cylinder does not (see \cref{fig:grifdir_samples_cylinder}), and MINO only includes a mean aggregation over nodes in the neighbourhood.

\begin{table*}[!ht]
\centering
\caption{Results for \Cylinder and \Pinball. MMD and SWD are based on \num{10000} samples for \Cylinder and $1024$ samples for \Pinball. For \Cylinder, DPS-MSE and DPS-ES use $42$ sensors with farthest-point placement; for \Pinball, $30$ sensors with random (but fixed) placement. Best results are shown in \textbf{bold} and second-best results are \underline{underlined}.}
\label{tab:results_cylinder_pinball}

\begin{minipage}[t]{0.49\textwidth}
\centering
\Cylinder\\[2pt]
\resizebox{\textwidth}{!}{%
\begin{tabular}{lcccc}
\toprule
Method & MMD $(\downarrow)$ & SWD $(\downarrow)$ & DPS-MSE $(\downarrow)$ & DPS-ES $(\downarrow)$ \\
\midrule
MP-GNN
& \textbf{0.0187}
& \textbf{0.0115}
& $1.71\times10^{-4}$
& \underline{0.4246} \\
MINO
& 0.0307
& 0.0165
&{$\mathbf{1.11\times10^{-4}}$}
& 0.4427 \\
GRIFDIR
& \underline{0.0197}
& \underline{0.0117}
& \underline{$1.56\times10^{-4}$}
& \textbf{0.3788} \\
\bottomrule
\end{tabular}%
}
\end{minipage}
\hfill
\begin{minipage}[t]{0.49\textwidth}
\centering
\Pinball\\[3pt]
\resizebox{\textwidth}{!}{%
\begin{tabular}{lcccc}
\toprule
Method & MMD $(\downarrow)$ & SWD $(\downarrow)$ & DPS-MSE $(\downarrow)$ & DPS-ES $(\downarrow)$ \\
\midrule
Spline-Conv
& 0.3627
& 0.4416
& 0.1738
& 33.1317 \\
Geo-FNO
& 0.1686
& 0.2256
& 0.1783
& 23.3092 \\
MLP-Conv
& 0.2779
& 0.3097
& \underline{0.0778}
& {17.1494} \\
MINO
& \textbf{0.0839}
& \textbf{0.0920}
& 0.0927
& \underline{16.8855} \\
GRIFDIR
& \underline{0.0907}
& \underline{0.0954}
& \textbf{0.0721}
& \textbf{14.2480} \\
\bottomrule
\end{tabular}%
}
\end{minipage}

\end{table*}

\paragraph{Scaling to 3D}
Finally, we show that GRIFDIR scales to 3D meshes. For this, we generalise the FEM kernel to 3D via a trilinear interpolation of the kernel and evaluate on the \DeepJEB dataset, see Appendix~\ref{app:deepjeb}. We train GRIFDIR to generate stress fields across different bracket geometries (see \cref{fig:overview} for examples). The model is conditioned on the point cloud of each bracket. 

We train separate models at \num{6000} and \num{24000} points and evaluate each at its training resolution and at $\pm33\%$ of it. Unlike \GaussianBlobs and \Pinball, where every sample shares a common domain and MMD/SWD compare two populations of draws from one distribution, each \DeepJEB bracket defines its own geometry and has a single ground-truth field rather than a distribution. We therefore evaluate goodness of fit against that single reference instance, mirroring the per-instance metrics used for conditional sampling in \cref{sec:conditional_sampling} rather than the distributional metrics. In particular, we compute the relative $L^2$ error and a stress-weighted variant (w-rel-$L^2$), the Pearson correlation $r$ between predicted and true stress magnitude (evaluating whether the spatial \emph{pattern} of stress is correct), and the intersection-over-union of the predicted and true top-$q\%$ stress regions (IoU@$q$, against a chance baseline of roughly $q/2$). Training and sampling time across resolutions are in \cref{app:deepjeb}, additional samples and ground truth stress fields are shown in Figure~\ref{fig:deep_jeb_conditional}.

The results are summarised in \cref{tab:deepjeb}. At both training resolutions, GRIFDIR places roughly half of the true top-1\% stress correctly (IoU@$1\% \approx 0.52$) with $r \approx 0.77$ indicating that the overall pattern is captured well. Evaluating at a \emph{finer} resolution than training is mildly better on IoU and $r$ (e.g.\ at 6000 nodes: IoU@$1\%$: $0.523\to0.538$ and $r$: $0.764\to0.779$), whereas evaluating at a \emph{coarser} resolution degrades performance, e.g., $9$-$14\%$ for $r$. The table also includes a lumped-quadrature variant (\cref{sec:fem_conv}) at matched evaluation resolutions. Unlike the graded 2D meshes in \cref{tab:discretisation_invariance}, lumping consistently underperforms the default aggregation here on every metric, as the \DeepJEB point clouds are fairly uniform. For a single sample, GRIFDIR achieves a relative $L^2$ error of $0.64$ at both training resolutions. 

\paragraph{Comparison to supervised baselines}
For context, GINOT~\citep{liu2025ginot} reports $0.356$, GI-FNO $0.314$, and GNO $0.673$ relative error on the same JEB benchmark. However, these models are trained in a supervised way to directly minimise the relative $L^2$ error.

\begin{table}[t]
\centering
\small
\caption{\DeepJEB: the \num{6000} and \num{24000} models evaluated at $\pm 33\%$ of their training density. Mean $\pm$ std over sampling seeds; each seed is averaged over the held-out test brackets.}
\label{tab:deepjeb}
\resizebox{\textwidth}{!}{
\begin{tabular}{llcccccc}
\toprule
Train pts. & Eval pts. & rel-$L^2$ $\downarrow$ & w-rel-$L^2$ $\downarrow$ & IoU@$1\%$ $\uparrow$ & IoU@$5\%$ $\uparrow$ & IoU@$10\%$ $\uparrow$ & $r$ $\uparrow$ \\
\midrule
\num{6000} & \num{4000} & 0.698{\tiny$\pm$0.002} & 0.648{\tiny$\pm$0.002} & 0.416{\tiny$\pm$0.007} & 0.386{\tiny$\pm$0.002} & 0.446{\tiny$\pm$0.002} & 0.689{\tiny$\pm$0.002} \\
\num{6000} & \num{6000} & 0.645{\tiny$\pm$0.004} & 0.575{\tiny$\pm$0.004} & 0.523{\tiny$\pm$0.007} & 0.479{\tiny$\pm$0.004} & 0.545{\tiny$\pm$0.002} & 0.764{\tiny$\pm$0.002} \\
\num{6000} & \num{8000} & 0.695{\tiny$\pm$0.003} & 0.595{\tiny$\pm$0.004} & 0.538{\tiny$\pm$0.006} & 0.490{\tiny$\pm$0.004} & 0.554{\tiny$\pm$0.002} & 0.779{\tiny$\pm$0.002} \\
\midrule
\num{24000} & \num{16000} & 0.734{\tiny$\pm$0.002} & 0.658{\tiny$\pm$0.003} & 0.404{\tiny$\pm$0.003} & 0.381{\tiny$\pm$0.005} & 0.439{\tiny$\pm$0.002} & 0.666{\tiny$\pm$0.002} \\
\num{24000} & \num{24000} & 0.638{\tiny$\pm$0.002} & 0.564{\tiny$\pm$0.003} & 0.510{\tiny$\pm$0.005} & 0.483{\tiny$\pm$0.005} & 0.541{\tiny$\pm$0.001} & 0.771{\tiny$\pm$0.002} \\
\num{24000} & \num{32000} & 0.654{\tiny$\pm$0.003} & 0.561{\tiny$\pm$0.004} & 0.517{\tiny$\pm$0.004} & 0.493{\tiny$\pm$0.004} & 0.552{\tiny$\pm$0.002} & 0.785{\tiny$\pm$0.003} \\
\midrule
\num{6000} {\tiny(lumped)} & \num{6000} & 0.666{\tiny$\pm$0.004} & 0.593{\tiny$\pm$0.004} & 0.512{\tiny$\pm$0.004} & 0.446{\tiny$\pm$0.005} & 0.519{\tiny$\pm$0.002} & 0.750{\tiny$\pm$0.002} \\
\midrule
\num{24000} {\tiny(lumped)} & \num{24000} & 0.673{\tiny$\pm$0.002} & 0.591{\tiny$\pm$0.002} & 0.492{\tiny$\pm$0.004} & 0.430{\tiny$\pm$0.002} & 0.499{\tiny$\pm$0.002} & 0.745{\tiny$\pm$0.000} \\
\num{24000} {\tiny(lumped)} & \num{32000} & 0.730{\tiny$\pm$0.007} & 0.614{\tiny$\pm$0.007} & 0.496{\tiny$\pm$0.005} & 0.431{\tiny$\pm$0.002} & 0.498{\tiny$\pm$0.002} & 0.749{\tiny$\pm$0.002} \\
\bottomrule
\end{tabular}}
\end{table}

\section{Conclusion}
In this work, we introduced GRIFDIR, a multi-scale graph neural operator making use of FEM convolution, designed for generative modelling over irregular geometries. 
By construction, GRIFDIR is discretisation invariant and supports sampling at arbitrary resolution, making it well-suited as a backbone for score-based diffusion models defined in infinite-dimensional function spaces. 
We demonstrated the effectiveness of our approach across four datasets spanning a variety of domains, e.g., including squared, cross-shaped, and $L$-shaped geometries, consistently producing high-quality samples without retraining across resolutions.

Several directions remain for future work. 
On the conditional generation side, we plan to extend our framework to non-linear forward operators and to benchmark against learned guidance approaches such as conditional training \citep{baldassari2023conditional} and supervised guidance methods like SGT \citep{baker2026supervised}. 
Further, ablation regarding the underlying infinite-dimensional framework, such as the choice of covariance $C$ or SDE, would be valuable. 
%We also intend to broaden our comparisons to include other graph neural operator architectures that can serve as backbones within diffusion frameworks. 
%Finally, we note that comparisons to recent work in functional flow matching, for example MINO \citep{shi2025meshinformed}, which is in principle adaptable to diffusion-based modelling, are of particular interest.

\subsection*{AI use statement}
In this work, we used generative AI tools to assist in the writing of proofs (proof-checking) and to implement methods (coding some of the model components and several plotting utilities). We have not used generative AI tools to generate synthetic data sets, help develop theoretical models or conceptual frameworks, formulate mathematical claims, provide critical ingredients for proving mathematical claims, propose or refine hypotheses, design or provide feedback on research methodology or experiments, clean and reformat datasets, or support qualitative and thematic data analysis or the interpretation of results; assisting with translation is not applicable to this work.

Additionally, we used generative AI tools for brainstorming and for creating or modifying scientific figures or images, including adapting plotting functions in matplotlib and writing TikZ figure code. We have reviewed all AI-assisted work. LLM-generated code was verified by three authors, and all AI-assisted figures were reviewed and evaluated by the authors before inclusion. We take responsibility for the final content of this work, including text, claims, code, and artifacts produced with the aid of generative AI.

\section*{Acknowledgements}
AD acknowledges support from the EPSRC (EP/V026259/1). 
EB acknowledge support by funding from Villum Foundation Synergy project number 50091 entitled “Physics-aware machine learning” as well as the Center for Basic Machine Learning Research in Life Science (MLLS) through the Novo Nordisk Foundation (NNF20OC0062606).
JR acknowledges support from the EU Horizon MSCA-SE under project REMODEL, “Research Exchanges in the Mathematics of Deep Learning with Applications” (grant agreement no. 101131557).
BA acknowledges support from the Natural Sciences and Engineering Research Council of Canada (NSERC) through grant RGPIN/2470-2021 and FRQ (Fonds de recherche du Quebec) -- Nature et Technologies through grant 359708. 
This research was supported in part through the Maxwell computational resources operated at Deutsches Elektronen-Synchrotron DESY, Hamburg, Germany.

\bibliographystyle{iclr2027_conference}
\bibliography{bibliography}

%%%%%%%%%%%%%%%%%%%%%%%%%%%%%%%%%%%%%%%%%%%%%%%%%%%%%%%%%%%%

\newpage
\appendix

\section{Related Work}
\label{app:realted_work}
\paragraph{Graph Diffusion} 
It is important to distinguish the infinite-dimensional framework from recent works that apply GNNs within standard finite-dimensional diffusion models. 
For instance, \citet{valencia2025learning} propose \emph{Graph Diffusion Models} to learn the conditional distribution of a fluid's equilibrium state given a short simulation rollout. 
While this approach uses GNNs for diffusion, it operates within the finite-dimensional framework on discrete states corrupted by i.i.d.\ Gaussian noise.
By contrast, our infinite-dimensional formulation ensures the diffusion model operates in a resolution-invariant manner by design.

\paragraph{Functional Flow Matching}
Functional Flow Matching (FFM)~\citep{kerrigan2023functional} is a related framework to infinite-dimensional diffusion. FFM generalises flow matching to operate in infinite-dimensional function spaces. Rather than learning a vector field on a finite-dimensional latent, FFM defines a path of probability measures interpolating between a Gaussian measure (often $\mathcal{N}(0,C)$) and the data distribution, and learns a function-space vector field that generates this path. As in our setting, extending such generative models to function spaces requires discretisation-invariant network architectures capable of operating on irregular domains. This need is addressed by MINO~\citep{shi2025meshinformed}, which makes use of graph neural operators and cross-attention mechanisms. Our work shares this motivation: the infinite-dimensional diffusion model we develop similarly requires mesh-aware architectures to handle unstructured spatial domains, and the resulting model is likewise discretisation-invariant by design.

% \paragraph{SplineCNN \cite{fey2018splinecnn}}
% By contrast, SplineCNN  algebraically the closest comparator, since the Q1 case of our basis recovers their tensor-product B-spline at degree one evaluates its basis in mesh-derived pseudo-coordinates and aggregates over $1$-hop graph neighbours, so neither the filter's spatial semantics nor its physical receptive field survive mesh refinement.

\section{Background}

\subsection{The story in finite-dimensions}
\label{app:finite_dim_story}
One way of using diffusion models for function space data, is first by discretising the data, and then applying the finite-dimensional formulation. For this, suppose we have a data distribution $p_{\data}$ with support in state space $\rbb^d$. In the case of function space data, $p_\data$ could, for instance, be a distribution of 2D magnetic fields discretised on a uniform grid. 

To learn to generate samples from $p_\data$ we follow the score-based diffusion model setup \citep{song2021scorebased}. We consider a forward stochastic differential equation (SDE) that noises data samples, such that for large $T$ the marginal distribution is approximately Gaussian:
\begin{align}
    \drm x_t = f(x_t,t) \drm t + g(t) \drm W_t, \quad x_0 \sim p_\text{data}(x),
\end{align}
where $f:\R^d \times [0,T] \to \R^d$ and $g:[0,T] \to \R_{> 0}$ are the drift and diffusion coefficients, chosen such that $p_T \approx \mathcal{N}(0, I)$.
By simulating trajectories of the time-reversal of this SDE, run backwards in time from $x_T \sim p_T \approx \mathcal{N}(0,I)$, we obtain samples from $p_\data$ at time $t=0$:
\begin{align}\label{eq:time_reversal}
\begin{split}
    \drm x_t &= \hat{f}(x_t, t) \drm t + g(t) \drm W_t,
    \\
    \hat{f}(x, t)&\coloneqq [f(x, t) - g(t)^2 \nabla_x \log p_t(x)].
    \end{split}
\end{align}
The reverse SDE depends on the unknown score function which we approximate with a neural network $s_\theta(x,t) \approx \nabla_x \log p_t(x)$ trained via denoising score matching.

Now consider an inverse problem of the form
\begin{align}
    y = Ax + \eta, \quad x \sim p_\text{data}(x),
\end{align}
where $A$ is a (possibly non-linear) forward operator and $\eta$ is Gaussian observation noise. Then, instead of sampling from the prior distribution $p_\data$, we wish to sample from the posterior, $p_\data(\cdot \mid y).$ 
This can be achieved by replacing the reverse SDE with a conditional reverse SDE:
\begin{align}
\begin{split}
    \drm x_t &=  \tilde{f}(x_t, t) \drm t + g(t) \drm W_t,
    \\
    \tilde{f}(x_t, t) &\coloneqq [f(x, t) - g(t)^2 \nabla_x \log p_t(x\mid y)]
    \end{split}
\end{align}
which relies on the conditional score $\nabla_x \log p_t(x\mid y)$. Using Bayes' theorem we can decompose the conditional score as
\begin{align}
    \nabla_x \log p_t(x\mid y) &= \nabla_x \log p_t(x) +\nabla_x \log p_t(y\mid x),
\end{align}
where the first term is approximated by the learned unconditional score and  $\nabla_x \log p_t(y\mid x)$ is the guidance term, a noisy version of the data likelihood $p_\text{lkhd}(y \mid x_0)$. The guidance term is intractable as can be seen by writing the guidance term as a conditional expectation
\begin{align}
    p_t(y\mid x_t) &= \int p_\text{lkhd}(y\mid x_0) p(x_0 \mid  x_t) dx_0 
    \\
    &= \mathbb{E}[p_\text{lkhd}(y\mid x_0) \mid  x_t].
\end{align}
Therefore, to sample from the posterior, we either need to approximate this guidance term, or a way to learn the conditional score directly.

\subsection{Guidance for Conditional Sampling}\label{app:bg_guidance}
In order to sample from the conditional distribution $\pi^y$, the fundamental step is to formulate or approximate a conditional score $s^y(t, a)$. We detail several paradigms for achieving this below.

\paragraph{Conditional Training} 
For the conditional sampling problem, outlined in \cref{sec:conditional_sampling}, there are multiple ways to sample from $\pi^y$. One of such is learning the conditional score 
\begin{align}\label{eq:app_conditional_score}
    s^y(t, a) = -v_t^{-1}\left(a - m_t\ebb[A_0 \mid A_t=a, Y=y]\right),
\end{align},
where $v_t$ and $m_t$ denote the variance and mean coefficients of the forward transition, i.e. $\text{Law}(A_t \mid A_0) = \mathcal{N}(m_t A_0, v_t C)$,
as outlined by \citet{baldassari2023conditional}. Using the time-reversal with the unconditional score as defined in \eqref{eq:backwardSDE} leads to samples from the distribution $\pi$. However, \citet{baldassari2023conditional} show that replacing the unconditional score by the conditional one in \eqref{eq:app_conditional_score}, the time-reversal leads to samples from $\pi^y$.
The conditional score is learned directly (as opposed to relying on knowledge of the unconditional score), using the loss
\begin{align}
   \ebb_{(a_0, y)\sim \text{Law}(A_0, Y) a_t\sim \text{Law}(A_t \mid A_0 =a_0)}\|-(1-e^{-t})^{-1}(a_t -e^{-\frac{t}{2}}a_0 - s_{\theta}(t, a_t, y)\|^2_{\mathcal{H}},
\end{align}
where $\mathcal{H}$ is the Hilbert space to which the data $a$ belongs. \citet{baldassari2023conditional} show that if the conditional expectation of the score function is bounded, then the above loss is equivalent to the intractable loss function 
\begin{align}
       \ebb_{(a_t, y)\sim \text{Law}(A_t, Y)}
       \|s^y(t, a_t) - s_{\theta}(t, a_t, y)\|^2_{\mathcal{H}}.
\end{align}

\paragraph{Joint Embedding Models} 
Rather than explicitly computing the guidance term \eqref{eq:guidance_inf}, this approach circumvents it by learning a joint score. 
Using paired data $(a_0,y)$, an unconditional diffusion model is trained to estimate $\pi_\text{joint}(a_0, y)$. 
During training, the observed data $y$ is treated as a fixed mask. 
By marginalising over this joint measure conditioned on this mask, one samples from the posterior $\pi^y$ \citep{yao_guided_2025,huang2024diffusionpde}. 
This approach requires training a new joint model for every new forward operator $\mathcal{L}$.

\paragraph{Diffusion Priors and Guidance: (Fun-)DPS}
Another approach relies on training an unconditional diffusion prior solely on the measure $\pi$ of the parameter $a_0$, and then approximating the intractable guidance term \eqref{eq:guidance_inf}.
For instance, \citet{lin2026decoupled} utilise a training-free approximation. 
At each denoising step, they use the unconditional score to compute the Tweedie estimate $\hat{a}_t(a) = \ebb[A_0\mid A_t=a]$ and approximate the guidance via 
\begin{align}\label{eq:approx_guidance_tweedie}
\log h^u(t, a) \approx -\Phi(\hat{a}_t(a), u).
\end{align}
This estimate is then used to correct the unconditional trajectory toward the conditional distribution. This mechanism is closely related to posterior samplers like DPS \citep{chung2022diffusion} in finite dimensions and FunDPS \citep{yao_guided_2025} in infinite dimensions. In its full form, FunDPS combines a joint embedding model with this Tweedie-based approximation, though in our setting we can replace the learned joint model with a standalone diffusion prior and solely adopt the approximated guidance term \eqref{eq:approx_guidance_tweedie}.

% \paragraph{(Fun-)DPS (heuristic guidance)}
% Instead of learning the conditional score directly, another approach is to first train a model to learn a prior distribution, and then obtain samples through a posterior sampler, as is the approach taken by \citet{chung2022diffusion} in finite dimensions and \citet{yao_guided_2025} in infinite dimensions. In its full form FunDPS combines (i) a joint embedding model trained on paired data $(a_0, y)$, which effectively learns an unconditional prior over a masked joint space, and (ii) a guidance term obtained via a Tweedie-based approximation of the conditional score. While the original formulation learns this joint embedding explicitly, in our setting we adopt only the second component: we approximate the guidance term using \eqref{eq:approx_guidance_tweedie}, but replace the learned joint model with a standalone diffusion prior. 

\paragraph{DAPS} Decoupled Annealing Posterior sampling (DAPS) 
\citep{zhang2025improving, lin2026decoupled} is a posterior sampling method that decouples the denoising step from the data-consistency update. In contrast to (Fun-)DPS, DAPS avoids backpropagation through the score model and instead performs Langevin update steps in the denoised space. We adapt DAPS to the function space setting, leading to the Function-Space DAPS (Fun-DAPS) algorithm, shown in Algorithm~\ref{alg:fs_daps}. 
Similar to the finite-dimensional setting, the conditional distribution $A_0 | A_t=a_t$ is generally intractable.
Following \citet{zhang2025improving}, we use a Gaussian approximation 
\begin{align}
    A_0 | A_t=a_t \sim \mathcal{N}(\hat{a}_0(a_t), r_t^2 C),
\end{align}
where $C$ is the prior covariance operator and $\hat{a}_0(a_t)$ denotes a Tweedie estimate computed from the score model. Assume measurements are generated according to
\begin{align}
y = \mathcal{L}(a) + \xi, \qquad \xi \sim \mathcal{N}(0, \sigma_\xi^2 I),
\end{align}
with $\mathcal{L}: \mathcal{H} \to \mathbb{R}^m$ (e.g., finite-dimensional observations), which results in a likelihood potential $\Phi(a) = \frac{1}{2\sigma_\xi^2}\|\mathcal{L}(a) - y\|^2$.
The function-space (preconditioned) Langevin dynamics then take the form
\begin{align}
\hat{a}_0^{(j+1)} 
&\leftarrow 
\hat{a}_0^{(j)}
- \eta_t C
\left[
\frac{1}{r_t^2}
C^{-1}
\left(
\hat{a}_0^{(j)} - \hat{a}_0^{(0)}
\right)
+
\nabla \Phi(\hat{a}_0^{(j)})
\right]
+
\sqrt{2\eta_t}\,\epsilon_j,
\quad
\epsilon_j \sim \mathcal{N}(0,C),
\end{align}
with a decaying step size $\eta_t$.

% \begin{algorithm}[t]
% \caption{Decoupled Annealing Posterior Sampling (DAPS)}
% \label{alg:daps}
% \begin{algorithmic}[1]
% \REQUIRE Score model $s_\theta$, measurement $y$, noise schedule $\sigma_t$, $\{t_i\}_{i=0}^{N_A}$
% \STATE Sample $x_T \sim \mathcal{N}(0, \sigma_T^2 I)$
% \FOR{$i = N_A, N_A-1, \dots, 1$}
%     \STATE Compute $\hat{x}_0^{(0)} = \hat{x}_0(x_{t_i})$ by solving the probability flow ODE with $s_\theta$ (or a one step Tweedie estimate)
%     \FOR{$j = 0, \dots, N-1$}
%         \STATE \textbf{Langevin dynamics:}
%         \STATE
%         \[
%         \hat{x}_0^{(j+1)} \leftarrow \hat{x}_0^{(j)} + \eta_t \Big(
%         \nabla_{\hat{x}_0} \log p(\hat{x}_0^{(j)} \mid x_{t_i})
%         + \nabla_{\hat{x}_0} \log p(y \mid \hat{x}_0^{(j)})
%         \Big)
%         + \sqrt{2\eta_t}\,\epsilon_j,
%         \quad \epsilon_j \sim \mathcal{N}(0, I)
%         \]
%     \ENDFOR
%     \STATE Sample $x_{t_{i-1}} \sim \mathcal{N}(\hat{x}_0^{(N)}, \sigma_{t_{i-1}}^2 I)$
% \ENDFOR
% \STATE \textbf{Return:} $x_0$
% \end{algorithmic}
% \end{algorithm}

\begin{algorithm}[t]
\caption{Function-Space Decoupled Annealing Posterior Sampling (Fun-DAPS)}
\label{alg:fs_daps}
\begin{algorithmic}[1]
\REQUIRE Score operator $s_\theta$, measurements $y$, covariance operator $C$, noise schedule $\sigma_t$, $\{t_i\}_{i=0}^{N_A}$, likelihood potential $\Phi(\mathbf{x})$, $r_t=\sigma_t$, step size $\eta_t$
\STATE Sample $a_T \sim \mathcal{N}(0, \sigma_T^2 C)$
\FOR{$i = N_A, N_A-1, \dots, 1$}
    \STATE \textit{Tweedie Estimate} (or e.g. probability flow)
    \STATE $\hat{a}_0^{(0)} = a_{t_i} + \sigma_{t_i}^2 C\, s_\theta(a_{t_i}, t_i)$
    
    \FOR{$j = 0, \dots, N-1$}
        \STATE \textit{Function-space (preconditioned) Langevin dynamics}
        \STATE
        \[
        \hat{a}_0^{(j+1)}
        \leftarrow
        \hat{a}_0^{(j)}
        -
        \eta_t C
        \left[
        \frac{1}{r_{t_i}^2}
        C^{-1}
        \left(
        \hat{a}_0^{(j)} - \hat{a}_0^{(0)}
        \right)
        +
        \nabla \Phi(\hat{a}_0^{(j)})
        \right]
        +
        \sqrt{2\eta_t}\, \xi_j, \quad \xi_j \sim \mathcal{N}(0, C)
        \]
    \ENDFOR
    \STATE \textit{Annealed Gaussian update}
    \STATE
    \[
    a_{t_{i-1}}
    \sim
    \mathcal{N}
    \left(
    \hat{a}_0^{(N)},
    \sigma_{t_{i-1}}^2 C
    \right)
    \]
\ENDFOR
\STATE \textbf{Return:} $a_0$
\end{algorithmic}
\end{algorithm}

\paragraph{Trained Guidance (DEFT/SGT)}
In finite dimensions \citet{denker2024deft} propose a training method, DEFT, to learn the guidance term $\nabla_a \log p_t(y \mid a).$ This is extended into infinite-dimensions by \citet{baker2026supervised} who introduce SGT to learn the infinite-dimensional guidance term $\nabla_a \log h^y(t, a) \coloneqq \nabla \log \xi^{-1}\ebb_{\pbb}\!\left[\exp(-\Phi(A_0, y))\mid A_t = a\right]$ as defined in \eqref{eq:guidance_inf}. Here, we detail the infinite-dimensional counterpart SGT, though the finite-dimensional DEFT works similarly. In order to learn the guidance term, \citet{baker2026supervised} work with the variance preserving SDE \eqref{eq:backwardSDE} and define the loss function
\begin{align}
    \label{eq:deft_loss}
    \begin{split}
       L(\phi) \coloneqq \mathbb{E}\left[\left\| \frac{A_t - e^{-\frac{t}{2}}A_0}{1-e^{-t}} +s(t,A_t) +u_{\phi}(t,A_t,Y) \right\|_K^2\right],
\end{split}
\end{align}
where the expectation is taken over $t\sim \mathcal{U}[0, T],$ $A_0, Y \sim \text{Law}(A_0, Y),$ $A_t \sim \text{Law}(A_t\mid A_0)$, where $K$ is an appropriately chosen Hilbert space that depends on the regularity of the data distribution. Then for the minimiser ${\phi^*}$ of $L(\phi)$, it holds that $u_{\phi^*} = C \nabla \log h^y(t, a).$ 

% \begin{algorithm}[t]
% \caption{Supervised Guidance Training (SGT)}
% \label{alg:guidance_training}
% \begin{algorithmic}[1]
% \REQUIRE Covariance operator $C$, pre-trained score model $s_\theta(t,z)$, loss weighting, training data $\{ X^{(i)}, Y^{(i)}\}_{i=1}^N$, batch size $B$, guidance function $u_\phi$
% \WHILE{Metrics not good enough}
% \STATE Subsample $\{ x^{(i)}_0, y^{(i)} \}_{i=1}^B$ from $\{ X^{(i)}, Y^{(i)}\}_{i=1}^N$
% \STATE $\epsilon^{(i)} \sim \mathcal{N}(0,I)$ 
% \STATE $t_i \sim \mathcal{U}([0,T])$
% \STATE $\sigma_{t_i} = \sqrt{1 - e^{-t_i}}$
% \STATE $x_t^{(i)} = e^{-\frac{1}{2} t_i} x_0^{(i)} + \sigma_{t_i} C^{\frac{1}{2}} \epsilon^{(i)}$ 
% \STATE $\hat{s}_\theta^{(i)} = \colorbox{blue!10}{$\texttt{stopgrad}(s_\theta(x_t^{(i)}, t_i))$}$ \label{alg:line_stopgrad}
% %\COMMENT{evaluate uncond. score model} 
% \STATE $\hat{u}_\phi^{(i)} = u_\phi(x_t^{(i)}, y^{(i)},t_i)$
% \STATE $r^{(i)} = (\hat{s}_\theta^{(i)} + \hat{u}_\phi^{(i)}) + \sigma_{t_i}^{-1}\, C^{\frac{1}{2}}\epsilon^{(i)}$ \label{alg:residual}
% \STATE $L(\phi)\!= \sum_{i=1}^B \| C^{-\frac{1}{2}} r^{(i)} \|_2^2$ \COMMENT{evaluate loss \eqref{eq:deft_loss}}
% \STATE Gradient step with $\nabla_\phi L(\phi)$
% \ENDWHILE
% \end{algorithmic}
% \end{algorithm}

%%%%%%%%%%%%%%%%%%%%%%%%%%%%%%%%%%%%%%%%%%%%%%%%%%%%%%%%%%%%%%%%%%%%%%%%%%%%%%%
%%%%%%%%%%%%%%%%%%%%%%%%%%%%%%%%%%%%%%%%%%%%%%%%%%%%%%%%%%%%%%%%%%%%%%%%%%%%%%%

%%%%%%%%%%%%%%%%%%%%%%%%%%%%%%%%%%%%%%%%%%%%%%%%%%%%%

\section{Proofs}\label{app:proofs}

\begin{proposition}[Generalisation of CNNs]
\label{prop:gen_cnns}
Let $\Omega \subset \mathbb{R}^2$. Let the mesh be the uniform Cartesian grid $\{x_i+h(m,n):m,n\in\mathbb Z\}$ with the bilinear basis, so $\operatorname{supp}\psi_j = \{y : \|y-x_j\|_\infty \le h\}$. Let $P$ be odd and set the FEM patch width to $r=\tfrac{(P-1)h}{2}$. Then, we have
\[
(\mathcal K^{\mathrm{FEM\text{-}Conv}}a)(x_i) = h^2\!\!\sum_{m,n=-(P-1)/2}^{(P-1)/2}\!\! w_{p(m,n)}\;a\bigl(x_i+h(m,n)\bigr),
\]
i.e.\ $\mathcal K^{\mathrm{FEM\text{-}Conv}}$ reduces \emph{exactly} to a standard discrete $P\times P$ cross-correlation with kernel $\{w_p\}_{p=1}^{P^2}$, up to the constant rescaling $h^2$.
\end{proposition}

\begin{proof}
For the uniform grid the neighbours are given by $\mathcal N(i):=\{x_i+h(m,n):m,n\in\{-\tfrac{P-1}2,\ldots,\tfrac{P-1}2\}\}$. For every neighbour $x_j \in \mathcal{N}(i)$ we have that 
\begin{align*}
    \Phi_{x_i}(x_j) = \frac{h(m,n)}{r} = \frac{2(m,n)}{P-1} =: \xi_{p(m,n)}
\end{align*}
is exactly a reference grid point, i.e., $(m,n)\mapsto\xi_{p(m,n)}$ a bijection between $\mathcal N(i)$ and the $P^2$ points $\{\xi_p\}$ of the reference patch. Conversely, if $x_j\notin\mathcal{N}(i)$, then $\|x_j-x_i\|_\infty>r$, so $x_j\notin B_r(x_i)$. 
Then for every $x_j \in \mathcal{N}(i)$, and using $\phi_p(\xi_q) = \delta_{pq}$, we have 
\begin{align*}
\kappa\bigl(\Phi_{x_i}(x_j)\bigr) = \sum_{p=1}^{P^2} w_p\,\phi_p\bigl(\xi_{p(m,n)}\bigr) = w_{p(m,n)}.
\end{align*}
For the uniform cartesian grid we have $V_{i,j}=h^2$ on $\mathcal N(i)$ and as such 
\begin{align*}
  (\mathcal K^{\mathrm{FEM\text{-}Conv}}a)(x_i) = \sum_{x_j\in\mathcal N(i)} h^2\,w_{p(m,n)}\,a(x_j) = h^2\!\!\sum_{m,n=-(P-1)/2}^{(P-1)/2}\!\! w_{p(m,n)}\,a\bigl(x_i+h(m,n)\bigr),  
\end{align*}
which is exactly a discrete $P\times P$ convolution with kernel $\{w_p\}$, up to the constant $h^2$.
\end{proof}

\begin{proposition}[Mesh-refinement consistency]
\label{prop:mesh_refinement_consistency}
Let $\Omega$ be compact. For a fixed kernel $\kappa(\xi) = \sum_{p=1}^{P^2} w_p\,\phi_p(\xi)$ and $a \in C(B_r(x_i))$ we have
\begin{align}
    (\mathcal{K}^{\mathrm{FEM\text{-}Conv}}a)(x_i) \;\longrightarrow\; (\mathcal{K}a)(x_i) = \int_{B_r(x_i)} \kappa\bigl(\Phi_{x_i}(y)\bigr)\,a(y)\,\mathrm{d}y
\end{align}
as $h_n \to 0$.
\end{proposition}

\begin{proof}
Let $f(y) := \kappa(\Phi_{x_i}(y))\,a(y)$. Since $\kappa$ is a finite sum of continuous basis functions, it is continuous on $\hat\Omega$, $\Phi_{x_i}$ is continuous, and $a$ is continuous by assumption. So, $f$ is continuous. Further, $f$ is uniformly continuous as $B_r(x_i)$ is compact. 

Using $\sum_{j\in N(i)}\psi_j\equiv1$ on $B_r(x_i)$  we can rewrite the integral 
\begin{align*}
    \int_{B_r(x_i)} f(y)\,dy = \sum_{j\in N(i)} \int_{B_r(x_i)} \psi_j(y)\,f(y)\,dy.
\end{align*}
With this we can bound 
\begin{align*}
\Bigl|(\mathcal{K}^{\mathrm{FEM\text{-}Conv}}a)(x_i) - (\mathcal{K}a)(x_i)\Bigr|
&= \Bigl|\sum_{j\in N(i)} V_{i,j}\,f(x_j) - \int_{B_r(x_i)} f(y)\,dy\Bigr| \\
&= \Bigl|\sum_{j\in N(i)} \int_{B_r(x_i)} \psi_j(y)\bigl(f(x_j)-f(y)\bigr)\,dy\Bigr| \\
&\le \sum_{j\in N(i)} \int_{B_r(x_i)} \psi_j(y)\,\bigl|f(x_j)-f(y)\bigr|\,dy \\
\end{align*}
Fix $\varepsilon>0$. By uniform continuity of $f$, choose $\delta>0$ such that $|y-y'|<\delta \Rightarrow |f(y)-f(y')|<\varepsilon$. If $h_n<\delta$, then every $y\in\operatorname{supp}\psi_j$ satisfies $|y-x_j|\le h_n<\delta$ for $j\in N(i)$. 
We then get 
\begin{align*}
\Bigl|(\mathcal{K}^{\mathrm{FEM\text{-}Conv}}a)(x_i) - (\mathcal{K}a)(x_i)\Bigr| \le \varepsilon \sum_{j\in N(i)} \int_{B_r(x_i)} \psi_j(y)\,dy = \varepsilon\,|B_r(x_i)|,
\end{align*}
using the identity $\sum_{j\in N(i)} V_{i,j} = |B_r(x_i)|$ from before. As $\varepsilon>0$ was arbitrary, $(\mathcal{K}^{\mathrm{FEM\text{-}Conv}}a)(x_i) \to (\mathcal{K}a)(x_i)$ as $h_n\to0$.
\end{proof}

% \begin{proposition}[Kernel Consistency]
% \label{prop:kernel_consistency}
% {\color{red}TODO}
% Let $w_p=\kappa^\ast(\xi_p)$ be the kernel values at neighbour offsets $\xi_p$. For any continuous kernel $\kappa^\ast$, and let $h_n$ be the largest dual-cell diameter meeting $B_r(x_i)$. Then
% \begin{align}
%    \Bigl|(\mathcal{K}^{\mathrm{FEM-Conv}}a)(x_i)-\int_{B_r(x_i)}\kappa^\ast(\Phi_{x_i}(y-x_i))\,a(y)\,dy\Bigr| \le C_1\bigl(\omega_{f^\ast}(h_n)+h_n\bigr)+C_2\,\omega_{\kappa^\ast}\bigl(\tfrac{2\sqrt{2}}{P-1}\bigr) 
% \end{align}
% so the lumped FEM convolution converges to the true continuous convolution as $h_n\to0$ and $P\to\infty$, jointly and in either order ($O(P^{-2})$ for $\kappa^\ast\in C^2$).
% \end{proposition}

% \begin{proof}
%     TODO
% \end{proof}

\section{GRIFDIR Architecture Details}\label{app:architecture}

We parametrise the score network $s_\theta(t, a)$ as a multi-resolution graph U-Net whose intra-level operator is the FEM-basis convolution of Equation~\eqref{eq:fem_message}. We describe the forward pass; encoder, V-cycle across $L$ mesh resolutions, global latent transformer at the coarsest level, and decoder and embedded supporting components (FiLM timestep conditioning, learned restriction/prolongation, position re-injection, multi-domain heads).

\paragraph{Mesh hierarchy.}
Meshes are built once per domain and cached. For square conductivity ($32\times32$ DG-$0$ cells) we use a uniform refinement chain $32\!\to\!16\!\to\!8\!\to\!4$ and store sparse pool/unpool matrices per level. For unstructured domains (\ref{app:gaussian_blobs}-L-shape, circle, plus and \ref{app:pinball} Pinball), Gmsh
produces independent triangulations at decreasing $h_{\max}$ and the pool/unpool maps are built by k-NN matching of node positions. %Per-level node counts are summarised in Table~\ref{tab:mesh_stats}.

\paragraph{Per-node input encoding.}
Each node $i$ of the finest mesh carries a noisy observation $\mathbf{a}_{t,i}\in\mathbb{R}^C$ and a physical coordinate $x_i\in\mathbb{R}^d$. The diffusion timestep $t\in[0,T]$ is mapped to a $d_t$-dimensional random Fourier-feature embedding $\gamma(t) = [\sin(2\pi\boldsymbol{\omega}t);\,\cos(2\pi\boldsymbol{\omega}t)]$ with $\boldsymbol{\omega}\sim\mathcal{N}(0,\sigma^2 I)$. A two-layer MLP with SiLU activations lifts the per-node vector $[\mathbf{a}_{t,i};\,x_i;\,\gamma(t)]$ to the hidden dimension $H$, producing the level-$0$ representation $h^{(0)}\in\mathbb{R}^{N_0\times H}$. The same embedding $\gamma(t)$ is reused later as the conditioning input to per-level FiLM modulation in the V-cycle and to AdaLN-Zero conditioning in the latent transformer.

\paragraph{Multi-resolution V-cycle.}
At each level $\ell\in\{0,\dots,L{-}1\}$, $K$ FEM-convolutional layers (Section~\ref{sec:fem_conv} update node features in place. Between levels, learned restriction $\mathcal{R}^{(\ell)}$ and prolongation $\mathcal{P}^{(\ell)}$ operators move features fine-to-coarse and coarse-to-fine. Restriction averages each coarse node's pre-image and applies a small MLP,
\begin{align*}
    h_c^{(\ell+1)} = \mathrm{MLP}\Bigl(\tfrac{1}{|\mathcal{R}^{-1}(c)|}
        \sum_{i\in\mathcal{R}^{-1}(c)} h_i^{(\ell)}\Bigr).
\end{align*}
Prolongation broadcasts each coarse feature back to its fine pre-image and combines it with the down-pass skip through a residual MLP,
\begin{align*}
    h_i^{(\ell)} = h_{\mathcal{P}(i)}^{(\ell+1)}
        + \mathrm{MLP}\bigl([h_{\mathcal{P}(i)}^{(\ell+1)};\,
        h_{\mathrm{skip},i}^{(\ell)}]\bigr).
\end{align*}
After every FEM-convolutional layer we apply zero-initialised FiLM modulation~\citep{perez2018film},
$h \leftarrow h\odot(1+\boldsymbol{\alpha}^{(\ell)})+\boldsymbol{\beta}^{(\ell)}$
with $(\boldsymbol{\alpha}^{(\ell)},\boldsymbol{\beta}^{(\ell)})
= \mathrm{MLP}_\ell(\gamma(t))$, so the diffusion timestep modulates each level's features and the network starts from an identity mapping. To keep the FEM filter coordinate-aware after each restriction we additionally re-inject the level's physical coordinates via a per-level linear projection added to the hidden features.

\paragraph{Cross-attention heads to the latent space.}
Between the V-cycle bottom and the global latent transformer, the variable-size coarsest-mesh feature tensor $h^{(L-1)} \in\mathbb{R}^{N_{L-1}\times H}$ is mapped to a fixed $M$-token latent through a cross-attention head whose queries are $M$ learnable tokens that cross-attend to $h^{(L-1)}$. A symmetric decoder head maps the processed latent back to the coarsest mesh, coarse-mesh skip features cross-attend to the $M$-token latent, where the result is recombined with the V-cycle skip and propagated up. Both heads share their weights
across all input meshes, so any coarsest-mesh discretisation is mapped into a single shared latent space of $M$ pseudo-coarse tokens. For multi-domain training across $D$ shapes the queries on both heads additionally receive a domain conditioning bias, an MLP projection of the one-hot domain identity $d \in \{0,1\}^D$ added to the queries so the same shared latent is reached from any shape.

\paragraph{Global latent transformer.}
At the coarsest level $\ell=L{-}1$ the $N_{L-1}$ tokens are processed by $K_T$ DiT blocks~\citep{peebles2023dit} with AdaLN-Zero conditioning, capturing long-range dependencies that local message passing cannot resolve:
\begin{align*}
    h &\leftarrow h + \boldsymbol{\gamma}_1\odot
        \mathrm{MHSA}\bigl(\mathrm{AdaLN}(h;\,\gamma(t))\bigr), \\
    h &\leftarrow h + \boldsymbol{\gamma}_2\odot
        \mathrm{FFN}\bigl(\mathrm{AdaLN}(h;\,\gamma(t))\bigr),
\end{align*}
with output gates $\boldsymbol{\gamma}_1,\boldsymbol{\gamma}_2$ zero-initialised. Coarse-mesh coordinates are encoded by a small MLP and added as positional embeddings before the first block. The decoder mirrors the encoder, mapping $\mathbb{R}^H$ back to the $C$-channel signal at the finest mesh.

\paragraph{FEM filter family.}
Equation~\eqref{eq:fem_message} parametrises each message through a $P\times P$ filter $\{w_p\}_{p=1}^{P^2}$ evaluated on a fixed reference patch. We use the bilinear nodal $P_1$ basis: $P^2$ filter weights at a uniform reference grid, where each sample contributes to its four surrounding weights via bilinear interpolation summing to one. The basis returns $\mathbf{0}$ outside the patch, giving the layer a strict physical receptive field. The filter coefficients combine with source-node features through a per-edge vector gate $m_{ij} = (W_g b_{ij}) \odot (W_m h_j)$ with $W_g \in \mathbb{R}^{H \times P^2}$.

\subsection{Multi-Channel FEM Convolution}
\label{app:fem_conv_mult_channels}
We now consider the case of a signal $a: \Omega \to \mathbb{R}^C$ with $C$ channels and we want to map to $C'$ output channels. We denote $a^{(c)}$ as the $c$-th component of $a$. We write the filter for the $(c', c)$ channel pair as $\kappa^{(c',c)}(\xi) =
\sum_p w_p^{(c',c)}\,\phi_p(\xi)$, one $P{\times}P$ filter image per pair, directly
analogous to a standard CNN. The output at channel $c'$ is
\begin{align}
\label{eq:app_fem_message}
(\mathcal{K}a)^{(c')}(x_i)
   &= \sum_{c=1}^{C} \int_{B_r(x_i)}
        \kappa^{(c',c)}\bigl(\Phi_{x_i}(y)\bigr)\,a^{(c)}(y)\,\mathrm{d}y
   \nonumber \\
   &= \sum_{c=1}^{C}\sum_{p=1}^{P^2} w_p^{(c',c)} \!
        \int_{B_r(x_i)}\!\phi_p\!\bigl(\Phi_{x_i}(y)\bigr)\,a^{(c)}(y)\,\mathrm{d}y
   \nonumber \\
   &\approx \sum_{c=1}^{C}\sum_{p=1}^{P^2} w_p^{(c',c)} \cdot
        \sum_{j \in N(i)} V_{i,j}
        \phi_p\bigl(\Phi_{x_i}(x_j)\bigr)\,a^{(c)}(x_j),
\end{align}
giving a total of $C'{\times}C{\times}P^2$ trainable filter weights per layer.

\subsection{Hyperparameters}\label{app:hyperparams}
In \cref{tab:hyperparameters} we give the default hyperparameters that were used in our network architecture. We also provide the number of parameters for each model at their default configurations in \cref{tab:param_counts}.

\begin{table}[t]
\centering
\begin{minipage}[t]{0.48\textwidth}
\centering
\caption{Default architecture hyperparameters.}
\label{tab:hyperparameters}
\begin{tabular}{@{}ll@{}}
\toprule
Hidden dimension $H$ & 128 \\
Hierarchy levels $L$ & 3--4 \\
MP layers per level $K$ & 2 \\
FEM patch resolution $P$ & 5 \\
FEM basis type & P1 (bilinear) \\
Mixing mode & vector \\
Radius multiplier $\mu$ & 2.0--4.0 \\
Timestep embedding dim $d_t$ & 64 \\
Transformer blocks $K_T$ & 4 \\
Transformer heads & 4 \\
Pooling type & learned (residual) \\
\bottomrule
\end{tabular}
\end{minipage}%
\hfill
\begin{minipage}[t]{0.48\textwidth}
\centering
% from running scripts/audit_params.py
\caption{Parameter counts on \GaussianBlobs (square domain) at each model's default configuration.}
\label{tab:param_counts}
\begin{tabular}{@{}lr@{}}
\toprule
Model & Parameters \\
\midrule
GRIFDIR (ours)                 & 2,822,209 \\
% GRIFDIR (ours, fem\_conv, res-invariant=false)        &  2,822,209 \\
% GRIFDIR (ours, fem\_conv +lumped, res-invariant=true) &  2,558,017 \\
MP-GNN (simple\_mp)            & 4,580,417 \\
% FNO (no positional encoding)                          &    269,130 \\
FNO (grid positional encoding) & 2,891,018 \\
CNN U-Net                      & 2,122,946 \\
% GAOT ($32{\times}32$ latent)                          &  3,465,609 \\
GAOT ($64{\times}64$ latent)   & 3,471,753 \\
% MINO                                                  &  4,850,113 \\
% SplineCNN                                             & 14,138,973 \\ % (spline\_v3)
% MoNet                                                 & 14,141,745 \\ % (gmm\_conv)
\bottomrule
\end{tabular}
\end{minipage}
\end{table}

\section{Experimental Details}

The source code will be made publicly available upon acceptance.
\subsection{Datasets}\label{app:datasets}
\subsubsection{Gaussian Blobs}\label{app:gaussian_blobs}

Each sample is a conductivity field $\sigma: \Omega \to \mathbb{R}$, where $\Omega \subset \mathbb{R}^2$, generated as
\begin{equation}
    \sigma(x) = \min_{i=1,\dots,M} \left[ \sigma_b - (\sigma_b - \sigma_i^*)\, \exp\!\left(-\frac{1}{2}\left\|\mathbf{A}_i^{-1}\mathbf{R}_i^\top(x - \boldsymbol{\mu}_i)\right\|^2\right) \right],
\end{equation}
\begin{wrapfigure}{r}{0.3\linewidth}
    \centering
    \vspace{-\baselineskip}
    \includegraphics[width=\linewidth]{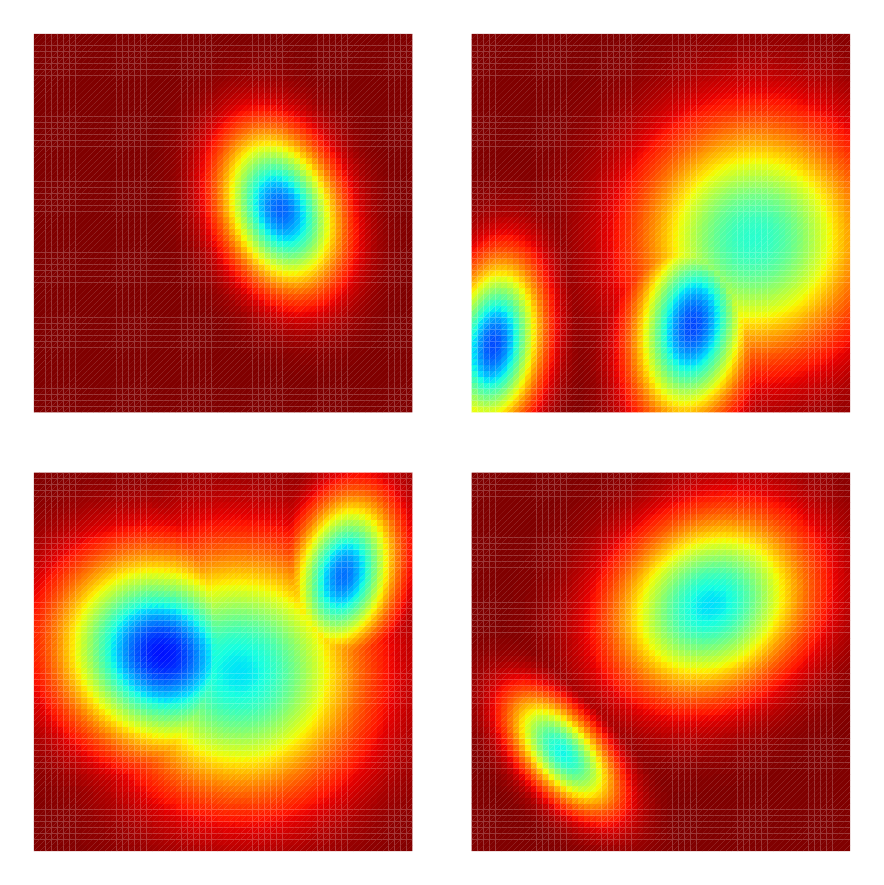}
    \caption{Ground truth samples from \GaussianBlobs.}
    \label{fig:gaussian_blob_gt}
    \vspace{-\baselineskip}
\end{wrapfigure}
where $M \sim \mathcal{U}\{1, \dots, N\}$ is the number of inclusions and $\sigma_b$ is the background conductivity. Each inclusion $i$ is parameterised by:
\begin{itemize}
    \item a centre $\boldsymbol{\mu}_i$, constrained such that the blob lies fully within $\Omega$;
    \item semi-axes $a_i \sim \mathcal{U}[a_{\min}, a_{\max}]$ and $b_i = r_i a_i$ with $r_i \sim \mathcal{U}[0.5, 1.0]$, collected in $\mathbf{A}_i = \mathrm{diag}(a_i, b_i)$;
    \item a rotation matrix $\mathbf{R}_i$ with angle $\alpha_i \sim \mathcal{U}[0, 2\pi)$;
    \item a minimum conductivity $\sigma_i^* = \sigma_b - d_i(\sigma_b - \sigma_{\min})$ with depth $d_i \sim \mathcal{U}[0.5, 1.0]$.
\end{itemize}

The same generator is used to build the following datasets that differ only in the support domain $\Omega$ and its triangulation.

\textbf{Square.} The domain $\Omega = [0,1]^2$ is tiled by a regular $32 \times 32$ Cartesian grid with each unit cell split into two triangles ($32 \times 32 \times 2 = 2{,}048$ triangles in total). 
The score network operates on the \emph{dual graph} of this triangulation: each triangle is a node carrying one DG-$0$ (piecewise-constant) scalar value, with an edge connecting any two triangles that share a mesh edge. 
This dual-graph layout gives $N_0 = 2{,}048$ nodes at the finest hierarchy level.

\textbf{Cross-domain.} Eight non-square shapes; \texttt{circle}, \texttt{circle\_with\_hole}, \texttt{e\_shape}, \texttt{l\_shape}, \texttt{plus}, \texttt{square\_with\_hole}, \texttt{x\_shape}, and a sub-sample of the square at the same hierarchy, each meshed by Gmsh \citep{geuzaine2009gmsh} at $h_{\max}=0.1$, used for the cross-shape generalisation experiments.

% \textbf{Graded.} Identical $32 \times 32$ topology to Square,
% but with one of four spatially varying coordinate warps applied to the
% base mesh, anisotropic refinement towards the left or right edge, or
% radial refinement towards the centre or boundary, so the four
% discretisations share a node budget but differ in node density
% distribution.

For every dataset we evaluate $\sigma$ at the cell-centroid coordinates of its triangulation, store the field as one channel per cell, and apply a $90/10$ train/test split. Table~\ref{tab:gblob_datasets} reports total sample counts and per-level
node counts of the multi-resolution hierarchy used by the score network.

\begin{figure}[tb]
    \centering
    \includegraphics[width=0.9\linewidth]{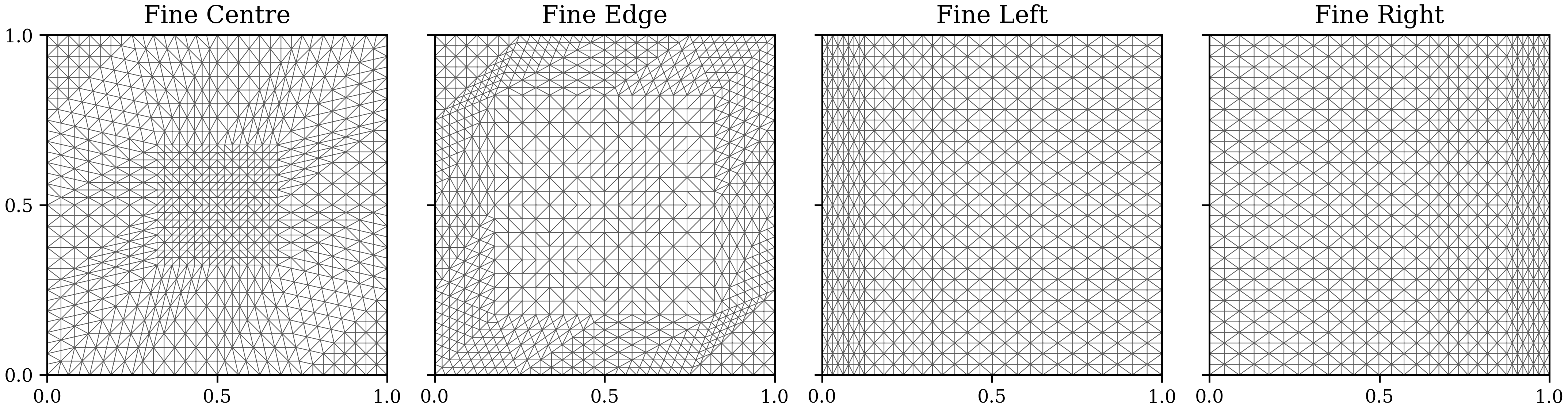}
    \caption{To test the discretisation invariance of the method we use the following graded meshes where the density of nodes is uneven across the mesh.}
    \label{fig:graded_meshes}
\end{figure}

\begin{table}[tb]
\centering
\caption{\GaussianBlobs datasets: total samples
($\lceil 0.9N\rceil$ assigned to train at runtime) and node counts at
each hierarchy level $N_0,N_1,N_2,N_3$ (finest to coarsest).}
\label{tab:gblob_datasets}
\begin{tabular}{@{}llrrrrr@{}}
\toprule
& Domain / pattern & $N_{\text{samples}}$ & $N_0$ & $N_1$ & $N_2$ & $N_3$ \\
\midrule
\multicolumn{7}{@{}l}{\emph{Square}}\\
& \texttt{nx32\_ny32}            & 10,000 & 2,048 & 512   & 128 & 32  \\
\midrule
\multicolumn{7}{@{}l}{\emph{Cross-domain (8 shapes, $h_{\max}=0.1$)}}\\
& circle                         & 1,000  & 2,974 & 757   & 212 & 119 \\
& circle\_with\_hole             & 1,000  & 2,736 & 714   & 202 & 104 \\
& e\_shape                       & 1,000  & 3,380 & 860   & 246 & 81  \\
& l\_shape                       & 1,000  & 2,814 & 732   & 190 & 58  \\
& plus                           & 1,000  & 4,718 & 1,218 & 318 & 130 \\
& square                         & 1,000  & 2,048 & 512   & 128 & 32  \\
& square\_with\_hole             & 1,000  & 3,642 & 942   & 248 & 173 \\
& x\_shape                       & 1,000  & 2,804 & 726   & 206 & 98  \\
% \hline\hline
% \multicolumn{7}{@{}l}{\emph{Graded (this work; warped $32{\times}32$ topology)}}\\
% & \texttt{fine\_left}            & 10,000 & 2,048 & 512   & 128 & 32  \\
% & \texttt{fine\_right}           & 10,000 & 2,048 & 512   & 128 & 32  \\
% & \texttt{fine\_centre}          & 10,000 & 2,048 & 512   & 128 & 32  \\
% & \texttt{fine\_edge}            & 10,000 & 2,048 & 512   & 128 & 32  \\
\bottomrule
\end{tabular}
\end{table}

\subsubsection{Pinball}\label{app:pinball}
\begin{figure}[tb]
    \centering
    \includegraphics[width=1.0\linewidth]{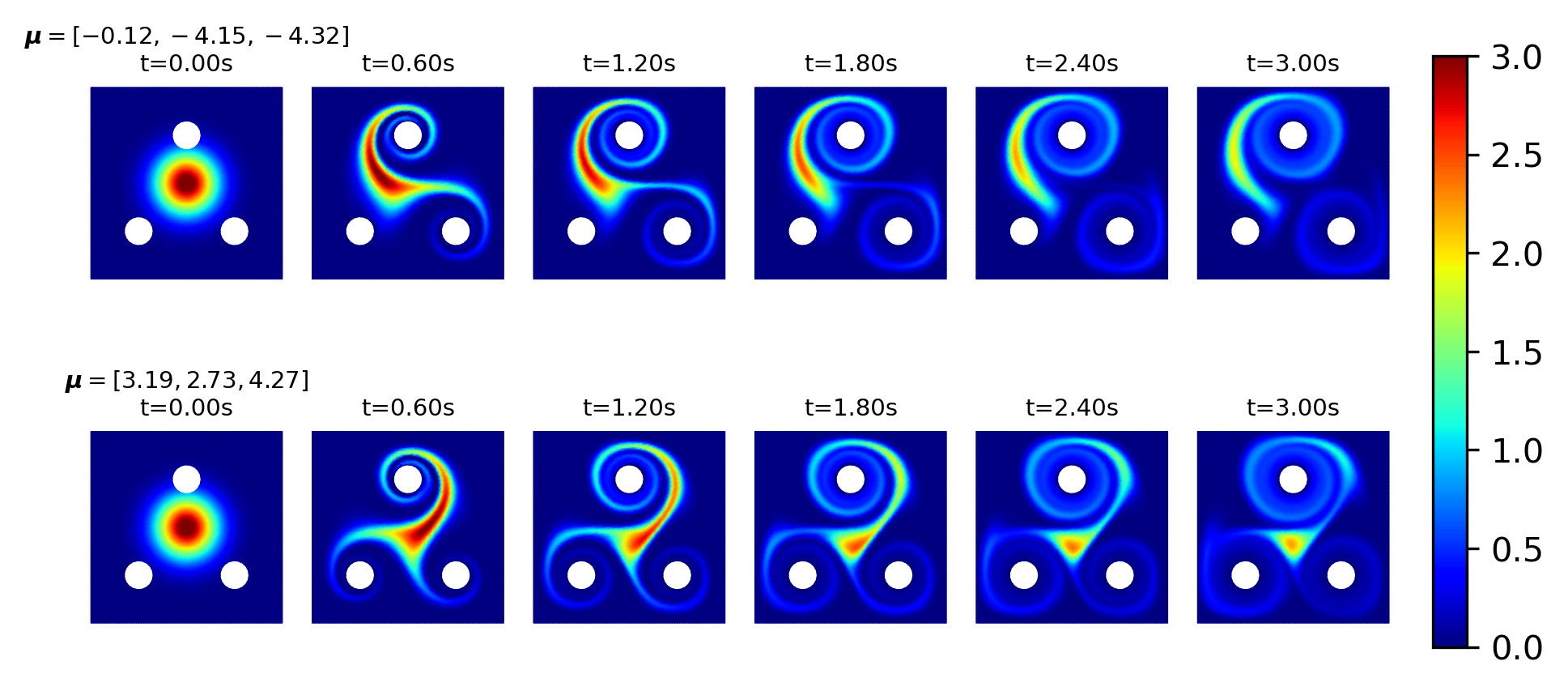}
    \caption{Two rollouts of \Pinball for different velocities $\boldsymbol{\mu}$ from the training set.}
    \label{fig:pinball_dataset}
\end{figure}

\begin{figure}
    \centering
    \includegraphics[width=0.9\linewidth]{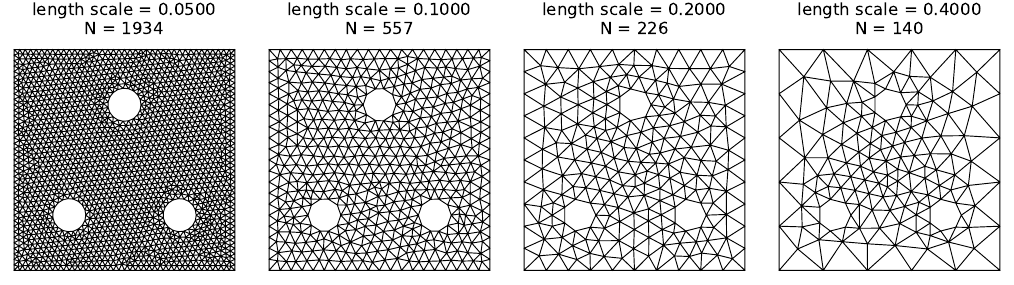}
    \caption{The \Pinball mesh for different length scales and number of elements $N$.}
    \label{fig:pinball_mesh}
\end{figure}

We consider the fluid pinball setup from \citet{tomasetto2025reduced} (Section D.). The pinball example involves a 2D advection-diffusion problem with an implicit parametric dependence. In particular, they consider a (incompressible) fluid in a square domain $[-1,1]^2$ with three rotating cylinders centred at $(-0.5, -0.5), (0.5, -0.5)$ and $(0.0, 0.5)$ with radius $0.15$. The constant rotation is given by velocities $\boldsymbol{\mu} = [v_1, v_2, v_3]$. The rotation leads to a fluid motion inside the square domain, where the velocity $v:[-1,1]^2 \to \mathbb{R}^2$ and pressure $p:[-1,1]^2 \to \mathbb{R}$ are determined by the (steady-state) Navier-Stokes equation. \citet{tomasetto2025reduced} use viscosity $\nu=1.0$ and no-slip boundary conditions on the external walls and Dirichlet boundary conditions on the cylinders. They then consider a quantity (e.g., mass or particle density) $\rho:[-1,1]^2 \times [0,T] \to \mathbb{R}$ that spreads inside the domain according to 
\begin{align}
    \rho_t + \nabla \cdot (-\eta \nabla \rho + v(\boldsymbol{\mu}) y) = 0,
\end{align}
with homogeneous Neumann boundary conditions, $\eta=0.001$, $T=3$ and $v(\boldsymbol{\mu})$ as the solution of the Navier-Stokes equation given the velocity of the cylinders $\boldsymbol{\mu}$. For a fixed initial condition 
\begin{align}
    \rho(x, 0) = \frac{10}{\pi} \exp(-10 x_1^2 - 10 x_2^2)
\end{align}
the distribution $\rho$ depends directly on the fluid velocity $v(\boldsymbol{\mu})$. We collect roll-outs of $\rho_t$ for 400/50/50 varying $\boldsymbol{\mu}$ as our training/validation/test sets and aim to learn this distribution (independent of $t$). This is a challenging distribution as the domain has three holes (where the cylinders are). Examples of the pinball dataset are shown in Figure~\ref{fig:pinball_dataset}. We plot the mesh for different length scales in Figure~\ref{fig:pinball_mesh}. In addition to the meshes shown there we use meshes with a length scale $0.03$ and $0.025$ containing 5,266 and 7,525 mesh vertices.

\begin{figure}
    \centering
    \includegraphics[width=0.6\linewidth]{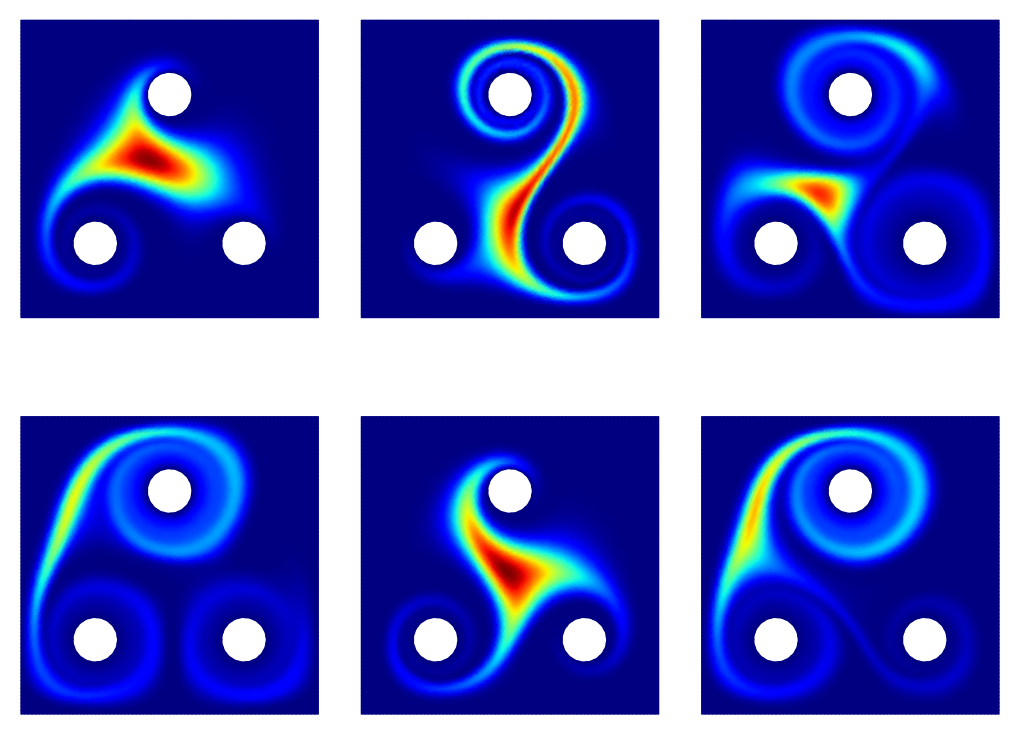}
    \caption{Unconditional samples for \Pinball. See Figure~\ref{fig:pinball_dataset} for samples from the dataset. The same colour scale is used in both figures, with a minimum of 0 and a maximum of 3.}
    \label{fig:uncond_samples_pinball}
\end{figure}

\subsubsection{Cylinder Flow}\label{app:cylinder_flow} 
We make use of the \Cylinder dataset from \cite{han2022predicting}. The dataset describes the flow around a cylinder on a fixed (unstructured) mesh of \num{1699} nodes. The dataset consists of $101$ simulations each for $400$ timesteps. Every sample has three channels, i.e., velocity in $x$-direction ($u$), velocity in $y$-direction ($v$) and pressure $p$. We use the same train-test split as \cite{shi2025meshinformed} and use \num{30000} samples for training and \num{5000} for testing.

\subsubsection{DeepJEB}
\DeepJEB \citep{hong2025deepjeb} contains \num{2138} 3D bracket geometries, each with the von Mises stress field induced by a vertical load. The stress concentrates in small hotspots near the mounting points, and the geometries themselves vary a lot, so the model has to generalise across shapes. We use \num{1924} brackets for training and the remaining 100 for testing.

We use the 3D version of our model (via a trilinear interpolation of the kernel), with the same multi-scale architecture. The brackets are represented as point clouds and we create a graph using kNN with 8 neighbours. We train two models on different resolutions, one capping the sampled nodes at \num{6000} points and one at \num{24000} points. For both, we use three downsampling ratios, taking $100\%, 25\%$ and $6.25\%$ of the input points.

\subsection{Baseline Approaches}
\label{app:baselines_arch}

\paragraph{MP-GNN}
GNN based on the message passing framework, modelled after the MP-PDE Solver \citep{brandstetter2022message}. Each node corresponds to a spatial point on the mesh, and edges encode local spatial connectivity. For each edge $(i, j)$, the message is computed from the hidden states $h_i, h_j$, the difference in input values $a_i - a_j$, and the relative position $x_i - x_j$:
\begin{align*}
    m_{ij} = \phi_m\bigl([h_i, h_j, a_i - a_j, x_i - x_j]\bigr)
\end{align*}
where $\phi_m$ is a two-layer MLP with Swish activations. Using differences rather than raw values makes the messages translation-equivariant in both signal and position space. Aggregated messages (mean pooling) are combined with the current hidden state via a residual connection:
\begin{align*}
    h_i \leftarrow h_i + \phi_u\bigl([h_i,, \bar{m}_i]\bigr),
\end{align*}
followed by instance normalisation. The time step $t$ is embedded via a Fourier feature embedding and concatenated with the node features in the encoder, allowing the network to function as a time-conditioned score network $s_\theta(a, t, x)$ for diffusion-based generative modelling over function spaces. The configuration used hidden dimension $128$, $4$ message-passing layers, mean aggregation, time embedding dimension equal to the hidden dimension. %The final model has \num{4580353} parameters.

\paragraph{FNO}
The Fourier Neural Operator \citep{li2020fourier} parametrises the score in the spectral domain. Each FNO block lifts the input with a $1\times 1$ convolution, applies a spectral convolution that truncates the 2D FFT to its top-$K$ low-frequency modes, multiplies by a learned complex weight tensor at those modes, and inverse-transforms back to the spatial domain; the spectral output is added to a residual $1\times 1$ convolution and passed through a GELU non-linearity:
\begin{align*}
    h \leftarrow \sigma\bigl(\mathcal{F}^{-1}\!\bigl(R \cdot \mathcal{F}(h)\bigr)_{|k|\le K}\,+\,W h\bigr).
\end{align*}
Because the network operates on a regular grid, the unstructured centroid mesh is first interpolated onto a $32 \times 32$ image grid (matching the conductivity dataset's reference resolution), processed by the FNO stack, and bilinearly interpolated back to the mesh nodes for loss computation. Time conditioning follows the same Fourier-feature embedding as the GNN baselines, concatenated as an extra channel before the first lift. Default configuration: $4$ FNO blocks, $K=8$ modes per spatial axis, hidden dimension $128$, we use grid-coordinate positional encoding and optional $12.5\%$ domain padding to handle non-periodic boundaries. %The final model has \num{2890506} parameters.

\paragraph{GAOT}
The Geometry-Aware Operator Transformer (GAOT) \citep{wen_geometry_2025} is a continuous-kernel encoder/decoder paired with a transformer processor. The diffusion time step $t$ is mapped through a Fourier-feature time embedding and projected to $n_{\text{time}}$ extra channels which are concatenated to every node's input alongside the function value $a(x_j)$ (and an optional domain one-hot $d_j$ in the multi-domain setting), giving a per-node feature vector $\tilde a(x_j) = [a(x_j),\, \tau(t),\, d_j]$. The MAGNO encoder then lifts these features onto a fixed regular grid of latent tokens via a continuous radius-ball kernel: for each latent location~$\xi$,
\begin{align*}
    z(\xi) = \frac{1}{|N(\xi)|} \sum_{x_j \in B_r(\xi)} \kappa_\theta\bigl(\xi - x_j,\, \tilde a(x_j)\bigr),
\end{align*}
where $\kappa_\theta$ is a per-edge MLP and the aggregation is normalised by neighbour count. The latent token grid is patchified and processed by a stack of standard ViT-style transformer blocks with multi-head self-attention. A symmetric MAGNO decoder lifts the processed tokens back onto the mesh, and a learned signed-distance geometric embedding is added to the latent tokens before the processor. Default configuration: latent grid $64 \times 64$, MAGNO radius $0.033$ in unit-square coordinates, $3$-layer MAGNO MLP with hidden dimension $64$ and lifting dimension $64$, transformer hidden dimension $256$, time embedding dimension $128$ projected to $n_{\text{time}}=8$ per-node channels, $3$ transformer blocks, patch size $2$, geometric embedding and self-attention enabled, absolute positional encoding. %The final model has \num{3463497} parameters.

\paragraph{CNN}
We additionally evaluate a CNN-based baseline. As CNNs operate on regular grids, we first interpolate the mesh representation onto a fixed image grid. A standard Attention U-Net \citep{dhariwal2021diffusion} is then applied in image space, after which the predictions are interpolated back onto the original mesh. Default configuration: hidden dimension $128$, $4$ down/up-sampling stages with single convolutions per block, residual connections, attention at the bottleneck, max-pool downsampling and bilinear upsampling, no dropout. %The final model has \num{2122882} parameters.

\paragraph{MINO}
The Mesh-Informed Neural Operator (MINO) \citep{shi2025meshinformed} is a transformer-based
neural operator designed for generative modelling of functions on arbitrary
discretisations. A graph neural operator encoder aggregates the input function, together
with the diffusion time embedding, from the input mesh onto a fixed set of $N \times N$ latent grid in the domain via a radius-ball kernel integral, analogous to \cref{eq:gno_discrete}.
The resulting latent tokens are processed by a stack of attention blocks. A cross-attention
decoder, in which the coordinates of the output nodes act as queries attending to the
latent tokens, returns the score at every mesh node. Because the latent query set is
independent of the input mesh, MINO is discretisation-agnostic by construction. Unlike
GRIFDIR, it has no hierarchy of local convolutions: all mesh information is compressed
into the latent tokens in a single encoding step. Time conditioning follows the same
Fourier-feature embedding as the other baselines. We use MINO's code at \url{https://github.com/yzshi5/MINO} for our experiments, however we swap the flow matching objective for a diffusion objective. {Default configuration:} latent grid $64 \times 64$, hidden dimension $128$, encoder transformer depth $4$, attention heads $4$, decoder depth $2$ and a GNO neighbour radius of $0.07$.

% \paragraph{Geo-FNO}
% Geo-FNO \citep{li2023fourier} extends FNO to irregular domains by learning a deformation
% $\phi_\theta:\Omega\to[0,1]^2$ from the physical domain to a uniform latent computational
% domain. The first layer replaces the FFT with a direct non-uniform Fourier transform
% evaluated at the mesh points,
% $\hat h(k) = \sum_{j} h(x_j)\, e^{-2\pi i \langle k, \phi_\theta(x_j)\rangle}$,
% truncated to the lowest $K$ modes. The latent representation is then processed by standard
% FNO blocks on the regular grid. The final layer evaluates the inverse transform at the
% deformed coordinates $\phi_\theta(x_j)$ of the output nodes, so that no interpolation
% between mesh and grid is required. We parametrise $\phi_\theta$ as the identity plus a
% learned MLP residual in the coordinates. Time conditioning follows the FNO baseline. \textcolor{red}{Default configuration:}

\subsection{Evaluation Metrics}
\label{app:evaluation_metrics}
For conditional sampling we obtain a set of samples $\{\hat{a}_i^{(k)}\}, k=1, \dots, K$ for every observation $y_i$, $i=1, \dots, n$. Here $n$ is the number of observations and $K$ the number of posterior samples per observation. We compute the MSE of the posterior mean, the energy score.
To evaluate the unconditional models we evaluate the MMD. 

\paragraph{MSE/RMSE of the Posterior Mean}
For each observation $y_i$, we compute the posterior mean as
\begin{align*}
    \bar{a}_i = \frac{1}{K} \sum_{k=1}^{K} \hat{a}_i^{(k)}.
\end{align*}
The MSE/RSME is then defined as
\begin{align*}
    \mathrm{MSE} = \frac{1}{n} \sum_{i=1}^{n}
\left\| a_i - \bar{a}_i \right\|^2, \quad \mathrm{RMSE} = \sqrt{\frac{1}{n} \sum_{i=1}^{n}
\left\| a_i - \bar{a}_i \right\|^2}
\end{align*}
This metric measures the average deviation between the ground truth parameters $a_i$ and the posterior mean estimates.

\paragraph{Energy Score (ES) \citep{gneiting2007strictly}}
For each observation $y_i$ with posterior samples $\{\hat{a}_i^{(k)}\}_{k=1}^{K}$, the energy score with $\beta = 1$ is defined as
\begin{align*}
    \mathrm{ES}_i =
\frac{1}{K} \sum_{k=1}^{K}
\left\| \hat{a}_i^{(k)} - a_i \right\|^\beta
-
\frac{1}{2K^2}
\sum_{k=1}^{K} \sum_{l=1}^{K}
\left\| \hat{a}_i^{(k)} - \hat{a}_i^{(l)} \right\|^\beta.
\end{align*}
The final energy score is obtained by averaging over all samples,
\begin{align*}
    \mathrm{ES} = \frac{1}{n} \sum_{i=1}^{n} \mathrm{ES}_i.
\end{align*}
The ES jointly evaluates calibration and sharpness of the posterior distribution. Lower values indicate better probabilistic predictions.

\paragraph{Maximum Mean Discrepancy (MMD) \citep{gretton2012kernel}}
Given the sets of samples $X = \{ x_1, \dots, x_n\}$ and $Z=\{z_1, \dots z_m\}$  for $n,m \in \mathbb{N}$, we compute the squared
MMD using the standard unbiased U-statistic estimator
\begin{align}
    \text{MMD}^2_u(X, Z) = \frac{1}{n(n-1)} \sum_{i \ne j} k(x_i, x_j)
    + \frac{1}{m(m-1)} \sum_{i \ne j} k(z_i, z_j)
    - \frac{2}{nm} \sum_{i=1}^n \sum_{j=1}^m k(x_i, z_j),
\end{align}
for a Gaussian kernel $k(\cdot, \cdot)$ with length scale $\ell$, defined as
\begin{align}
    k(x, z) = \exp\left(-\frac{\|x - z\|^2}{2\ell^2}\right).
\end{align}

\paragraph{Sliced Wasserstein Distance (SWD) \citep{bonneel2015sliced}}
The sliced Wasserstein distance measures distances between two distributions $\mu$ and $\nu$ in high-dimensional spaces by choosing random directions in the space and projecting the data onto it, resulting in two 1D distributions for which we can cheaply find the Wasserstein distance. 
For $p \geq 1$, the sliced
$p$-Wasserstein distance between distributions $\mu$ and $\nu$ on $\mathbb{R}^N$ is
\begin{equation}
    \mathrm{SW}_p(\mu,\nu) = \left( \int_{\mathbb{S}^{N-1}}
    W_p^p\big(\theta_\# \mu, \theta_\# \nu\big)\, \mathrm{d}\sigma(\theta) \right)^{1/p},
\end{equation}
where $\sigma$ is the uniform measure on the unit sphere $\mathbb{S}^{N-1}$ and
$\theta_\# \mu$ is the push-forward of $\mu$ under $x \mapsto \langle \theta, x \rangle$. To approximate this, we use the sliced Wasserstein distance function using \texttt{POT} \citep{pot_lib}, with $p=2$ and $256$ projections. We draw compute the SWD with $5$ different seeds for the random projection and compute the mean as the final score.

\subsection{Computational Resources}\label{app:comp_resources}
Training of the models was performed on a cluster where each node comprises 2 Intel Xeon Platinum 8468 CPUs (96 cores total) and 4 Intel Data Center GPU Max 1550 (Ponte Vecchio) GPUs with 128 GB memory per GPU. Training took about $24$h for 100 epochs on the Pinball dataset, $6$h for the Gaussian Blobs for 250 epochs, Cylinder Flow $24$h for 200 epochs and DeepJEB $20$h for 50 epochs. Inference (unconditional and conditional sampling) was performed on a GPU cluster comprising NVIDIA V100, A100 and H100 GPUs. 

\begin{wraptable}{r}{0.55\linewidth}
\centering
\small
\vspace{-\baselineskip}
\caption{Reconstruction errors for the Poisson inverse problem on \GaussianBlobs.}
\label{tab:fun_dps_daps}
\begin{tabular}{lcc}
\toprule
& RMSE $(\downarrow)$ & Energy Score $(\downarrow)$ \\
\midrule
Fun-DPS  & $0.0197 \pm 0.0128$ & $0.6709 \pm 0.4479$ \\
Fun-DAPS & $\mathbf{0.0069 \pm 0.0049}$ & $\mathbf{0.2549 \pm 0.1919}$ \\
\bottomrule
\end{tabular}
\vspace{-\baselineskip}
\end{wraptable}

\section{Further results and ablations}
In this section, we provide further results for the experiments reported in \cref{sec:invariance,sec:geometry,sec:realistic}. Namely, in \cref{app:conditional} we provide results from the conditional experiments, in \cref{app:ablations} we provide ablations of our model and in \cref{app:deepjeb} we provide scaling details on \DeepJEB.

\begin{table}[!ht]
\centering
\caption{We evaluate the models on graded meshes for the \GaussianBlobs dataset and evaluate using $1024$ samples for all models. We report the MMD and SWD metrics. Best results are in \textbf{bold}, second best are \underline{underlined}.}
\resizebox{\linewidth}{!}{
\begin{tabular}{lcccccccc}
\toprule
& \multicolumn{2}{c}{{Fine left}}
& \multicolumn{2}{c}{{Fine centre}} 
& \multicolumn{2}{c}{{Fine edge}} 
& \multicolumn{2}{c}{{Fine right}} \\
\cmidrule(lr){2-3} \cmidrule(lr){4-5} \cmidrule(lr){6-7} \cmidrule(lr){8-9}
&  MMD $(\downarrow)$ &  SWD $(\downarrow)$
&  MMD $(\downarrow)$ &  SWD $(\downarrow)$
&  MMD $(\downarrow)$ &  SWD $(\downarrow)$
&  MMD $(\downarrow)$ &  SWD $(\downarrow)$ \\
\midrule
GAOT 
& 0.730 & 6.078 & 0.687 & 4.015
& 0.756 & 3.582 & \underline{0.191} & \underline{0.077}\\
MINO 
& \textbf{0.136} & \textbf{0.046} & \textbf{0.056} & \textbf{0.021}
& \underline{0.183} & \underline{0.063} & \textbf{0.136} & \textbf{0.048}\\
MP-GNN
& \underline{0.223} & \underline{0.076} & \underline{0.106} & \underline{0.036}
& 0.218 & 0.076 & 0.234 & 0.083\\
GRIFDIR
& 0.231 & 0.079 & 0.110 & 0.044
& \textbf{0.062} & \textbf{0.026} & 0.239 & 0.080\\
\bottomrule
\end{tabular}}
\label{tab:discretisation_invariance}
\end{table}

\begin{figure*}[tb]
    \centering
    \textbf{Fun-DPS}\\
    \begin{subfigure}[t]{0.9\linewidth}
    \centering
    \includegraphics[width=\linewidth]{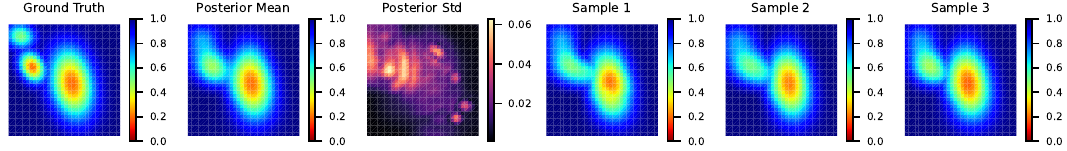}
    \caption*{}
    \label{fig:dps_poisson}
    \end{subfigure}\\
    \vspace{-10pt}
    % \centering
    \textbf{Fun-DAPS}\\
    \vspace{-5pt}
    \begin{subfigure}[t]{0.9\linewidth}
    \centering
        \label{fig:daps_poisson}
        \includegraphics[width=\linewidth]{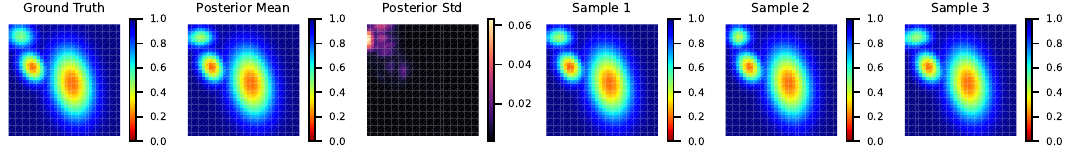}
    \end{subfigure}
        \caption{We use Fun-DPS and Fun-DAPS to sample conditionally on the Poisson operator for \GaussianBlobs. The posterior mean and standard deviation are taken across $100$ samples, with three such samples plotted on the right. }
        \label{fig:blob_poisson}
\end{figure*}

\subsection{Conditional Sampling}\label{app:conditional}
We use the unconditionally trained GRIFDIR as a prior and approximate the guidance
term with Fun-DPS and Fun-DAPS,
drawing $K=100$ posterior samples per observation. Fun-DPS backpropagates through the
score network at every step, whereas Fun-DAPS is backpropagation-free and hence cheaper.

\begin{figure*}[t]
    \centering
    \textbf{Fun-DPS}\\
    \begin{subfigure}[t]{0.9\linewidth}
        \centering
        \includegraphics[width=\linewidth]{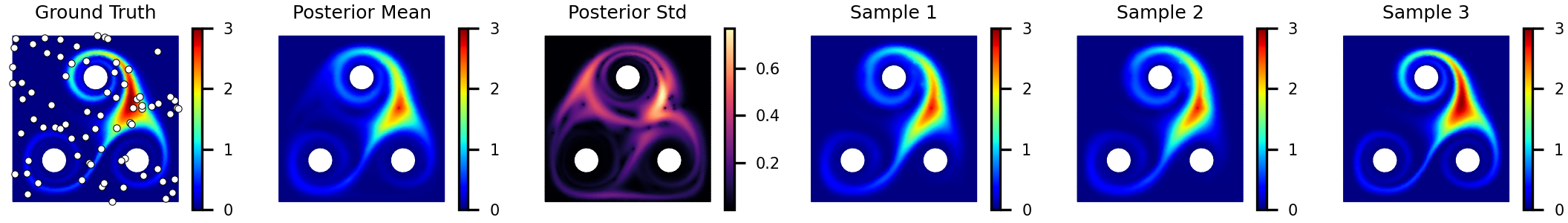}
        %\caption{Sparse sensor reconstruction with Fun-DPS for 75 randomly placed sensors.}
        \caption*{}
        \label{fig:fun_dps}
    \end{subfigure}
    \hfill 
    \textbf{Fun-DAPS}\\
    \begin{subfigure}[t]{0.9\linewidth}
        \centering
        \includegraphics[width=\linewidth]{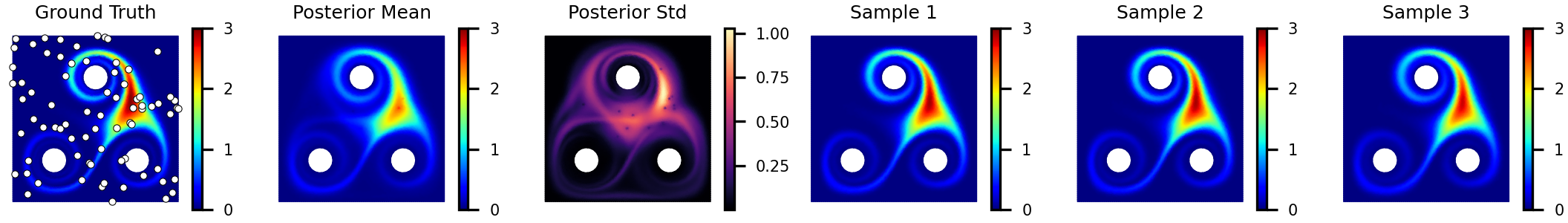}
        %\caption{Sparse sensor reconstruction with Fun-DAPS for 75 randomly placed sensors.}
        \caption*{}
        \label{fig:fun_daps}
    \end{subfigure}
    \caption{Sparse sensor reconstruction with 75 randomly placed sensors with Fun-DPS and Fun-DAPS for \Pinball. For each method, we show the ground truth, posterior mean and standard deviation (computed with $100$ samples), and three posterior samples.}
    \label{fig:reconstruction_comparison}
\end{figure*}

\begin{figure}[tb]
\centering
\text{Performance vs. Number of Sensors on \GaussianBlobs}\\
\begin{subfigure}{0.45\linewidth}
\includegraphics[width=\linewidth]{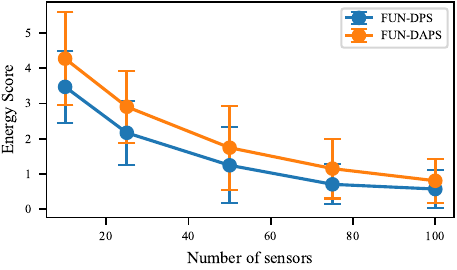}
\end{subfigure}
\begin{subfigure}{0.45\linewidth}
\includegraphics[width=\linewidth]{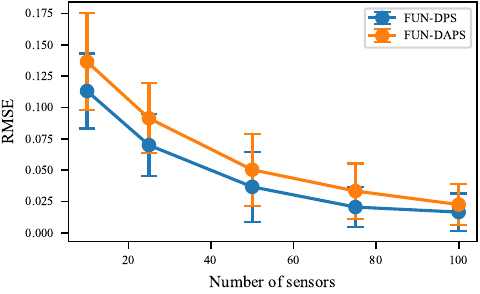}
\end{subfigure}
    \caption{Conditional Sampling for \GaussianBlobs. We compare the conditional methods Fun-DPS and Fun-DAPS for sampling from the conditional distribution arising from an increasing number of randomly placed sensors.}
    \label{fig:blobs_sensor_vs_error}
\end{figure}

\paragraph{Sparse Sensors}
We condition on $10$ to $100$ randomly placed point observations for the \GaussianBlobs (see \cref{fig:blobs_sensor_vs_error}) and $25$ to $100$ for the \Pinball dataset (\cref{fig:reconstruction_comparison}, \cref{tab:sensors}). On both datasets, RMSE and energy score decrease steadily as sensors are added, as we would expect. Fun-DPS is consistently more accurate than Fun-DAPS. However, Fun-DPS has a higher computational cost, as it requires to backpropagate through the score model at every sampling step.

% \subsubsection{Gaussian Blobs}
% \textbf{Sparse Sensors} 
% We next consider using our model for conditional sampling. 
% First, we consider the sparse sensor experiment described in \cref{sec:experiment_setup}. 
% Our model is trained unconditionally on the Gaussian blob dataset with a square domain. 
% In \cref{fig:gaussian_blob_sensors} we plot one sample from Fun-DPS and one from Fun-DAPS conditioned on $10, 50$ and $100$ sensors. 
% We see that as the number of sensors increase, the conditional samples become more accurate, more closely resembling the ground truth. 
% In \cref{fig:blobs_sensor_vs_error}, we plot the energy scores and RMSE, alongside standard deviations.
% For this we use $n=10$ observations, with $K=100$ samples per observation. 
% Similarly to what we observe in \cref{fig:gaussian_blob_sensors}, as the number of sensors increases, so does the reconstruction performance. 
% This is in line with what we would expect, since by increasing the number of sensors, we are increasing the information on which we condition. 
% % We also note that DPS performs better than DAPS

\paragraph{Poisson}
We apply our model to the Poisson inverse problem in the noiseless setting for \GaussianBlobs. For Fun-DPS we use a guidance weight of $2.0$ and for Fun-DAPS, we set the noise level to $\beta_y=1e-5$, with $20$ Langevin steps. For both we use $400$ sampling steps. We report the mean and standard deviation over $100$ independent reconstructions for one observation, alongside three samples, in Figure~\ref{fig:blob_poisson}. 
Doing the same, we get the reconstruction errors reported in Table~\ref{tab:fun_dps_daps}, where we see in this case Fun-DAPS is more accurate. 

\begin{table}[t]
\centering
\caption{Performance for varying number of sensors on \Pinball.}
\label{tab:sensors}
\resizebox{\linewidth}{!}{%
\begin{tabular}{lcccccccc}
\toprule
& \multicolumn{2}{c}{\textbf{25 sensors}}
& \multicolumn{2}{c}{\textbf{50 sensors}}
& \multicolumn{2}{c}{\textbf{75 sensors}}
& \multicolumn{2}{c}{\textbf{100 sensors}} \\
\cmidrule(lr){2-3} \cmidrule(lr){4-5}
\cmidrule(lr){6-7} \cmidrule(lr){8-9}
& RMSE $(\downarrow)$ & ES $(\downarrow)$
& RMSE $(\downarrow)$ & ES $(\downarrow)$
& RMSE $(\downarrow)$ & ES $(\downarrow)$
& RMSE $(\downarrow)$ & ES $(\downarrow)$ \\
\midrule
Fun-DPS
& $0.289\pm0.052$ & $16.6\pm3.33$
& $0.188\pm0.054$ & $9.21\pm3.59$
& $0.140\pm0.042$ & $6.18\pm2.54$
& $0.110\pm0.042$ & $4.53\pm2.29$ \\
Fun-DAPS
& $0.332\pm0.059$ & $20.1\pm4.45$
& $0.249\pm0.064$ & $14.03\pm4.37$
& $0.218\pm0.057$ & $11.5\pm3.82$
& $0.183\pm0.054$ & $8.90\pm3.65$ \\
\bottomrule
\end{tabular}}
\end{table}

\begin{figure}[h]
    \centering
    \includegraphics[width=0.5\linewidth]{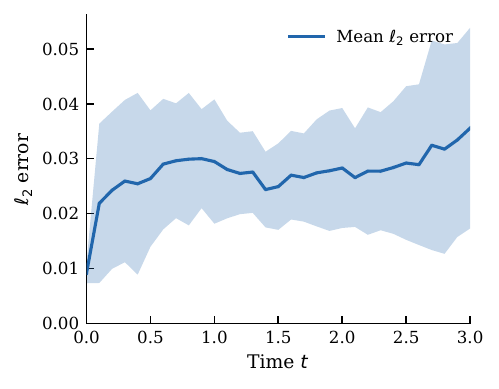}
    \caption{The mean $\ell_2$-error of the conditional GRIFDIR model trained on \Pinball over the rollout time $t$.}
    \label{fig:pinball_cond_l2}
\end{figure}

\paragraph{Conditional sampling for \Pinball}
We evaluate conditional sampling for \Pinball on 50 random velocity parameters $\boldsymbol{\mu}$, each rolled out over 30 time steps in $[0, T]$. Here, we train GRIFDIR conditional on the cylinder velocities $\mathbf{\mu}$ and time $t$. For each case, we draw one conditional sample and report the mean $\ell^2$ error against the ground truth $\rho_t$.
As shown in Figure~\ref{fig:pinball_cond_l2}, the error is minimal at $t=0$ since the initial condition is fixed and independent of $\boldsymbol{\mu}$. It increases during the early transient phase as $\rho$ is advected by $v(\boldsymbol{\mu})$, while the standard deviation remains moderate, indicating consistent generalisation across parameters.
Overall, the model captures the parametric dependence of the advection–diffusion dynamics and produces accurate samples over the full time horizon.

\begin{figure}
    \centering
    \includegraphics[width=0.5\linewidth]{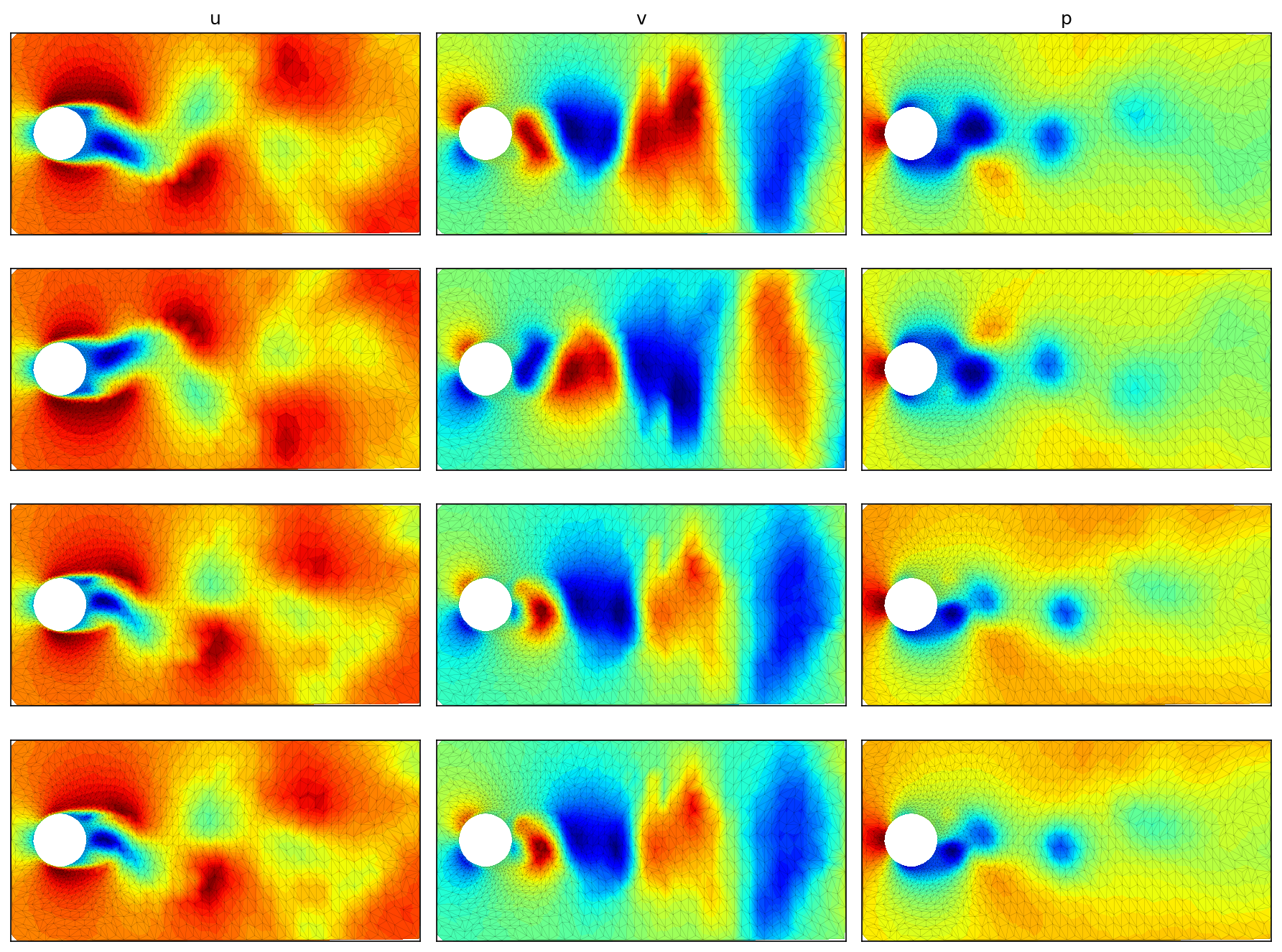}
    \caption{Unconditional samples from \Cylinder dataset using GRIFDIR.}
    \label{fig:grifdir_samples_cylinder}
\end{figure}

\begin{figure}
    \centering
    \includegraphics[width=1.0\linewidth]{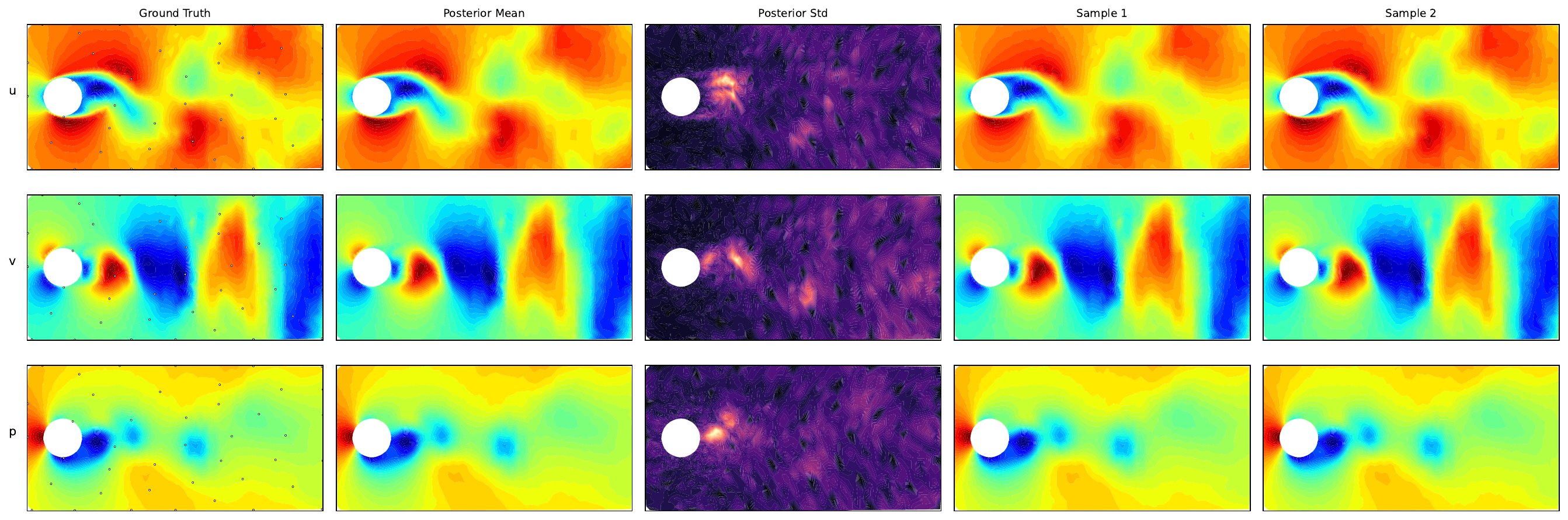}
    \caption{Conditional Samples from \Cylinder dataset using GRIFDIR using $42$ sensor points with a farthest-point placement of the sensors. We show the ground truth, the posterior mean, the posterior pointwise standard deviation and two samples.}
    \label{fig:grifdir_samples_cylinder_posterior}
\end{figure}

\subsection{Ablations}\label{app:ablations}

\paragraph{Transformer, FiLM and Aggregation}

First, we compare mean aggregation against volume-weighted (``lumped'') aggregation and the effect of the latent transformer and the FiLM embedding. We do this on \GaussianBlobs, trained on $2{,}048$ nodes and report the results in \cref{tab:component_ablation}. 
We see that for evaluating on higher resolutions, using FiLM, lumped aggregation and the transformer performs best. 
Keeping FiLM and the transformer fixed, using lumped aggregation tends to improve results and in the cases it doesn't we get roughly the same scores. Indeed, without it, we see many of the experiments on the coarser and finer resolutions collapse to generating near-constant samples.

\begin{table}[!ht]
    \centering
    \caption{Component ablation of the latent transformer (Trans.), the FiLM time embedding and the lumped aggregation for \GaussianBlobs. MMD computed with $1024$ generated samples against $2000$ reference samples; SWD computed with $1024$ samples. Columns give the number of nodes in the evaluation discretisation; ``Mean'' is the average over the three discretisations.}\label{tab:component_ablation}
    \resizebox{\textwidth}{!}{%
    \begin{tabular}{lllcccccccc}
    \toprule
        & & & \multicolumn{4}{c}{MMD $\downarrow$} & \multicolumn{4}{c}{SWD $\downarrow$} \\
        \cmidrule(lr){4-7} \cmidrule(lr){8-11}
        Trans. & FiLM & Lumped & $1152$ pts. & $2048$ pts. & $8192$ pts. & Mean & $1152$ pts. & $2048$ pts. & $8192$ pts. & Mean \\
        \midrule
        \checkmark & \checkmark & \checkmark & $0.064$ & $0.090$ & $\mathbf{0.053}$ & $\mathbf{0.069}$ & $0.029$ & $0.033$ & $\mathbf{0.025}$ & $\mathbf{0.029}$ \\ 
        \checkmark & \checkmark & ---        & $0.471$ & $0.088$ & $0.379$ & $0.313$ & $0.162$ & $0.033$ & $0.130$ & $0.108$ \\ 
        \checkmark & ---        & \checkmark & $0.094$ & $\underline{0.078}$ & $0.253$ & $0.142$ & $0.035$ & $\mathbf{0.028}$ & $0.089$ & $0.051$ \\ 
        \checkmark & ---        & ---        & $0.652^\dagger$ & $0.097$ & $0.650$ & $0.466$ & $0.260^\dagger$ & $0.035$ & $0.256$ & $0.184$ \\ 
        ---        & \checkmark & \checkmark & $\underline{0.055}$ & $\mathbf{0.075}$ & $0.234$ & $\underline{0.121}$ & $\underline{0.026}$ & $\mathbf{0.028}$ & $0.082$ & $\underline{0.045}$ \\ 
        ---        & \checkmark & ---        & $0.642^\dagger$ & $0.111$ & $0.654^\dagger$ & $0.469$ & $0.255^\dagger$ & $0.040$ & $0.265^\dagger$ & $0.187$ \\ 
        ---        & ---        & \checkmark & $0.069$ & $0.101$ & $\underline{0.212}$ & $0.127$ & $0.028$ & $0.036$ & $\underline{0.075}$ & $0.046$ \\ 
        ---        & ---        & ---        & $\mathbf{0.048}$ & $0.085$ & $0.241$ & $0.125$ & $\mathbf{0.022}$ & $\underline{0.031}$ & $0.086$ & $0.046$ \\ 
        \bottomrule
    \end{tabular}%
    }
    \par\vspace{3pt}
    {\footnotesize $^\dagger$Generated samples collapse to a near-constant field.}
\end{table}

\paragraph{FEM radius and multiscale hierarchy depth}
We vary the radius multiplier of the FEM convolution applied to \GaussianBlobs. Throughout the experiments, we use a multiplier of $2$ as the default. The ablation for the radius multiplier is in Table~\ref{tab:abl_radius}. Here, we see that a small radius multiplier ($\times 1.0$) is not working as well, in particular for the coarse $1152$ points mesh. This is because in the lowest resolution with a small radius, the number of neighbours inside the radius is too small. Note, that increasing the radius also increases the computational cost as more neighbours are taken into account for the graph convolution. The experiments for the number of downsampling steps is in Table~\ref{tab:abl_multires}. 
We note that, although using the single scale gives good results, since the mesh is not downsampled, it is also more computationally heavy. The metrics remain relatively stable across different downsampling regimes.

%From these ablations we are able to conclude there is sensitivity to the multi-scale backbone, but the main effect is caused by the FEM-convolutional kernel.

\begin{table}[!ht]
    \centering
    \caption{Ablation of the FEM-Conv radius for GRIFDIR on \GaussianBlobs.}
    \begin{tabular}{lcccccc}
    \toprule
        & \multicolumn{3}{c}{MMD $\downarrow$} & \multicolumn{3}{c}{SWD $\downarrow$} \\
        \cmidrule(lr){2-4} \cmidrule(lr){5-7}
        Radius & $1152$ pts. & $2048$ pts. & $8192$ pts. & $1152$ pts. & $2048$ pts. & $8192$ pts. \\
        \midrule
        radius $\times 1.0$ & 0.3969 & 0.1273 & 0.1992 & 0.1378 & 0.0443 & 0.0709 \\
        radius $\times 2.0$ & \textbf{0.0585} & \textbf{0.0975} & \textbf{0.0287} & \textbf{0.0248} & \textbf{0.0353} & \textbf{0.0191} \\
        radius $\times 3.0$ & 0.0943 & 0.1087 & 0.1353 & 0.0345 & 0.0391 & 0.0549 \\
    \bottomrule
    \end{tabular}%
    \label{tab:abl_radius}
\end{table}

    % {\color{red} Note that none of these runs is using mass lumping! Maybe we should just show results on 2048 pts (train res.)}
% \begin{table}[!ht]
%     \centering
%     \caption{Component ablation on the Gaussian blob dataset.  }
%     \begin{tabular}{lcccccc}
%     \toprule
%     & \multicolumn{3}{c}{MMD $\downarrow$} & \multicolumn{3}{c}{SWD $\downarrow$} \\
%     \cmidrule(lr){2-4} \cmidrule(lr){5-7}
%     Depth & $1152$ pts. & $2048$ pts. & $8192$ pts. & $1152$ pts. & $2048$ pts. & $8192$ pts. \\
%     \midrule
%     $1$ --- $2048$ (single scale) & 0.086 & 0.111 & 0.098 & 0.032 & 0.040 & 0.035 \\
%     $2$ --- $2048/512$              & 0.503 & 0.102 & 0.049 & 0.178 & 0.036 & 0.024 \\
%     $3$ --- $2048/512/128$       &  0.571 & 0.103 & 0.423 & 0.201 & 0.036 & 0.146 \\
%     $4$ --- $2048/512/128/32$      & 0.166 & 0.100 & 0.548 & 0.036 & 0.059 & 0.200  \\
%     \bottomrule
%     \end{tabular}%
%     \label{tab:abl_multires}
% \end{table}
\begin{table}[!ht]
    \centering
    \caption{Component ablation on the Gaussian blob dataset ($2048$ points).}
    \begin{tabular}{lcc}
    \toprule
    Depth & MMD $\downarrow$ & SWD $\downarrow$ \\
    \midrule
    $1$ --- $2048$ (single scale)  & 0.111            & \underline{0.040} \\
    $2$ --- $2048/512$             & \underline{0.102} & \textbf{0.036} \\
    $3$ --- $2048/512/128$         & 0.103            & \textbf{0.036} \\
    $4$ --- $2048/512/128/32$      & \textbf{0.100}   & 0.059 \\
    \bottomrule
    \end{tabular}%
    \label{tab:abl_multires}
\end{table}

\subsection{Kernel choice}

% \paragraph{Regression Task} 
% To evaluate the effect of the kernel parametrisation from the rest of the architecture, we replace the full diffusion framework with a simpler supervised regression task: recovering a fixed Mexican-hat potential on a 2D domain from values at randomly placed points, using a single convolution layer. Each kernel is trained with Adam optimiser for 2,000 steps training budget on a mesh of $2048$ points, then evaluated without retraining at $512$ and $8192$ points. Quantitative results are in \cref{tab:mexican_hat_regression}. FEM-Conv outperforms both spline- and MLP-parametrised kernels at the largest patch size ($P=9$). At the lower capacity ($P=3$) the FEM-Conv variant is worse than the Spline kernel with $5$ and $9$ knots. The MLP kernel does not perform well under this setup and even reduces in accuracy with a larger width.

\paragraph{Comparison of kernels in diffusion model setup} 
We replace our FEM kernel (FEM-Conv) with an MLP kernel (see, e.g., \cite{gilmer2017neural}), a spline kernel \citep{fey2018splinecnn} (Spline-Conv), a standard GNO matrix kernel \citep{li2020neural} (GNO-Conv), and a MoNet-style Gaussian-mixture kernel (GMM-Conv) \citep{monti2017geometric}, keeping the rest of the architecture fixed. For the Spline kernel, the standard implementation uses a $k$-hop mesh-adjacency neighbourhood rather than a physical radius. We report both the original $k$-hop mesh-adjacency neighbourhood and a version using the physical-radius neighbourhood, to separate the effect of the kernel from the effect of the neighbourhood definition.
We evaluate the MMD and SWD; Results are in Table \ref{tab:gaussian_kernel_choice}. 

\begin{table}[!ht]
    \centering
    \caption{MMD and SWD for \GaussianBlobs on different resolutions for different choices of the kernel parametrisation in GRIFDIR. All models use the same backbone architecture.}
    \begin{tabular}{lcccccc}
    \toprule
        & \multicolumn{3}{c}{MMD $\downarrow$} & \multicolumn{3}{c}{SWD $\downarrow$} \\
        \cmidrule(lr){2-4} \cmidrule(lr){5-7}
        Kernel & $1152$ pts. & $2048$ pts. & $8192$ pts. & $1152$ pts. & $2048$ pts. & $8192$ pts. \\
        \midrule
        MLP-Conv  & 0.281 & 0.284 & 0.222 & 0.101 & 0.104 & 0.084 \\
        GNO-Conv  & 0.329 & 0.339 & - & 0.111 & 0.115 & - \\
        GMM-Conv  & 0.491 & 0.402 & 0.218 & 0.162 & 0.127 & 0.084 \\
        Spline-Conv (kNN)  & 0.315 & 0.098 & 0.384 &  0.106 & 0.035 & 0.134 \\
        Spline-Conv (radius)  & 0.647 & 0.085 & 0.070 & 0.256 & 0.030 & 0.039 \\
        FEM-Conv (ours) & 0.163 & 0.088 & 0.109 & 0.060 & 0.033 & 0.041 \\
    \bottomrule
    \end{tabular}%
    \label{tab:gaussian_kernel_choice}
\end{table}

\paragraph{Mexican Hat kernel recovery}

\begin{figure}
    \centering
    \includegraphics[width=1.0\linewidth]{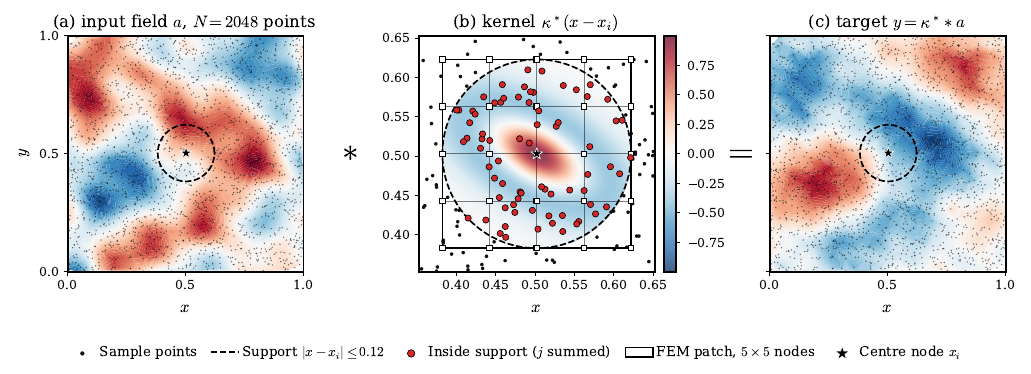}
    \caption{Operator-recovery task at $N=2{,}048$ points. (a) An input field $a$. (b) The target kernel $\kappa^*$ around one centre point $x_i$: the points inside the support $|x-x_i|\le 0.12$ (red) are summed, and the square is a $5\times5$ FEM-Conv patch. (c) The target $y$ that a single kernel layer must learn.}
    \label{fig:grifdir_mexican_toy}
\end{figure}

To isolate the kernel parametrisation from the rest of the architecture, we replace the diffusion model with a supervised operator-learning task solved by a single convolution layer (\cref{fig:grifdir_mexican_toy}). A fixed target kernel $\kappa^*$, a rotated anisotropic Mexican hat supported on the disc $|x|\le r$ with $r=0.12$, defines the operator $y_i=\sum_j w_j\,\kappa^*(x_i-x_j)\,a_j$ on $N$ uniformly random points in the unit square, where $w_j$ is the Voronoi area of point $j$ and the inputs $a$ are Gaussian random fields with length scale $0.1$. The support of each kernel is enforced being applied on an $r-$radius graph formed by the $N$ random point cloud. Each kernel is applied between a linear encoder and decoder and is trained on 64 input-output pairs at $N=2048$ for $2000$ Adam steps; all kernels share the same support, data and training budget. We compare our FEM-Conv with lumped quadrature, the same FEM kernel with mean aggregation, SplineConv which has mean aggregation \cite{fey2018splinecnn}, and an MLP kernel that takes as input the offset and distance, $(\Delta x,\Delta y,|\Delta x|)/r$, with lumped or mean aggregation. For FEM-Conv and SplineConv the resolution is an $R\times R$ node grid on $[-r,r]^2$; for the MLP it is a hidden width of $R^2$. Each model is linear in its input, so its learned kernel can be read out exactly and compared with $\kappa^*$.

\begin{figure}
    \centering
    \includegraphics[width=1.0\linewidth]{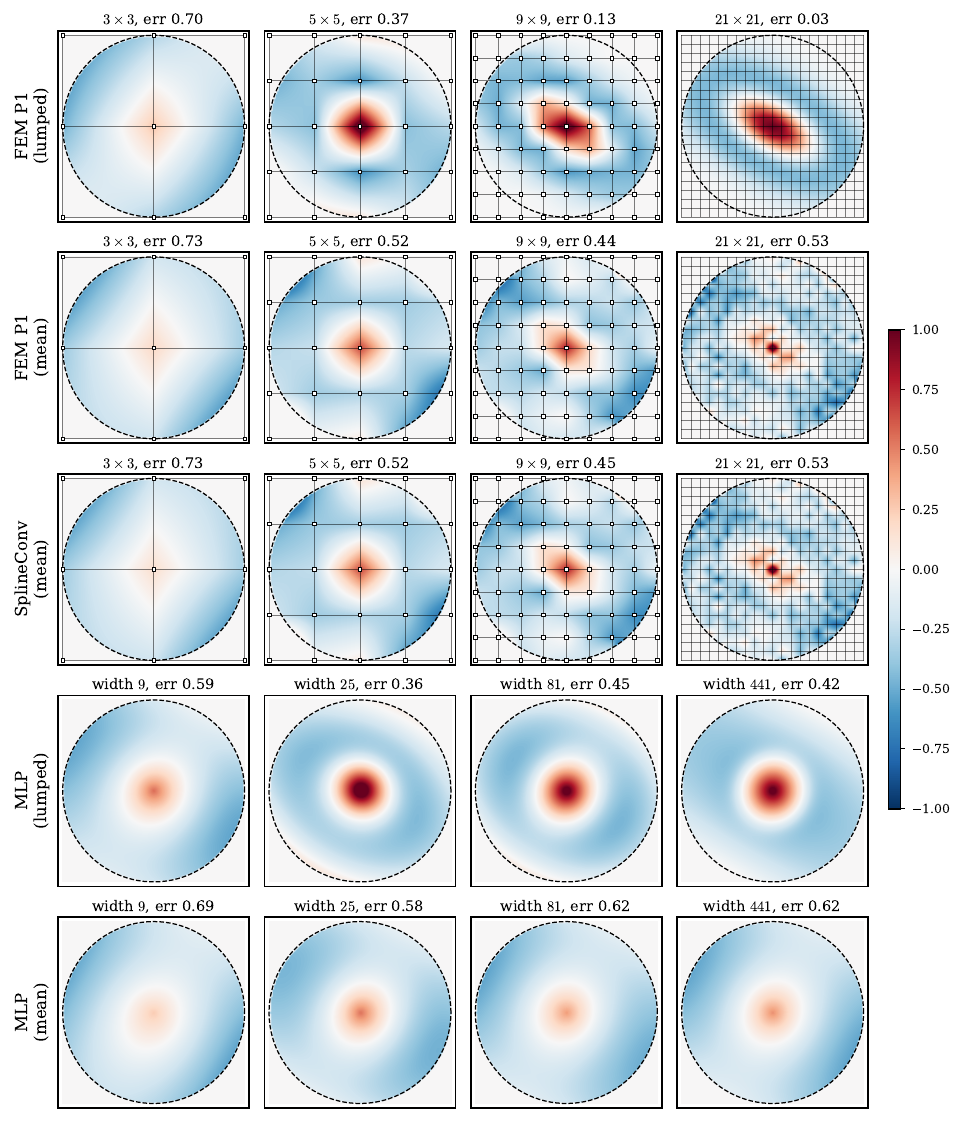}
    \caption{Kernels recovered by a single layer trained at $N=2{,}048$ points, at increasing resolution: $R\times R$ nodes for FEM-Conv and SplineConv, hidden width $R^2$ for the MLP.}
    \label{fig:grifdir_mexican_kernals}
\end{figure}

After training, each model is evaluated without retraining on new input fields and new point clouds of $512$, $2048$ and $8192$ points; we report the relative MSE of its output $y$ and the relative $L^2$ error of its kernel against $\kappa^*$. Recovered kernels are evaluated on the relative $L^2$ error on the support (dashed circle), outside which no convolution evaluates the kernel. 
All mean aggregation kernels divide by the number of neighbours, so each kernels graph is smaller than $\kappa^*$ by a near-constant factor; each kernel is therefore scaled back towards $\kappa^*$ by a single least-squares factor and the remaining error is reported. 
Full results are shown in Table~\ref{tab:mexican_hat_regression} and the recovered kernels are plotted in Figure~\ref{fig:grifdir_mexican_kernals}.

In the special case of using a degree-1 B-spline kernel, the SplineConv (mean aggregation) recovers the FEM-conv (mean aggregation) and give the same kernels and errors at every resolution as both use piecewise-linear basis functions on a regular grid. SplineConv applies its basis to mesh-normalised pseudo-coordinates and averages over neighbours, whereas FEM-Conv defines the kernel on a fixed physical patch and aggregates with the lumped finite-element quadrature. Only lumped FEM-Conv recovers the rotated, anisotropic target converges to the true kernel. Its kernel error falls from $0.70$ at $3\times3$ to $0.13$ at $9\times9$ and $0.03$ at $21\times21$. This is in accordance with Proposition~\ref{prop:mesh_refinement_consistency}. The MLP kernel failed on the task. Although its input, $(\Delta x,\Delta y,|\Delta x|)/r$, carries the information to learn the direction, it learns a nearly isotropic filter.

% \begin{table}[!ht]
%     \centering
%     \caption{Regression error (MSE) when fitting a fixed Mexican-hat potential from function values at randomly placed points, using a single convolution layer per kernel parametrisation. All kernels are trained with the same optimiser and training budget on a mesh of $2048$ points, and evaluated without retraining at $512$ and $8192$ points. Best value \textbf{bold}.}
%     \begin{tabular}{lllll}
%     \toprule
%         Kernel & Kernel info. & $512$ pts. $\downarrow$ & $2{,}048$ pts. $\downarrow$ & $8{,}192$ pts. $\downarrow$ \\
%         \midrule
%         FEM Conv & $3\times3$ patch ($P=3$) & $0.215$ & $0.016$ & $0.005$ \\ 
%         FEM Conv & $5\times5$ patch ($P=5$) & $0.059$ & $0.006$ & $0.008$ \\ 
%         FEM Conv & $9\times9$ patch ($P=9$) & $\mathbf{0.011}$ & $\mathbf{0.002}$ & $\mathbf{0.0004}$ \\
%         Spline basis & $3$ knots & $0.306$ & $0.079$ & $0.048$ \\ 
%         Spline basis & $5$ knots & $0.196$ & $0.071$ & $0.046$ \\ 
%         Spline basis & $9$ knots & $0.163$ & $0.068$ & --- \\ 
%         MLP kernel & width $9$ & $0.33$ & $0.085$ & $0.055$ \\ 
%         MLP kernel & width $25$ & $0.301$ & $0.083$ & $0.052$ \\ 
%         MLP kernel & width $81$ & $0.437$ & $0.176$ & $0.143$ \\ 
%         \bottomrule
%     \end{tabular}\label{tab:mexican_hat_regression}
% \end{table}

\begin{table}[!ht]
    \centering
    \small
    \caption{Mexican-hat kernel recovery with a single convolution layer trained on $2{,}048$ points: relative MSE of the output $y$ on new point clouds, and relative $L^2$ error of the recovered kernel against $\kappa^*$; mean of two seeds. Best value \textbf{bold}.}
    \begin{tabular}{llcccc}
    \toprule
         & & \multicolumn{3}{c}{Output rel.\ MSE $\downarrow$} & Kernel \\
        \cmidrule(lr){3-5}
        Kernel & Kernel info. & $512$ pts. & $2048$ pts. & $8192$ pts. & rel.\ $L^2$ $\downarrow$ \\
        \midrule
        FEM Conv (lumped) & $3\times3$ patch & $0.22$ & $0.016$ & $0.0046$ & $0.70$ \\
        FEM Conv (lumped) & $5\times5$ patch & $0.059$ & $0.0056$ & $0.00079$ & $0.37$ \\
        FEM Conv (lumped) & $9\times9$ patch & $\mathbf{0.011}$ & $0.0018$ & $\mathbf{0.00039}$ & $0.13$ \\
        FEM Conv (lumped) & $21\times21$ patch & $0.016$ & $\mathbf{0.00059}$ & $0.00043$ & $\mathbf{0.02}$ \\
        \midrule
        FEM Conv (mean) & $3\times3$ patch & $0.28$ & $0.087$ & $0.059$ & $0.73$ \\
        FEM Conv (mean) & $5\times5$ patch & $0.19$ & $0.070$ & $0.046$ & $0.52$ \\
        FEM Conv (mean) & $9\times9$ patch & $0.16$ & $0.067$ & $0.046$ & $0.44$ \\
        FEM Conv (mean) & $21\times21$ patch & $0.25$ & $0.067$ & $0.054$ & $0.56$ \\
        \midrule
        SplineConv & $3\times3$ knots & $0.31$ & $0.079$ & $0.048$ & $0.73$ \\
        SplineConv & $5\times5$ knots & $0.20$ & $0.071$ & $0.046$ & $0.52$ \\
        SplineConv & $9\times9$ knots & $0.16$ & $0.067$ & --- & $0.47$ \\
        SplineConv & $21\times21$ knots & $0.27$ & $0.070$ & $0.052$ & $0.56$ \\
        \midrule
        MLP (lumped) & width $9$ & $0.17$ & $0.013$ & $0.0039$ & $0.64$ \\
        MLP (lumped) & width $25$ & $0.14$ & $0.011$ & $0.0036$ & $0.52$ \\
        MLP (lumped) & width $81$ & $0.076$ & $0.0072$ & $0.0011$ & $0.45$ \\
        MLP (lumped) & width $441$ & $0.063$ & $0.0053$ & $0.0016$ & $0.34$ \\
        \midrule
        MLP (mean) & width $9$ & $0.28$ & $0.078$ & $0.051$ & $0.70$ \\
        MLP (mean) & width $25$ & $0.23$ & $0.078$ & $0.053$ & $0.64$ \\
        MLP (mean) & width $81$ & $0.24$ & $0.079$ & $0.053$ & $0.63$ \\
        MLP (mean) & width $441$ & $0.31$ & $0.17$ & $0.14$ & $0.62$ \\
    \bottomrule
    \end{tabular}\label{tab:mexican_hat_regression}
\end{table}

%%%%%%

\subsection{DeepJEB Scaling and additional Samples}\label{app:deepjeb}

\paragraph{Training} We consider training two models, one capped at \num{6000} nodes and the second at \num{24000} nodes.
From this we find that training scales sub-linearly with the number of points: For both models we train for $30$ epochs and get a $1.48\times$ wall time increase for $4\times$ the number of points ($7.7$ hours for \num{6000} nodes to $11.4$ hours for \num{24000} nodes).

\paragraph{Sampling} \cref{tab:timing_deepjeb} shows a scaling law for the sampling time for increasing node densities. We report the sampling cost per bracket for 200 sampling steps and average over 8 held-out brackets. At the higher resolution caps some test bracket data samples have fewer nodes than the cap, so we also report the average number of nodes. Sampling cost grows sub-linearly with resolution: a $4.6\times$ increase in nodes (i.e. going from \num{6000} to \num{27700} nodes) costs $3.5\times$ the sampling time.

\begin{table}[!ht]
    \centering
    \caption{Sampling time of DeepJEB}
    \begin{tabular}{llllll}
    \toprule
        Cap & Avg. nodes & Sampling (s) & Node scale & Time scale & Time / node \\
        \midrule
        \num{6000} & \num{6000} & $45.6$ & $1.00$ & $1.00$ & $1.00$ \\ 
        \num{12000} & \num{12000} & $59.3$ & $2.00$ & $1.30$ & $0.65$ \\ 
        \num{24000} & \num{23427} & $122.7$ & $3.90$ & $2.69$ & $0.69$ \\ 
        \num{32000} & \num{27702} & $158.6$ & $4.62$ & $3.48$ & $0.75$ \\ 
        \bottomrule
    \end{tabular}
    \label{tab:timing_deepjeb}
\end{table}

\begin{figure}[t]
    \centering

    \begin{subfigure}{0.24\textwidth}
        \centering
        \captionsetup[subfigure]{justification=centering}
        \includegraphics[width=\linewidth]{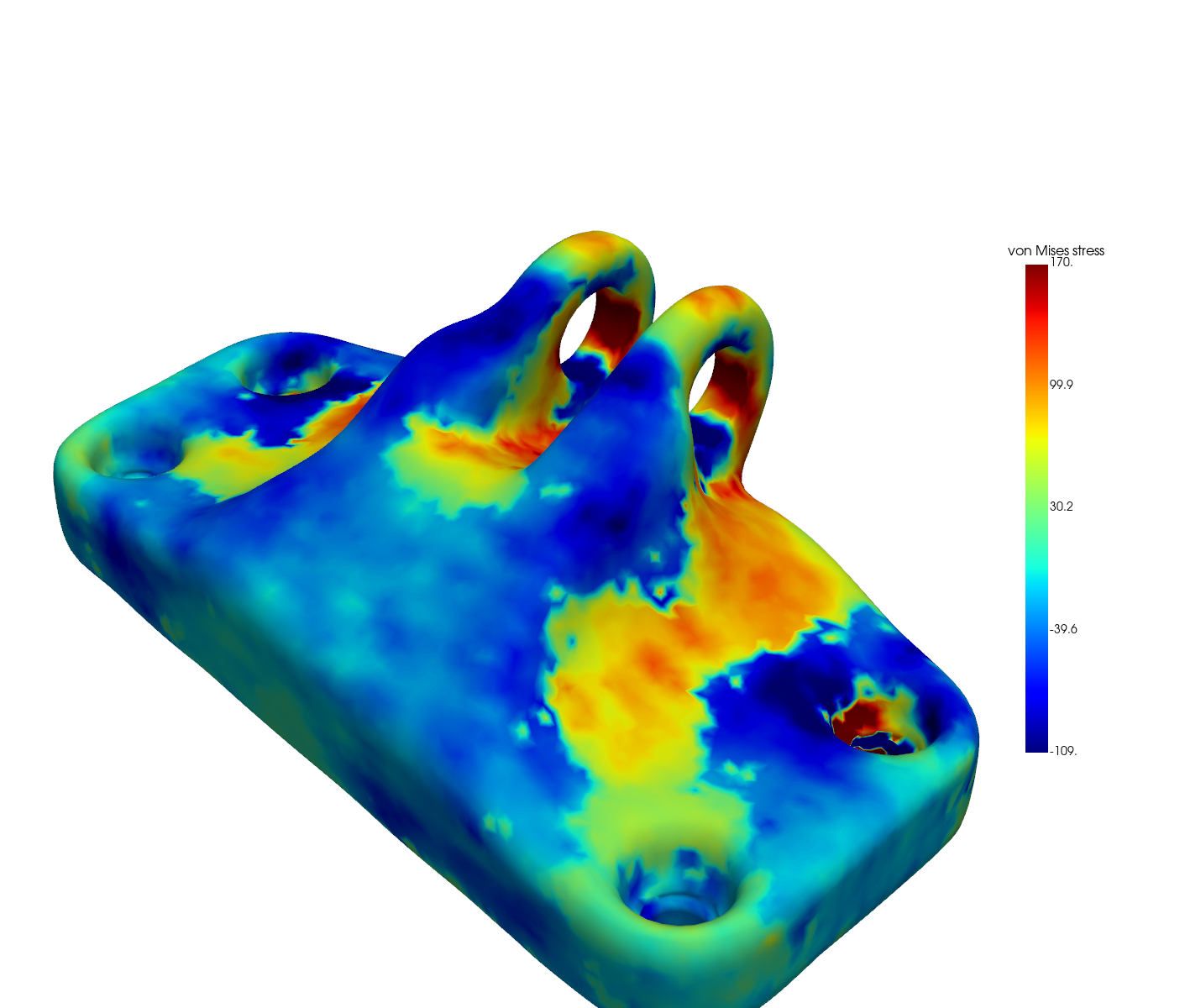}
        \caption{Generated sample \\ (load index $0$)}
    \end{subfigure}
    \hfill
    \begin{subfigure}{0.24\textwidth}
        \centering
        \captionsetup[subfigure]{justification=centering}
        \includegraphics[width=\linewidth]{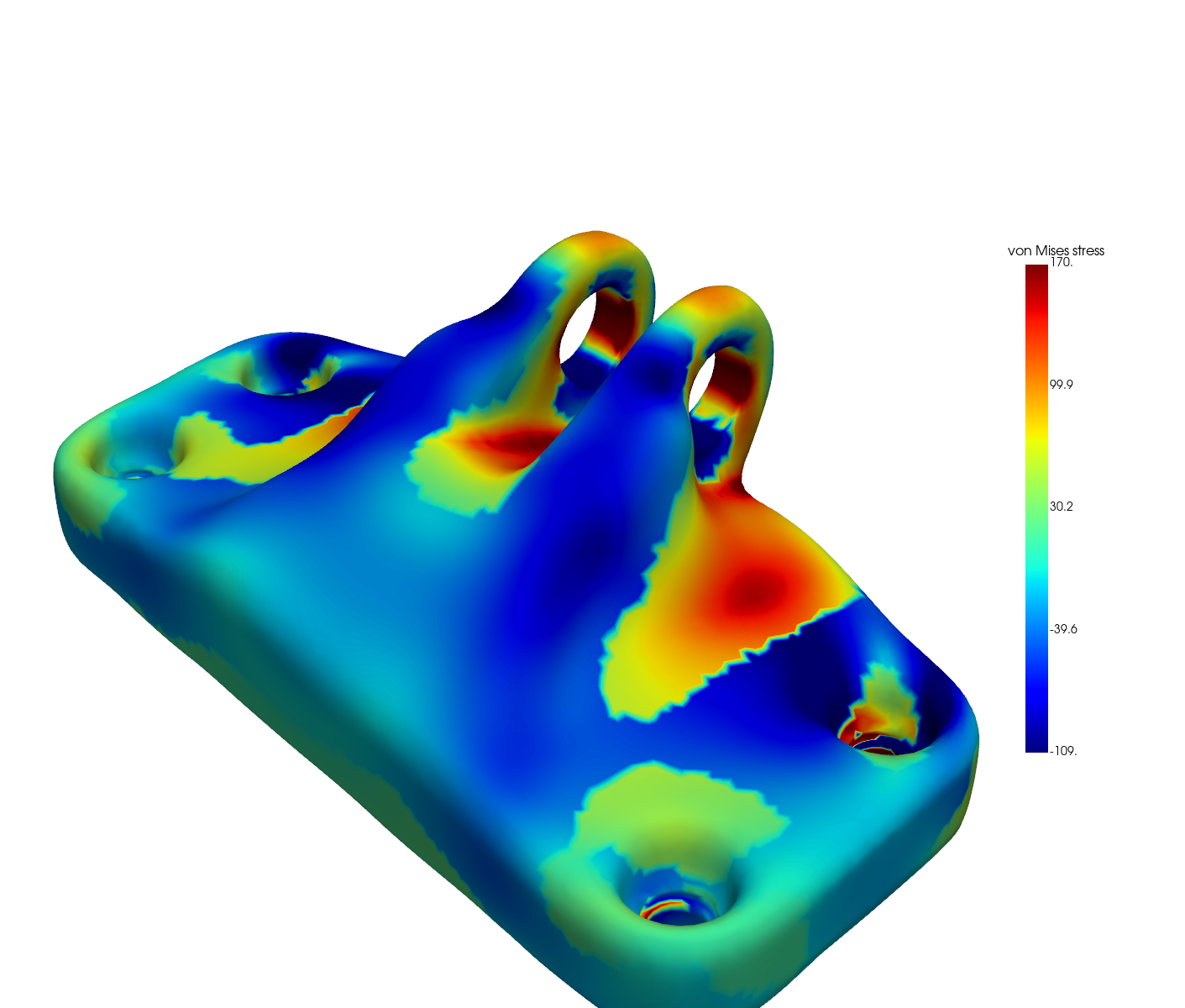}
        \caption{Ground truth \\ (load index $0$)}
    \end{subfigure}
    \hfill
    \begin{subfigure}{0.24\textwidth}
        \centering
        \captionsetup[subfigure]{justification=centering}
        \includegraphics[width=\linewidth]{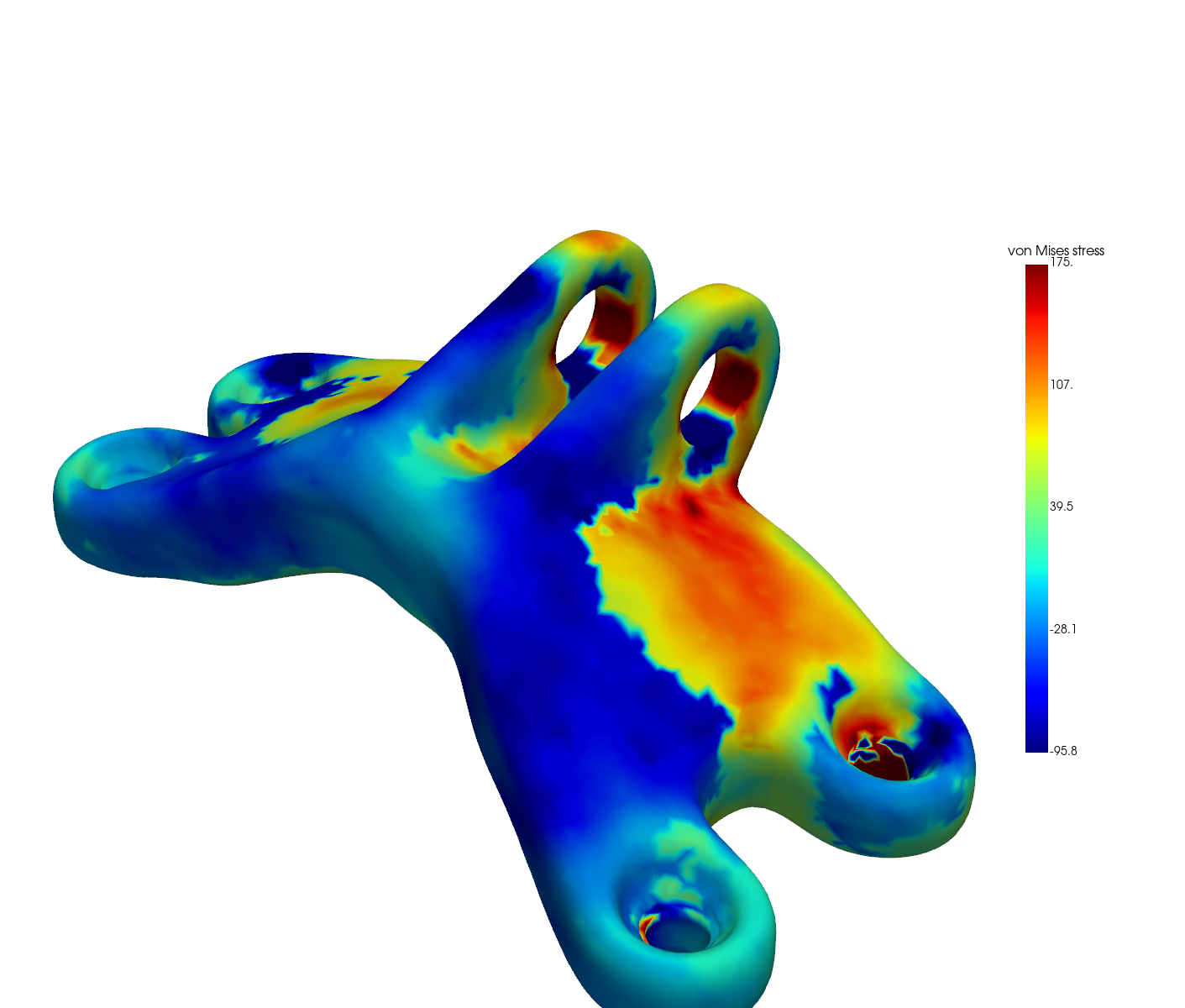}
        \caption{Generated sample \\ (load index $10$)}
    \end{subfigure}
    \hfill
    \begin{subfigure}{0.24\textwidth}
        \centering
        \captionsetup[subfigure]{justification=centering}
        \includegraphics[width=\linewidth]{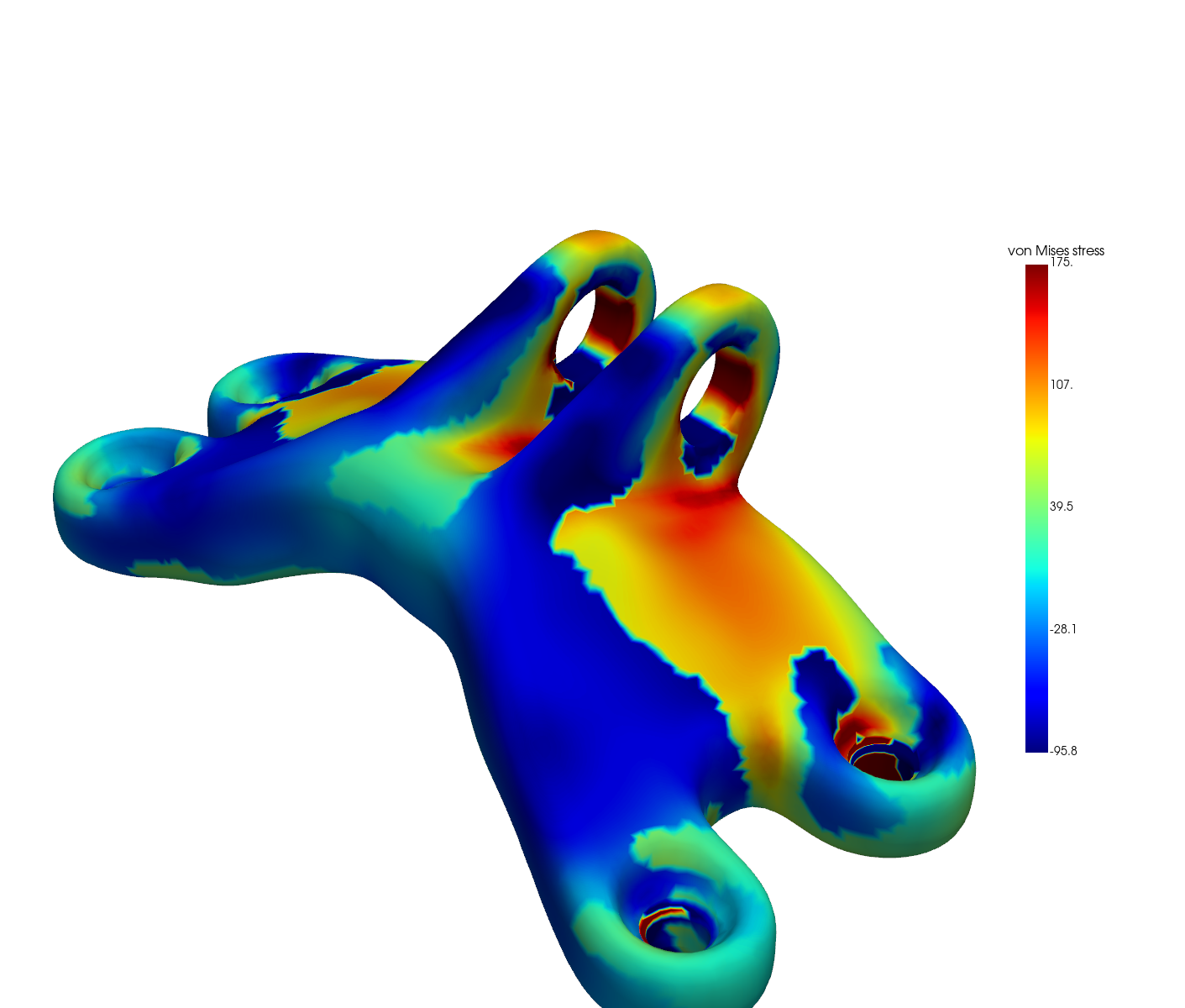}
        \caption{Ground truth \\ (load index $10$)}
    \end{subfigure}

    \caption{Conditional generation with \DeepJEB. The GRIFDIR model is conditioned on both the 3D bracket geometry and distance to the bracket surface. The figure compares generated von Mises stress fields with the corresponding ground-truth fields for two vertical-load conditions. Samples are generated from the model trained and evaluated at \num{24000} points without lumping from Table~\ref{tab:deepjeb} with 200 sample steps.}
    \label{fig:deep_jeb_conditional}
\end{figure}

\end{document}